%% file: main.tex
\documentclass[10pt,twocolumn,letterpaper]{article}

\usepackage{cvpr}              

\input{configs/packages.tex}

\definecolor{cvprblue}{rgb}{0.21,0.49,0.74}
\usepackage[pagebackref,breaklinks,colorlinks,allcolors=cvprblue]{hyperref}

\usepackage[capitalize]{cleveref}
\input{configs/commands.tex}

\def\confName{CVPR}
\def\confYear{2027}

\title{Progress Predicts Policy}

\author{Licheng Zhong \quad Gim Hee Lee\\
Department of Computer Science, National University of Singapore\\
{\tt\small zlicheng@u.nus.edu, gimhee.lee@nus.edu.sg}
}

\begin{document}
\twocolumn[{%
        \renewcommand\twocolumn[1][]{#1}%
        \maketitle
        \vspace{-1.5em}
        \begin{center}
            \includegraphics[width=\linewidth]{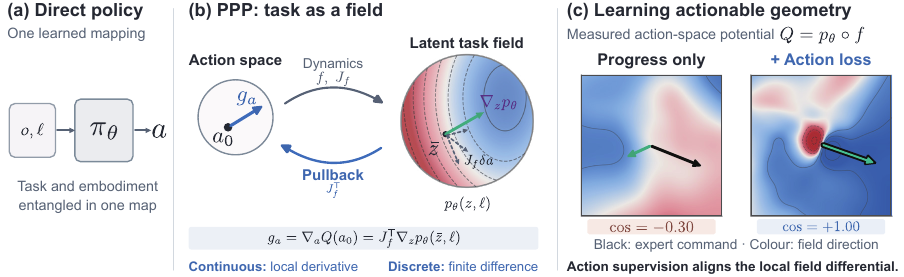}\vspace{-0.5em}
            \captionof{figure}{\textbf{Learning a field through its induced actions.}
                (a) A direct policy maps observations and instructions to actions.
                (b) PPP reads commands from a scalar field's differential, using a shared world model for action conditioning.
                (c) Measured action-space potential slices at a demonstration state show that progress supervision alone yields a gradient misaligned with the expert command (black).
                Action supervision aligns the gradient with that command.
                Project page: \url{https://zlicheng.com/ppp_page/}
            }
            \label{fig:teaser}
        \end{center}
        \vspace{0.6em}
    }]
\input{sec/0_abstract}
\input{sec/1_intro}
\input{sec/2_related}
\input{sec/3_method}
\input{sec/4_experiments}
\input{sec/5_conclusion}
\section*{LLM Usage Statement}
ChatGPT, Codex, and Claude Code were used to assist with code development, manuscript drafting and editing, and figure design feedback. These tools were not used to generate or alter experimental data, results, or figures. The authors reviewed all scientific content and take full responsibility.
    {
        \small
        \bibliographystyle{ieeenat_fullname}
        \bibliography{refers}
    }

\appendix
\input{sec/X_supplementary}

\end{document}

%% file: configs/packages.tex
\usepackage{graphicx}
\usepackage{amsmath}
\usepackage{amssymb}
\usepackage{booktabs}
\usepackage{balance}
\usepackage{lipsum}
\usepackage{caption}
\usepackage{algorithm}
\usepackage{algorithmic}
\usepackage{enumerate}
\usepackage{engord}
\usepackage{bm}
\usepackage{bbm}
\usepackage{amstext}
\usepackage{comment}
\usepackage{color}
\usepackage{booktabs}
\usepackage{multirow}
\usepackage{mathtools}
\usepackage{setspace}
\usepackage{pifont}
\usepackage{dsfont}
\usepackage{cancel}
\usepackage[inline]{enumitem}
\usepackage[toc,page]{appendix}
\usepackage{hhline}
\usepackage{adjustbox}
\usepackage{makecell}

%% file: configs/commands.tex
\definecolor{Gray}{gray}{0.9}
\definecolor{White}{gray}{1}
\definecolor{WhiteGray}{rgb}{0.9, 0.9, 0.9}
\definecolor{DGray}{gray}{0.8}
\definecolor{DDDDGray}{gray}{0.3}
\definecolor{citecolor}{HTML}{0071bc}

\definecolor{DeltaColor}{rgb}{0.039,0.73,0.71}
\definecolor{SigmaColor}{rgb}{0.98,0.45,0.0}
\definecolor{AlphaColor}{rgb}{0,0,0.8}
\definecolor{BetaColor}{rgb}{0.8,0,0.8}
\definecolor{GammaColor}{rgb}{0.514,0.34,0.224}
\definecolor{EpsilonColor}{rgb}{0.353,0.725,0.906}
\definecolor{GreenColor}{rgb}{0.137,0.573,0.565}
\definecolor{RedColor}{rgb}{0.949,0.275, 0.224}
\definecolor{Cardinal}{rgb}{0.549,0.082,0.082}

\DeclareMathAlphabet\mathbfcal{OMS}{cmsy}{b}{n}

\newcommand{\colorRef}[1]{\textcolor{black}{#1}}
\crefname{figure}{\colorRef{Fig.}}{\colorRef{Figs.}}
\Crefname{figure}{\colorRef{Figure}}{\colorRef{Figures}}
\crefname{section}{\colorRef{Sec.}}{\colorRef{Secs.}}
\Crefname{section}{\colorRef{Section}}{\colorRef{Sections}}

\crefname{table}{\colorRef{Tab.}}{\colorRef{Tabs.}}
\Crefname{table}{\colorRef{Table}}{\colorRef{Tables}}
\crefname{equation}{\colorRef{Eq.}}{\colorRef{Eqs.}}
\Crefname{equation}{\colorRef{Equation}}{\colorRef{Equation}}

\newcommand\method{PPP\xspace}
\newcommand{\qheading}[1]{\noindent\mbox{\textbf{#1}\;}}



%% file: sec/0_abstract.tex
\begin{abstract}
    A task specifies how the world should change, while a policy specifies what the robot should do.
    We introduce Progress Predicts Policy (PPP), which learns a language-conditioned scalar field through the commands induced by its differential.
    We obtain action conditioning by composing a latent progress field with a shared world model and read its action gradient at a fixed reference command.
    Continuous commands scale this gradient by a learned positive gain, while discrete commands use finite differences of the same scalar.
    This yields a hybrid policy without a separate action head or iterative search.
    Fitting demonstration progress values alone does not ensure that action gradients align with expert commands.
    PPP trains the induced commands by imitation and uses double backpropagation for the continuous gradient readout to shape the field's local control geometry.
    Demonstration timing calibrates field values on observed states without additional annotation.
    The continuous readout remains learnable without progress labels but these timing-derived targets support progress monitoring during execution.
    Experiments in simulated and real robot manipulation show that PPP outperforms behavioral cloning under a shared architecture.
\end{abstract}
\vspace{-5mm}

%% file: sec/1_intro.tex
\section{Introduction}
\label{sec:intro}

Imitation learning has produced capable visuomotor policies.
Diffusion policies model expressive action distributions~\cite{chi2023diffusionpolicy} and vision-language-action (VLA) models bring web-scale visual and linguistic knowledge to control~\cite{kim2025openvla,black2025pi0}.
These methods learn a map from observations and instructions to actions.
A task and a policy play different roles, however.
A task states what should change in the world: the bowl should end up on the plate.
The robot dynamics determine which changes its commands can produce.
We ask whether a learned scalar task evaluator can provide progress estimates and a differential from which a policy reads its commands.

We represent a \emph{task as a field}: a language-conditioned scalar progress field $p_\theta(z,\ell)$ over learned latent states $z$.
Timing labels supervise its values on observations while expert commands supervise its differential-induced actions.
Reading an action derivative requires dependence on a candidate command.
We obtain this dependence by composing the field with a shared world model, $Q_{z,\ell}(a)=p_\theta(f(z,a),\ell)$.
The chain rule at a fixed reference $a_0$ gives:
\begin{equation}
    g_a=\nabla_a Q_{z,\ell}(a_0)=J_f^\top\,\nabla_z p_\theta(\bar z,\ell),
    \label{eq:intro_pullback}
\end{equation}
where $\bar z=f(z,a_0)$ and $J_f$ is the action Jacobian of the learned control-query map at $a_0$.
\cref{fig:teaser} illustrates the pairing of task sensitivity with the local latent response of the query map.
Imitation trains both factors through the induced command.
Continuous commands scale $g_a$ by a learned positive gain while discrete commands use finite differences of $Q$.
One gradient evaluation supplies continuous commands without a separate action generator or iterative search.
The same field provides progress estimates on observations and control readouts on action-conditioned latents.
\cref{sec:exp_rq2} shows that a directly action-conditioned scalar yields an effective continuous gradient policy.

Fitting progress values alone does not ensure that this differential yields useful commands.
Progress labels constrain $p_\theta$ only at demonstrated states.
Functions agreeing on those values can have different nearby derivatives, which $g_a$ reads.
\cref{fig:grad_alignment} shows that progress-only fields rank training demonstration states almost perfectly but produce action gradients that align only weakly with expert commands.
A second gap separates local from global use of the field.
Searching for finite commands that maximize $Q$ need not recover the demonstrated behavior and almost never succeeds in our experiments.
The central learning problem is therefore to learn a \emph{control-aligned progress field} with actionable local geometry rather than merely an accurate progress estimator.

We introduce \textbf{Progress Predicts Policy (\method)}, which trains a task field through the behavior it induces.
Demonstrations supervise the field in two complementary ways: expert commands supervise the action produced by $g_a$ while demonstration-relative progress labels calibrate its values without extra annotation.
Imitation jointly shapes the field geometry and the action response of the query model.
Under a shared architecture, \method outperforms behavioral cloning (BC) on LIBERO~\cite{liu2023libero} and real robots.
Auxiliary progress prediction leaves BC essentially unchanged.
\cref{sec:exp_rq3} shows that progress labels are optional for continuous control and useful for execution monitoring.
Our contributions are:

(i) a scalar-field policy that reads continuous commands from action gradients and discrete commands from finite differences of the same scalar;

(ii) a training formulation that supervises induced commands by imitation and calibrates values on observed states using annotation-free demonstration timing;

(iii) an empirical study that separates progress-value fit from action-gradient alignment and closed-loop performance and characterizes the local scope of the learned readout.

%% file: sec/2_related.tex
\section{Related Work}
\label{sec:related}

\smallskip
\qheading{Visuomotor policies and world models.}
Most visuomotor policies predict actions from observations~\cite{zhao2023act,ghosh2024octo}.
Diffusion policies model multimodal action distributions~\cite{chi2023diffusionpolicy,ze2024dp3} while VLA models add action outputs to pretrained vision-language backbones~\cite{zitkovich2023rt2,kim2025openvla,black2025pi0}.
World models support planning through visual features in DINO-WM~\cite{zhou2025dinowm} and action-conditioned video prediction in V-JEPA~2~\cite{assran2025vjepa2}.
LeWorldModel learns latent dynamics from pixels~\cite{maes2026leworldmodel}.
Other predictive models couple prediction with action generation~\cite{lyu2025dywa,ye2026dreamzero,wu2024gr1,tian2025seer}.
Model-based policy learning differentiates through learned dynamics to train a separate actor~\cite{heess2015svg,hafner2020dreamer}.
\method reads commands from a scalar field through the world model without a separate action head.

\smallskip
\qheading{Reward, value, and progress representations.}
Visual and language-conditioned representations can induce reusable rewards~\cite{ma2023vip,ma2023liv,sontakke2023roboclip}.
GOMP uses a graspability field for reward shaping and termination in non-prehensile manipulation~\cite{zhong2026gomp}.
VLAC and Robometer scale progress and preference estimation across robot data~\cite{zhang2026vlac,liang2026robometer}.
Progress estimates inform action generation through reinforcement learning feedback in VLAC~\cite{zhang2026vlac} and verification of policy-proposed candidates in TapSampling~\cite{zhao2026tapsampling}.
The closest work is ProgressVLA, which backpropagates progress through a world model to guide a latent-action diffusion policy~\cite{yan2026progressvla}.
This parallels reward-gradient guidance of trajectory diffusion in Diffuser~\cite{janner2022diffuser}.
These methods guide an independent action generator.
In \method, the scaled pulled-back progress differential defines the continuous command and is trained by imitation.
Inverse reinforcement learning learns objectives to match demonstrated behavior~\cite{abbeel2004apprenticeship,ziebart2008maxent} but recovers actions by optimization.

\smallskip
\qheading{Potential-based control and differentiable planning.}
Classical robotics generates motion from scalar potential gradients.
Artificial potential fields steer manipulators away from obstacles and toward goals~\cite{khatib1986potential} while navigation functions remove spurious minima by construction~\cite{rimon1992navigation}.
These potentials are designed in configuration space for a known goal.
Learning-from-demonstration methods fit stable dynamical systems~\cite{khansari2011seds,rana2020euclideanizing} or potential functions~\cite{khansari2017potential} to demonstrated motion in robot state space.
Geometric policy frameworks compose task-space controllers via pullbacks~\cite{cheng2020rmpflow,mukadam2020rmpfusion}.
\method extends this perspective to hybrid control through a language-conditioned perceptual latent potential and learned, rather than known, dynamics.
Implicit behavioral cloning selects actions by minimizing a learned energy at inference~\cite{florence2022implicit}.
Differentiable planners learn through action optimization~\cite{amos2018differentiablempc,srinivas2018universal,wan2024difftori} and GRASP improves gradient-based planning with visual world models~\cite{psenka2026grasp}.
TD-MPC methods use latent models and learned values for sampling-based planning~\cite{hansen2022tdmpc,hansen2024tdmpc2}.
\method reads continuous commands with one gradient evaluation and no inference-time optimization.
Supervising the induced command connects \method to Sobolev training, which supervises network derivatives alongside values~\cite{czarnecki2017sobolev}.

%% file: sec/3_method.tex
\input{sec/fig/pipeline.tex}

\section{Method}
\label{sec:method}

\method learns a scalar task field through the commands induced by its action-conditioned differential (\cref{fig:pipeline}).


\subsection{Representing Tasks as Progress Fields}
\label{sec:method_field}

A visual encoder $E$ maps an observation $o$ to a latent state $z=E(o)\in\mathbb{R}^d$.
The instruction is encoded separately as an embedding $\ell$.
A world model learns latent transitions under continuous commands $a\in\mathbb{R}^m$ and supplies the control representation through its action-conditioned query $f(z,a)$.
We instantiate $E$ and $f$ with LeWM~\cite{maes2026leworldmodel} and suppress the latent and action history of $f$ in the notation.
We represent the task as a language-conditioned scalar potential:
\begin{equation}
    p_\theta:\ \mathbb{R}^d\times\mathcal{L}\rightarrow\mathbb{R},\qquad (z,\ell)\mapsto p_\theta(z,\ell),
    \label{eq:field}
\end{equation}
We call $p_\theta(\cdot,\ell)$ the \emph{task field} of instruction $\ell$.
Action supervision shapes its local differential into a control signal while timing labels calibrate values on observations to demonstration-relative progress

\smallskip
\qheading{Demonstration-relative progress.}
For a demonstrated observation $o_i$, we denote its elapsed time by $t_i$ and the terminal time on the world model transition grid by $T$.
We consider two progress targets.
\emph{Normalized phase} uses $p_i^\star=\rho_i-1$, where $\rho_i\in[0,1]$ is the normalized time within the demonstration.
\emph{Absolute time-to-go} uses $p_i^\star=-(T-t_i)/T_u$, with a fixed time unit $T_u>0$ shared across demonstrations.
Both targets are zero at completion but differ in how they represent duration.
Normalized phase starts at $-1$ for every demonstration while time-to-go retains differences in remaining duration across demonstrations.
For either form we fit values and increments for pairs $i<j$ from the same demonstration:
\begin{equation}
    \begin{aligned}
        \mathcal L_{\mathrm{val}} & =\tfrac12\textstyle\sum_{x\in\{i,j\}}\bigl(p_x-p_x^\star\bigr)^2, \\
        \mathcal L_{\mathrm{inc}} & =\bigl[(p_j-p_i)-(p_j^\star-p_i^\star)\bigr]^2,
    \end{aligned}
    \label{eq:progress_supervision}
\end{equation}
with $p_x=p_\theta(E(o_x),\ell)$.
We also use a temporal ranking loss, a terminal boundary loss, and a contrast loss that scores the true instruction above a mismatched one.
Their weighted sum defines $\mathcal L_p$.
These losses calibrate demonstration-relative progress but do not guarantee physical progress.
\cref{sec:limit} discusses this limitation.

\smallskip
\qheading{Values do not determine the differential.}
$\mathcal L_p$ constrains $p_\theta$ only at states the demonstrations visit.
Adding a smooth function that vanishes on those states preserves the fitted values while allowing derivatives in unconstrained directions to differ.
Composition with the world model transfers this freedom to the action gradient.
Even if $f(z,a_E)$ matches the next demonstrated state, the labels constrain only the progress difference along the expert query:
$p_\theta(f(z,a_E),\ell)-p_\theta(z,\ell)\approx p_{t+k}^\star-p_t^\star$.
They do not determine values at other candidate commands or the direction of fastest increase in the composed scalar.
Progress supervision can therefore calibrate values without producing a useful control direction.
\cref{sec:exp_rq2} measures this gap.

\subsection{Action Conditioning and Local Control}
\label{sec:method_pullback}

To define the action-dependent scalar used for control, we compose the field with the control-query branch of the world model:
\begin{equation}
    Q_{z,\ell}(a)=p_\theta\bigl(f(z,a),\ell\bigr).
    \label{eq:Q}
\end{equation}
Control queries share predictor weights but use separate normalization statistics, fixed with respect to candidate actions to avoid BatchNorm coupling between samples.
Thus $Q$ is trained for action readout; its values need not be calibrated forecasts of candidate outcomes.

\smallskip
\qheading{Action conditioning with a world model.}
The readout requires dependence on a candidate command.
A world model provides a natural interface by mapping commands into the latent space on which the task field is defined.
We reuse this interface to construct $Q=p_\theta\circ f$ and expose the chain-rule structure of the readout.
\cref{sec:exp_rq2} evaluates Direct-AP as an alternative that retains the world model backbone but supplies the command directly to the scalar.

\smallskip
\qheading{Local control geometry.}
We use a fixed reference command $a_0$ at normalized zero, which corresponds to the corpus median command.
Let $\bar z=f(z,a_0)$ and $J_f=\partial f/\partial a|_{a_0}$ denote the reference latent and action Jacobian.
The chain rule gives:
\begin{equation}
    \boxed{\,g_a:=\nabla_aQ_{z,\ell}(a_0)=J_f^\top\,\nabla_zp_\theta(\bar z,\ell).\,}
    \label{eq:pullback}
\end{equation}
$J_f^\top$ pulls the task differential back through the query map under the Euclidean metric of normalized action coordinates.
The product $g_a^\top\delta a=\nabla_zp_\theta^\top(J_f\delta a)$ pairs task sensitivity with the first-order latent response of that map.
One reverse-mode pass computes this gradient exactly without forming $J_f$.
For a perturbation $\delta a$, the Taylor expansion is:
\begin{equation}
    Q_{z,\ell}(a_0+\delta a)-Q_{z,\ell}(a_0)=g_a^\top\delta a+O(\|\delta a\|^2),
    \label{eq:taylor}
\end{equation}
For a fixed direction $v$ with $g_a^\top v>0$, sufficiently small positive steps $\delta a=\alpha v$ increase $Q$.
This also holds for smooth direct action-conditioned scalars, which learn the combined dependence implicitly.
Factorization exposes the geometry but is not required for local ascent.
This local argument does not guarantee physical progress or improvement at the deployed step.
\cref{sec:exp_rq4} examines the scope of the local readout.

\subsection{Reading Actions from a Task Field}
\label{sec:method_readout}

\smallskip
\qheading{Continuous commands from the differential.}
The continuous policy follows the pulled-back gradient:
\begin{equation}
    \begin{aligned}[b]
        \pi(z,\ell) & =a_0+\eta_\phi(\bar z)\,g_a,                                                            \\
        \eta_\phi   & =\operatorname{softplus}\!\bigl(s_\phi(\operatorname{sg}\bar z)\bigr)+\varepsilon_\eta.
    \end{aligned}
    \label{eq:continuous_policy}
\end{equation}
A small MLP computes the positive mobility $\eta_\phi$ from the reference query with a stop-gradient on its input and no language input.
This positive scalar sets the magnitude of the displacement from $a_0$ but cannot rotate it.
The displacement therefore follows the field gradient and has a first-order change in $Q$ of $\eta_\phi\|g_a\|^2\ge0$.
The language instruction enters only through $p_\theta$.
This readout is a scalar-mobility instance of a generalized gradient field~\cite{mclachlan1998unified}.

\smallskip
\qheading{Hybrid actions from one potential.}
Consider a hybrid action $a=(u,c)$ consisting of a continuous pose increment $u\in\mathbb{R}^6$ and a discrete gripper command $c\in\{\mathrm{open},\mathrm{close}\}$.
The action-conditioned potential becomes $Q_{z,\ell}(u,c)=p_\theta(f(z,(u,c)),\ell)$.
At the reference $(u_0,c_{\mathrm h})$ we set $u_0=0$ and hold the current gripper command $c_{\mathrm h}$.
The pose follows the derivative:
\begin{equation}
    u=u_0+\eta_\phi(\bar z)\,\nabla_uQ_{z,\ell}(u_0,c_{\mathrm h}),
    \label{eq:pose_policy}
\end{equation}
the gripper decision uses a finite difference of the same function:
\begin{equation}
    \begin{aligned}[b]
        b & =Q_{z,\ell}(u_0,c_{\mathrm{alt}})-Q_{z,\ell}(u_0,c_{\mathrm h}),                                         \\
        c & =\begin{cases}c_{\mathrm{alt}},&b>\varepsilon_{\mathrm g},\\ c_{\mathrm h},&\text{otherwise,}\end{cases}
    \end{aligned}
    \label{eq:gripper_policy}
\end{equation}
where $c_{\mathrm{alt}}$ is the other gripper command.
The threshold $\varepsilon_{\mathrm g}\geq0$ sets the minimum difference required to switch.
Both readouts use the same reference and define a local hybrid controller rather than a joint maximizer over $(u,c)$.
Each decision requires one gradient evaluation and two potential evaluations without sampling or iterative search.

\subsection{Learning Control-Aligned Task Fields}
\label{sec:method_learning}

\smallskip
\qheading{Supervising the induced behavior.}
Let $\bar u_E$ be the normalized expert pose command.
The action loss compares the induced pose command with this expert command:
\begin{equation}
    \mathcal L_{\mathrm{act}}=\tfrac{\lambda_{\mathrm{vec}}}{6}\|u-\bar u_E\|^2+\lambda_{\mathrm{dir}}\bigl(1-\cos(u,\bar u_E)\bigr).
    \label{eq:action_supervision}
\end{equation}
We retain the computation graph of $\nabla_uQ$ and differentiate $\mathcal L_{\mathrm{act}}$ through the gradient readout by double backpropagation.
The resulting gradient reaches the potential through $\nabla_zp_\theta(\bar z,\ell)$ and reshapes the local field geometry at the query latent.
It reaches the world model and encoder through $J_f$ and $\bar z$ to reshape the latent action response of the query map.
The same loss trains the mobility that scales the continuous command.
Binary labels $y$ mark short windows ending at expert gripper switches.
A temperature-scaled logistic loss on $(2y-1)b$ trains the finite difference in \cref{eq:gripper_policy} and updates only $p_\theta$.

\smallskip
\qheading{Two kinds of supervision, one field.}
$\mathcal L_{\mathrm{act}}$ supervises the field \emph{differential} through the actions it induces.
$\mathcal L_p$ supervises its \emph{values} using timing targets available without additional annotation.
\cref{sec:exp_rq3} shows that the continuous readout remains learnable without $\mathcal L_p$.
Replacing $p_\theta$ by $c\,p_\theta+b$ and $\eta_\phi$ by $\eta_\phi/c$ for $c>0$ leaves the command in \cref{eq:continuous_policy} unchanged.
Continuous action supervision alone therefore does not fix the level or scale of the field.
Progress labels anchor values on observations while gripper supervision shapes query differences.

\smallskip
\qheading{Joint training.}
$E$ and $f$ are initialized from a world model pretrained on the demonstrations with latent prediction and the SIGReg regularizer~\cite{balestriero2025lejepa}.
During an initial warm-up phase we train $p_\theta$ with $\mathcal L_p$ while $E$ and $f$ remain frozen.
This stage produces the progress-only fields analyzed in \cref{sec:exp_rq2}.
We then jointly optimize the progress, action, gripper, and world model objectives:
\begin{equation}
    \mathcal L=\mathcal L_p+\lambda_{\mathrm{act}}\mathcal L_{\mathrm{act}}+\lambda_{\mathrm{grip}}\mathcal L_{\mathrm{grip}}+\lambda_{\mathrm{WM}}\mathcal L_{\mathrm{WM}},
    \label{eq:joint_objective}
\end{equation}
where $\mathcal L_{\mathrm{WM}}$ is the predictive objective of the world model on recorded transitions.
The world model uses a lower learning rate than the potential and mobility.

%% file: sec/fig/pipeline.tex
\begin{figure*}[t]
    \centering
    \includegraphics[width=\linewidth]{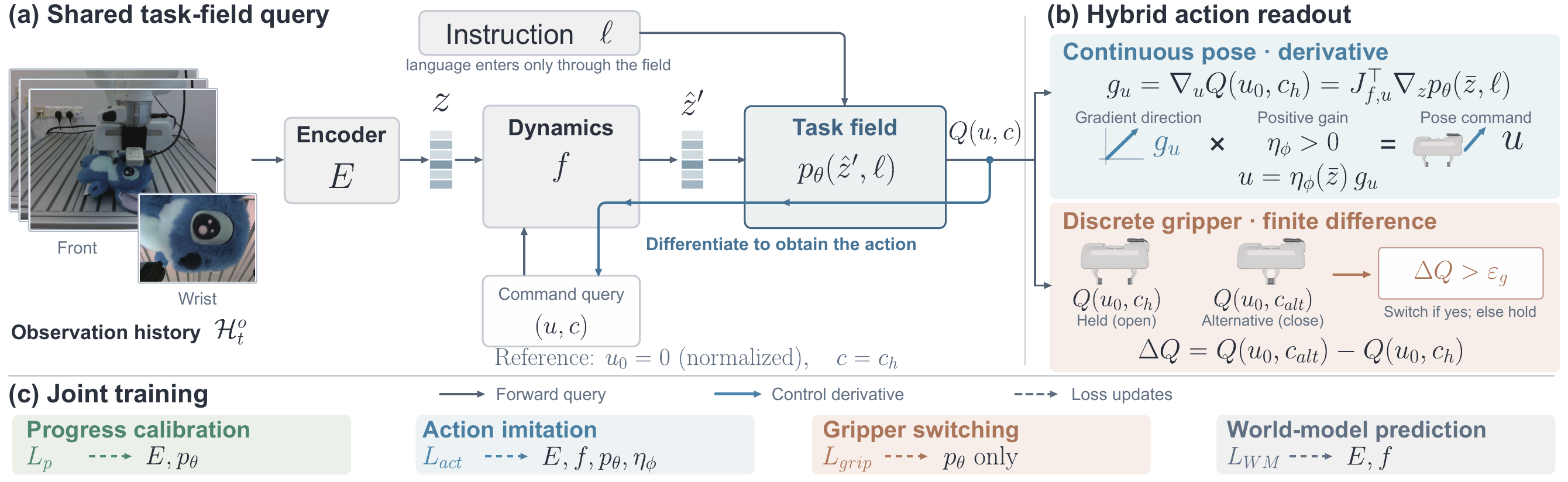}\vspace{-3mm}
    \caption{\textbf{Progress Predicts Policy.}
        \textbf{(a)} The language-independent control branch of the world model supplies latent queries that the language-conditioned task field scores through $Q(u,c)=p_\theta(f(z,(u,c)),\ell)$.
        \textbf{(b)} At the held-gripper reference, pose commands scale the pulled-back gradient by a learned positive gain while gripper switches use finite differences of the same $Q$.
        \textbf{(c)} Training combines progress supervision, pose imitation, gripper switching, and world model prediction.
        Pose imitation differentiates through the gradient readout while gripper supervision updates only $p_\theta$.
        \vspace{-4mm}
    }
    \label{fig:pipeline}
\end{figure*}

%% file: sec/4_experiments.tex
\section{Experiments}
\label{sec:experiments}

%
We evaluate \method on simulated and real robot manipulation, examining policy performance, the role of progress supervision, the locality of learned guidance, and hybrid action readouts.

\subsection{Experimental Setup}
\label{sec:experimental_setup}

\smallskip
\qheading{Shared architecture.}
\label{sec:experiment_implementation}
All methods share encoder, world model, and MLP-trunk architectures but use method-specific readouts and supervision.
A shared LeWM~\cite{maes2026leworldmodel} ViT encodes third-person and wrist images ($128\times128$) into one $384$-dimensional latent.
A transformer predicts the next latent from three past observations and their action chunks.
Each prediction spans $k=4$ raw ticks.
Both \method and BC regress the mean expert pose command and execute one tick before replanning.
LIBERO uses \texttt{all-MiniLM-L6-v2}~\cite{reimers2019sentencebert,wang2020minilm} instruction embeddings.
Real-robot tasks use one-hot vectors that remain constant in the single-task doll and bimanual experiments.
The potential and BC head are FiLM-conditioned residual MLPs~\cite{perez2018film} and the mobility in \method is a small MLP.

\smallskip
\qheading{Compared methods.}
\textbf{BC} predicts pose from the reference latent and uses a gripper-switching head with the same context and action loss as \method.
\textbf{BC+Progress} adds a progress output and the progress losses used by \method to the same trunk.
\textbf{BC-MH} predicts four pose candidates and trains progress to favor the candidate closest to the expert command.
We compare its learned selector with progress ranking at the same checkpoint~\cite{zhao2026tapsampling}.
\textbf{CEM} replaces the pose readout of \method with cross-entropy-method search~\cite{rubinstein1999cross} to maximize $Q$ within the training action range.
\textbf{CEM-Goal} uses goal-reaching planning in latent world models~\cite{zhou2025dinowm,maes2026leworldmodel}.
\textbf{\method-GH} keeps the pose readout of \method but uses the BC gripper-switching head.
\textbf{Direct-AP} uses a direct action-space scalar potential $Q(u)=\Phi(\bar z,u,\ell)$ and retains the world model reference prediction $\bar z=f(z,a_0)$ and gripper head of \method-GH.
\textbf{No $\mathcal L_p$} removes all progress losses and trains the field of \method-GH solely through its action readout.
\textbf{Seer-PPP} reads the task field through a small world model added to the action-free network of Seer~\cite{tian2025seer}.
Both Seer-PPP and Seer train from scratch with the Seer architecture and training recipe to test the field readout against a strong baseline.

\smallskip
\qheading{LIBERO protocol.}
The world model first trains alone for $100$ epochs on all $6{,}500$ demonstrations from $130$ LIBERO tasks, including LIBERO-90~\cite{liu2023libero}.
Each method then trains one multitask policy from this checkpoint on the same demonstrations and is compared at matched epochs.
Progress supervision uses the normalized-phase targets of \cref{sec:method_field}.
We evaluate Spatial, Object, Goal, and Long across $40$ tasks with $50$ initial states per task shared across methods.
All controllers replan end-effector pose increments and binary gripper commands at $20$\,Hz.
Success-rate differences use task-level sign-flip tests ($p_{\mathrm{task}}$).

\smallskip
\qheading{Gradient diagnostics.}
We measure progress MAE, ranking, and gradient--expert alignment on fixed training demonstration states with the deployed queries.
Closed-loop success on evaluation initial states separately measures control performance.
Progress-only fields train with the encoder and dynamics frozen.
The frozen-WM control uses the losses of \method-GH under the same freezing constraint.

\smallskip
\qheading{Real robot.}
A Franka arm uses wrist and front RGB cameras with the same input resolution, context, and control rate as LIBERO.
Both methods start from the same world model pretrained for $100$ epochs on the demonstrations for each task.
Bun picking requires placing the bun of the instructed color in a steamer.
This task uses normalized-phase supervision and $400$ demonstrations with evaluation on ten layouts under all four instructions.
Single-instruction doll straightening requires flipping and orienting a randomly posed doll.
This task uses absolute time-to-go supervision and $150$ demonstrations with $12$ evaluation trials per method.
\cref{sec:exp_bimanual} describes an additional bimanual task with two xArm7 arms.

\input{sec/fig/real_rollouts.tex}

\subsection{Task Fields as Policies}
\label{sec:exp_rq1}

\input{sec/table/libero_epoch120.tex}

\cref{tab:libero_epoch120} shows that \method improves on BC in every LIBERO suite at epoch $120$ with the largest gains on Object and Long.
These comparisons omit the gripper-settling procedure described in \cref{sec:exp_rq5}.
The stronger BC-branch narrows the gap but \method retains higher average success without settling under both schedules.
The supplementary material reports these comparisons and strong performance with a SEER~\cite{tian2025seer} backbone.
\cref{tab:real_bun_results} shows that \method reduces every failure type relative to BC on bun picking.
Wrong-target picks remain the most common failure for both methods.
Doll straightening requires no instruction grounding and tests control under a single instruction.
\cref{tab:real_stitch_results} shows that \method achieves two more successes in $12$ trials per method and completes successful trials faster on average.
\cref{fig:real_rollouts} shows successful rollouts and field predictions.

\input{sec/table/real_bun_results.tex}
\input{sec/table/real_stitch_results.tex}

\subsection{Predicting Progress Is Not Enough}
\label{sec:exp_rq2}

\input{sec/table/progress_vs_control.tex}

Success in \cref{tab:progress_vs_control} is pooled over epochs $70$--$120$ unless otherwise specified.
BC+Progress matches \method's progress fit but has no significant success gain over BC.

\input{sec/fig/grad_alignment.tex}

\smallskip
\qheading{Value accuracy versus action alignment.}
\cref{fig:grad_alignment} shows that progress-only fields rank demonstration states almost perfectly but have action gradients that align weakly with expert commands.
Longer progress training improves value fit without improving alignment and the progress gradients of BC+Progress drift toward orthogonality.
Action imitation yields near-perfect training-state alignment for \method and \method-GH across all $40$ tasks without degrading progress fit.
A positive gain cannot correct a misaligned direction.
\cref{tab:progress_vs_control} shows that the progress-only field with a trained gain barely moves and succeeds in only $1$ of $2{,}000$ episodes.

\smallskip
\qheading{Potential or dynamics?}
\cref{fig:grad_alignment} shows that action supervision can align gradients even with the encoder and dynamics frozen.
At epoch $120$ the mean cosine reaches $0.95$ compared with $0.22$ for progress supervision alone.
However, \cref{tab:progress_vs_control} reports only $5.0\%$ success for this frozen-WM control compared with $43.5\%$ for \method-GH.
Training-state alignment alone therefore does not ensure closed-loop control.
Exact dynamics also does not resolve the gap: ascending the learned field through the simulator succeeds in $1.5\%$ of episodes compared with $66.0\%$ for the command produced by \method.
That command raises the field at a similar rate to the expert direction ($72\%$ versus $74\%$).

\smallskip
\qheading{Direct or factorized action conditioning.}
Direct-AP reaches $41.6\%$ pooled success compared with $43.5\%$ for \method-GH with no significant difference detected ($p_{\mathrm{task}}=0.351$).
This result shows that a directly action-conditioned scalar also supports an effective continuous gradient readout in this setting.
Direct-AP retains the reference representation from the world model and adds action-value supervision.
This comparison does not establish a performance advantage for explicit composition.

\smallskip
\qheading{Instruction conditioning.}
At episode start we test whether the command under each instruction points closest to the demonstrated motion among tasks that share a scene.
\method matches the instructed task about twice as often as chance on the real robot and three times as often on LIBERO-Goal.
These results are comparable to BC readouts and measure instruction sensitivity of the initial command rather than generalization to unseen instructions.
Progress-only gradients remain at chance on LIBERO-Goal.

\subsection{What Progress Labels Add}
\label{sec:exp_rq3}

\input{sec/fig/rollout_progress.tex}

\smallskip
\qheading{Acting without progress labels.}
\cref{tab:progress_vs_control} shows that removing all progress losses from \method-GH leaves an effective policy (No $\mathcal L_p$) with comparable expert alignment and local action-space structure.
This comparison also changes when action learning begins.
Matched branch runs account for this difference and show no consistent control benefit from progress supervision.
Despite comparable action alignment, the scalar trained without progress labels correlates less strongly with demonstration progress.
Progress labels are unnecessary for continuous control in this setting but provide a clearer demonstration-relative ordering.

\smallskip
\qheading{What calibration adds.}
\cref{fig:rollout_progress} shows the field predicted by \method on its $2{,}000$ closed-loop rollouts.
On successful rollouts the predictions track elapsed fraction with a median per-rollout Spearman correlation of $1.00$ and MAE of $0.08$. They also end near the terminal label with a median phase of $0.97$ while failures tend to end lower.
Without tuning, final and $5$\,s values yield success AUROCs of $0.84$ and $0.78$ compared with $0.73$ and $0.56$ without progress labels. Each model is evaluated on its rollout distribution.
\cref{fig:rollout_progress}c shows the label-free field tracks elapsed fraction less closely even after an affine fit on demonstrations. These results support demonstration-relative progress monitoring but the AUROC comparison reflects differences in rollout distributions. The monitor remains imperfect: $30\%$ of failures reach within a tenth of the terminal label during execution.

\subsection{Locality of the Learned Field}
\label{sec:exp_rq4}

\input{sec/fig/local_linearity.tex}

\cref{fig:local_linearity} measures first-order fit and gradient stability around the reference command.
For \method and \method-GH the first-order approximation holds only within a small fraction of a typical expert command.
Gradient directions change sharply farther from the reference.
Despite this local scope, $Q$ still ranks equal-size commands toward the expert direction.
For \method-GH, this holds on $98\%$ of states when commands have the same norm as the expert command.
Such directional ranking does not establish calibrated progress forecasts for arbitrary finite commands.
\cref{tab:progress_vs_control} further shows that progress-based ranking does not significantly improve BC-MH and that CEM on the \method checkpoint fails in all $2{,}000$ rollouts.
These findings are limited to the tested search procedures and action ranges.

\subsection{One Field, Two Action Types}
\label{sec:exp_rq5}

\smallskip
\qheading{Gripper readout and settling.}
\label{sec:exp_gripper_settling}
\cref{tab:progress_vs_control} shows that finite-difference switching on $Q$ improves average success over the gripper head of \method-GH with the same pose-readout architecture ($p_{\mathrm{task}}<0.01$).
\cref{tab:libero_epoch120} reports gains on Object and Long but lower success on Spatial and Goal.
We also test a settling procedure to reduce premature post-grasp motion.
Settling keeps pose queries conditioned on an open gripper until finger feedback indicates that closure has stopped.
This procedure modestly improves epoch-$120$ success without retraining or changing switching decisions.
The comparisons above omit settling.

\smallskip
\qheading{Multiple discrete readouts.}
\label{sec:exp_bimanual}
In egg pouring with two xArm7 arms, one field supplies a joint gradient over $12$ pose coordinates and a finite difference for each gripper.
A single mobility scales the joint pose gradient.
The controller succeeds in $1$ of $5$ trials as shown in \cref{fig:real_rollouts}(c).

%% file: sec/fig/real_rollouts.tex
\begin{figure*}[t]
    \centering
    \includegraphics[width=\linewidth]{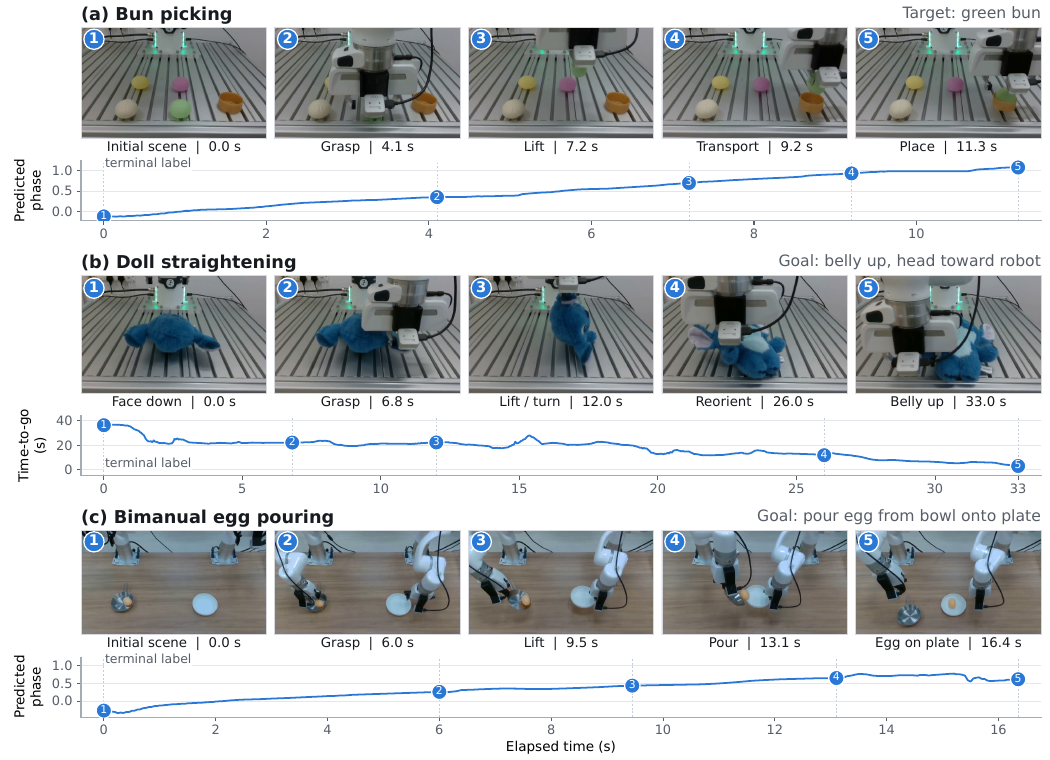}
    \caption{\textbf{Real-robot execution.}
        Successful \method rollouts for (a) green-bun picking, (b) doll straightening, and (c) bimanual egg pouring.
        Numbered frames correspond to samples on the curves, which show predicted demonstration phase in (a,c) and absolute time-to-go in seconds in (b).
        Predictions are unsmoothed and unclipped.
        Dashed lines indicate terminal training labels rather than measured completion.\vspace{-3mm}
    }
    \label{fig:real_rollouts}
\end{figure*}

%% file: sec/table/libero_epoch120.tex

\begin{table}[!ht]
    \centering
\caption{LIBERO success rates (\%) at epoch $120$ under the original protocol with $500$ rollouts per suite.
PPP and PPP w/o settling differ only in deployment-time gripper settling (\cref{sec:exp_gripper_settling}).
Bottom rows use the Seer backbone at epoch $50$ under its training recipe.} \vspace{-1mm}
    \label{tab:libero_epoch120}
    \small
    \renewcommand{\arraystretch}{0.78}
    \setlength{\tabcolsep}{3pt}
    \setlength{\aboverulesep}{1pt}
    \setlength{\belowrulesep}{1pt}
    \begin{tabular*}{\columnwidth}{@{\extracolsep{\fill}}lrrrrr@{}}
        \toprule
        \addlinespace[1pt]
        Method           & Spatial       & Object        & Goal          & Long          & Avg.          \\[1pt]
        \midrule
        BC               & 31.4          & 53.6          & 47.0          & 20.2          & 38.1          \\
        BC+Progress      & 33.8          & 52.2          & 49.0          & 20.0          & 38.8          \\
        Direct-AP        & 45.0          & 50.8          & 53.0          & 30.4          & 44.8          \\
        PPP-GH           & \textbf{46.8} & 64.6          & 53.0          & 27.4          & 48.0          \\
        PPP w/o settling & 40.4          & \textbf{76.2} & 52.4          & 37.0          & 51.5          \\
        PPP              & 46.0          & 75.4          & \textbf{53.6} & \textbf{42.2} & \textbf{54.3} \\
        \midrule
        Seer             & 98.8          & 99.0          & 97.8          & 93.6          & 97.3          \\
        Seer-PPP         & 99.4          & 99.8          & 98.2          & 91.0          & 97.1          \\
        \bottomrule
    \end{tabular*}
\end{table}

%% file: sec/table/real_bun_results.tex

\begin{table}[!ht]
    \centering
    \caption{Real-robot bun picking. Failure columns count the
        primary observed failure mode and sum to the number of failed trials.} \vspace{-1mm}
    \label{tab:real_bun_results}
    \small
    \renewcommand{\arraystretch}{0.78}
    \setlength{\tabcolsep}{3pt}
    \setlength{\aboverulesep}{1pt}
    \setlength{\belowrulesep}{1pt}
    \begin{tabular*}{\columnwidth}{@{\extracolsep{\fill}}lccccc@{}}
        \toprule
        \addlinespace[1pt]
        Method & Success
        & \makecell{Success\\[-2pt]rate}
        & \makecell{Wrong\\[-2pt]target}
        & \makecell{Pose\\[-2pt]inaccuracy}
        & \makecell{Gripper\\[-2pt]inaccuracy} \\[1pt]
        \midrule
        PPP & \textbf{21/40} & \textbf{52.5\%} & 10 & 8  & 1 \\
        BC  & 5/40          & 12.5\%          & 13 & 13 & 9 \\
        \bottomrule
    \end{tabular*}
\end{table}

%% file: sec/table/real_stitch_results.tex

\begin{table}[!ht]
    \centering
    \caption{Real-robot doll straightening. Time: mean time to success over successful trials.} \vspace{-1mm}
    \label{tab:real_stitch_results}
    \small
    \renewcommand{\arraystretch}{0.78}
    \setlength{\tabcolsep}{3pt}
    \setlength{\aboverulesep}{1pt}
    \setlength{\belowrulesep}{1pt}
    \begin{tabular*}{\columnwidth}{@{\extracolsep{\fill}}lccc@{}}
        \toprule
        \addlinespace[1pt]
        Method & Success & Success rate & Time (s) \\[1pt]
        \midrule
        PPP & \textbf{9/12} & \textbf{75.0\%} & \textbf{31.8} \\
        BC  & 7/12          & 58.3\%          & 46.0 \\
        \bottomrule
    \end{tabular*}
    \vspace{-3mm}
\end{table}

%% file: sec/table/progress_vs_control.tex
\begin{table}[!t]
    \centering
    \caption{Progress fit and control without settling.
    At epoch $120$, $\rho$ measures task-averaged progress correlation ($130$ tasks) and cos measures gradient--expert alignment ($40$ tasks).
    Succ.\ is closed-loop success (\%) pooled over epochs $70$--$120$, except progress, gain and the last two blocks (epoch $120$).
    No $\mathcal L_p$ omits progress supervision.
    $\Delta$ is the paired gain in percentage points over the preceding row at matched epochs.
    References are \method-GH for Direct-AP, No $\mathcal L_p$, and frozen-WM \method-GH, and gradient for CEM/CEM-Goal.
    $p_{\mathrm{task}}$ uses task-level sign flips.
    True-dyn.\ ascent differentiates through the simulator and compares with \method in the same harness ($66.0\%$, supplementary material).}
    \label{tab:progress_vs_control}
    \footnotesize
    \renewcommand{\arraystretch}{0.78}
    \setlength{\tabcolsep}{2.4pt}
    \setlength{\aboverulesep}{1pt}
    \setlength{\belowrulesep}{1pt}
    \begin{tabular*}{\columnwidth}{@{\extracolsep{\fill}}llccrrr@{}}
        \toprule
        \addlinespace[1pt]
        Method                 & From $Q$  & $\rho$ & cos   & Succ. & $\Delta$ & $p_{\mathrm{task}}$ \\[1pt]
        \midrule
        BC                     & --        & --     & --    & 36.2  & --       & --                  \\
        BC+Progress            & --        & 0.9973 & 0.08  & 36.5  & $+0.3$   & 0.86                \\
        \method-GH             & pose      & 0.9974 & 0.995 & 43.5  & $+6.9$   & 0.0002              \\
        \method w/o settling   & both      & 0.9970 & 0.995 & 50.5  & $+7.1$   & 0.0075              \\
        Direct-AP              & pose      & 0.9975 & --    & 41.6  & $-1.9$   & 0.351               \\
        \midrule
        \multicolumn{7}{@{}l@{}}{\method-GH without progress labels:} \\
        \quad No $\mathcal L_p$ & pose      & 0.6863 & 0.996 & 38.6  & $-4.9$   & 0.011               \\
        \midrule
        \multicolumn{7}{@{}l@{}}{World model frozen:} \\
        \quad \method-GH       & pose      & 0.9880 & 0.947 & 5.0   & $-38.5$  & $<\!10^{-5}$        \\
        \quad progress, gain   & pose      & 0.9929 & 0.225 & 0.05  & $-4.4$   & 0.0001              \\
        \midrule
        \multicolumn{7}{@{}l@{}}{BC-MH, same checkpoints:} \\
        \quad own pick         & --        & 0.9923 & --    & 31.2  & --       & --                  \\
        \quad progress pick    & selection & 0.9923 & --    & 30.7  & $-0.4$   & 0.66                \\
        \midrule
        \multicolumn{7}{@{}l@{}}{\method w/o settling, epoch $120$:} \\
        \quad gradient         & both      & 0.9970 & 0.995 & 51.5  & --       & --                  \\
        \quad true-dyn.\ ascent & gripper   & 0.9970 & 0.26  & 1.5   & $-64.5$  & $<\!10^{-5}$        \\
        \quad CEM              & search    & 0.9970 & --    & 0.0   & $-51.5$  & $<\!10^{-5}$        \\
        \midrule
        \multicolumn{7}{@{}l@{}}{World model only:} \\
        \quad CEM-Goal         & --        & --     & --    & 6.2   & $-45.4$  & $<\!10^{-5}$        \\
        \bottomrule
    \end{tabular*}
    \vspace{-3mm}
\end{table}

%% file: sec/fig/grad_alignment.tex
\begin{figure*}[t]
    \centering
    \captionsetup{skip=4pt}
    \includegraphics[width=\linewidth]{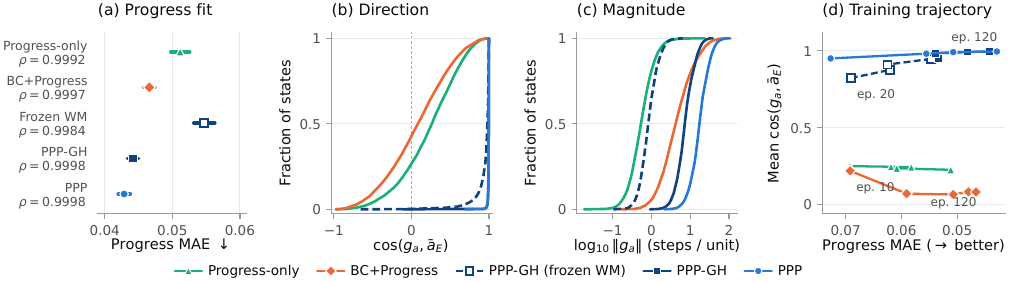}
    \caption{\textbf{Accurate progress values do not ensure aligned action gradients.}
        All panels measure fit using deployment queries on the same LIBERO training states.
        Progress-only uses only progress losses.
        Frozen WM trains \method-GH with the encoder and dynamics frozen.
        BC+Progress evaluates the progress output of the BC trunk.
        At epoch $120$, (a) shows progress MAE and per-trajectory Spearman $\rho$, and (b,c) show cumulative distributions of gradient--expert cosine similarity and gradient norm.
        (d) tracks fit and alignment across checkpoints.} \vspace{-3mm}
    \label{fig:grad_alignment}
\end{figure*}

%% file: sec/fig/rollout_progress.tex
\begin{figure*}[t]
    \centering
    \captionsetup{skip=4pt}
    \includegraphics[width=\linewidth]{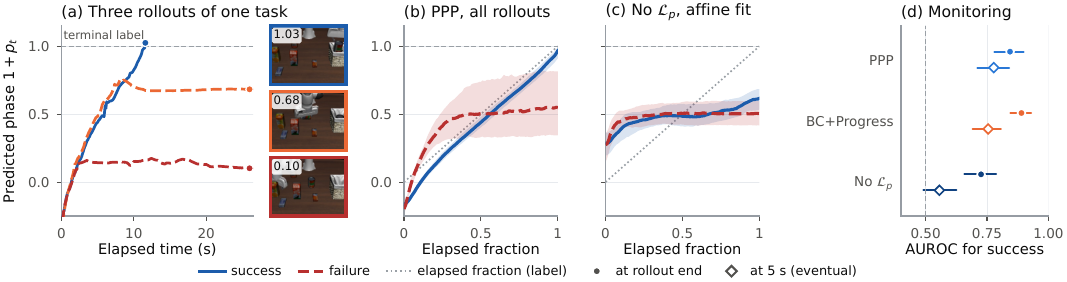}
    \caption{\textbf{Progress predictions during closed-loop rollouts.}
        (a) Predicted phase $1+p_t$ for three \method rollouts of one LIBERO-Long task with final frames and phases.
        (b,c) Median predictions and interquartile bands over normalized elapsed time for $2{,}000$ rollouts each of \method and No $\mathcal L_p$.
        The latter uses affine calibration on demonstration states.
        Dotted lines indicate normalized elapsed time.
        (d) Success AUROC at rollout end (filled) and at $5$\,s for ongoing rollouts (hollow) with $95\%$ task-bootstrap intervals.
        Results use epoch $120$ without settling.}\vspace{-3mm}
    \label{fig:rollout_progress}
\end{figure*}

%% file: sec/fig/local_linearity.tex
\begin{figure}[t]
    \centering
    \captionsetup{skip=4pt}
    \includegraphics[width=\linewidth]{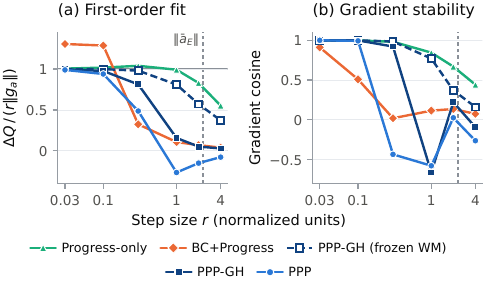}
    \caption{\textbf{Local scope of learned field guidance.}
        (a) Median ratio of $Q(a_0+r\hat g_a)-Q(a_0)$ to its first-order prediction $r\|g_a\|$ across diagnostic states.
        Step size $r$ is measured in normalized command units.
        (b) Cosine similarity between gradients at $a_0$ and $a_0+r\hat u_E$.
        Dashed lines mark the median expert command norm ($2.3$).} \vspace{-3mm}
    \label{fig:local_linearity}
\end{figure}

%% file: sec/5_conclusion.tex
\section{Conclusion}
\label{sec:conclusion}

\method learns a scalar task field through the commands it induces: continuous commands from action gradients and gripper switches from finite differences.
Composing the field with a shared world model supplies action conditioning and an explicit chain-rule interpretation; direct action conditioning is also effective.
Progress labels are optional for learning continuous control, but calibrate the field for execution monitoring.
Under a shared architecture, \method outperforms BC in simulation and on real robots.

\smallskip
\qheading{Limitations.}
\label{sec:limit}
Demonstration-relative progress may admit ``arm clock'' shortcuts based on arm motion rather than task-relevant scene changes.
The field has no Lyapunov guarantee, and its differential does not guarantee physical progress.
Progress-based recovery and sharing the task field across embodiments remain future work.

%% file: sec/X_supplementary.tex
\clearpage
\twocolumn[{\raggedright\LARGE\bfseries Appendix\par\vspace{1em}}]
\section{Implementation Details}
\label{sec:ppp_implementation}

\subsection{General Mobility and Local Improvement}
\label{sec:ppp_general_mobility}

\smallskip
\qheading{Latent geometry and the exact pullback.}
Fix the observation history, instruction, and query-normalization statistics, and let $Q(a)=p_\theta(f(z,a),\ell)$.
Write $\bar z=f(z,a_0)$ and $J_0=\partial f(z,a)/\partial a|_{a_0}$.
Suppose $f$ and $p_\theta$ are twice continuously differentiable near the reference query and its latent output.
For a small action perturbation, the query's latent displacement is
\begin{equation}
    \begin{aligned}
        \Delta z & =f(z,a_0+\delta a)-\bar z       \\
                 & =J_0\delta a+O(\|\delta a\|^2).
    \end{aligned}
    \label{eq:ppp_latent_displacement}
\end{equation}
Taylor expansion of the potential at $\bar z$ gives
\begin{equation}
    \begin{aligned}
        Q(a_0+\delta a)-Q(a_0)
              & =\nabla_zp_\theta(\bar z,\ell)^\top\Delta z+R_p, \\
        |R_p| & \leq\frac{L_p}{2}\|\Delta z\|^2,
    \end{aligned}
    \label{eq:ppp_latent_taylor}
\end{equation}
when $\|\nabla_z^2p_\theta\|_{\mathrm{op}}\leq L_p$ along the latent segment.
Substituting \cref{eq:ppp_latent_displacement} yields
\begin{equation}
    \begin{gathered}
        g_a=\nabla_aQ(a_0)=J_0^\top\nabla_zp_\theta(\bar z,\ell),\\
        Q(a_0+\delta a)-Q(a_0)=g_a^\top\delta a+O(\|\delta a\|^2).
    \end{gathered}
    \label{eq:ppp_pullback_taylor}
\end{equation}
The chain-rule identity for $g_a$ is exact; the Taylor approximation concerns finite changes in $Q$.
The expansion point is the reference query $\bar z$, which need not equal the current encoded state $z$.
Subtracting $p_\theta(z,\ell)$ instead of $Q(a_0)$ adds the action-independent offset $Q(a_0)-p_\theta(z,\ell)$ and leaves the derivative unchanged.

\smallskip
\qheading{A metric-dependent readout.}
The pulled-back differential $J_f^\top\mathrm{d}p_\theta$ is a covector on the action space.
Turning it into a command requires an action-space metric: under a positive-definite $G$, the steepest-ascent direction is $G^{-1}g_a$.
More generally, any operator whose symmetric part is positive semidefinite gives a nonnegative first-order term.
The generalized gradient field~\cite{mclachlan1998unified} is
\begin{equation}
    \begin{aligned}
        \delta a    & =\bigl(M_\phi(\bar z)+W_\phi(\bar z)\bigr)g_a, \\
        M_\phi^\top & =M_\phi\succeq0,\qquad W_\phi^\top=-W_\phi,
    \end{aligned}
    \label{eq:generalized_gradient_policy}
\end{equation}
whose first-order effect is $g_a^\top\delta a=g_a^\top M_\phi g_a\geq0$, since $g_a^\top W_\phi g_a=0$.
$M_\phi$ rescales the gradient and can redirect it, and $W_\phi$ adds motion tangent to the local level set of $Q$.
Both depend only on the reference query latent.
They could be parameterized as $M_\phi=C_\phi C_\phi^\top+\varepsilon I$ and $W_\phi=\tfrac12(K_\phi-K_\phi^\top)$.
All experiments use the scalar instance $M_\phi=\eta_\phi I$, $W_\phi=0$ of \cref{eq:continuous_policy}.
Its command direction is exactly that of $g_a$, and only its magnitude is learned.

\smallskip
\phantomsection
\qheading{Local justification of the control direction.}
\label{sec:ppp_local_control_proof}
Suppose $\|\nabla_a^2Q(a)\|_{\mathrm{op}}\leq L$ in a neighborhood of $a_0$, for $L\geq0$.
Hold the mobility operators at their reference-state values and let
\begin{equation}
    v=(M_\phi+W_\phi)g_a,
    \qquad \beta=g_a^\top M_\phi g_a.
    \label{eq:ppp_local_direction}
\end{equation}
For $a_\alpha=a_0+\alpha v$ whose connecting segment stays in this neighborhood, Taylor's theorem gives
\begin{equation}
    Q(a_\alpha)-Q(a_0)\geq\alpha\beta-\frac{L}{2}\alpha^2\|v\|^2 .
    \label{eq:ppp_finite_step_bound}
\end{equation}
Hence, if $\beta>0$ and $L>0$, every step with
\begin{equation}
    0<\alpha<\frac{2\beta}{L\|v\|^2}
    \label{eq:ppp_sufficient_step_size}
\end{equation}
that stays in the neighborhood strictly improves $Q$ relative to the reference command.
If $L=0$ and $\beta>0$, every positive step that stays in the neighborhood improves $Q$.
If $\beta=0$, the first-order term gives no strict-improvement guarantee; a nonzero skew component can still produce motion in this case.
For the scalar instance $v=\eta_\phi g_a$,
\begin{equation}
    Q(a_0+\alpha\eta_\phi g_a)-Q(a_0)\geq\alpha\eta_\phi\Bigl(1-\frac{L\alpha\eta_\phi}{2}\Bigr)\|g_a\|^2.
    \label{eq:ppp_scalar_step_bound}
\end{equation}
With $\eta_\phi>0$ and $g_a\neq0$, a sufficient condition is $0<\alpha\eta_\phi<2/L$ when $L>0$, together with the neighborhood condition; when $g_a=0$, this readout produces no perturbation.
The deployed controller uses $\alpha=1$ and a fitted gain without estimating $L$ or enforcing this step-size condition.
The bound therefore does not certify improvement of $Q$ for the executed command, and improvement of the learned query would itself require separate validation to imply physical task progress.
\Cref{fig:local_linearity} shows that in the trained models the first-order region is far smaller than a deployed command.

\smallskip
\qheading{Scope of the conservative-field statement.}
At fixed $z$ and $\ell$, the exact action-gradient field is conservative,
\begin{equation}
    \oint_{\mathcal C}\nabla_aQ(a)^\top\,\mathrm{d}a=0,
    \label{eq:ppp_conservative_action_gradient}
\end{equation}
for any closed curve $\mathcal C$ in its smooth domain.
This property holds in candidate-action space at a fixed query; successive control steps use different observations and histories, and hence different functions $Q$.
Multiplying by the scalar $\eta_\phi(\bar z)$, which is fixed for a given query, preserves this property, whereas $W_\phi\neq0$ need not.

All the local bounds on $Q$ and the conservative-field identity also apply to a twice continuously differentiable direct action-conditioned scalar, such as Direct-AP's $\Phi(\bar z,a,\ell)$ with $\bar z$ held fixed.
An explicit $p_\theta\circ f$ factorization is therefore not required for this gradient readout or its local ascent interpretation.
A direct scalar can learn the combined action dependence implicitly; these statements do not establish that it internally recovers dynamics and a task potential.

\subsection{Observations, Actions, and World-Model Queries}
\label{sec:ppp_query_details}

The visual and language encoders and the dynamics architecture are described
in Sec.~\ref{sec:experiment_implementation}. We suppress the language
encoder in the notation $p(z,\ell)$ and make the temporal conditioning of
the world model queries explicit below.

\smallskip
\qheading{Action coordinates.}
The six-dimensional continuous command contains a translation delta and an
axis-angle rotation delta. It is normalized elementwise as
\begin{equation}
    u=(u^{\mathrm{raw}}-\mu_u)\oslash\sigma_u,
    \label{eq:ppp_action_normalization}
\end{equation}
where $\mu_u$ is the corpus median and $\sigma_u$ is the corpus standard
deviation. Thus, the reference $u_0=0$ corresponds to the median raw command.
The positive scalar gain in the policy acts in these normalized coordinates.
For execution, we compute $\mu_u+\sigma_u\odot u$ and clip the result to the
robot's action limits.

\smallskip
\qheading{Temporal conditioning.}
Let $\tau$ index raw ticks and define
$Z_\tau=(z_{\tau-(H-1)k},\ldots,z_\tau)$. The past action chunks connect
successive observations in this context. A candidate $(u,c)$ is repeated
$k$ times to form $\operatorname{tile}_k(u,c)\in\mathbb{R}^{7k}$. Making the
conditioning suppressed in the main text explicit, a control query uses
\begin{equation}
    f(z,(u,c))
    :=f\!\left(Z_\tau,
    [\bar A_\tau,\operatorname{tile}_k(u,c)]\right),
    \label{eq:ppp_explicit_query}
\end{equation}
where $\bar A_\tau$ contains the past chunks with their pose channels
replaced by normalized zero and their gripper channels retained. This
history convention limits direct access to past pose commands when computing
the policy. It is used for the continuous gradient and both gripper queries.
The reference gripper $c^{\mathrm h}$ is the currently held command; the
reference pose contains no expert future command.

The held-action Jacobian accounts for every repeated slot:
\begin{equation}
    J_0
    =\left.\sum_{r=1}^{k}
    \frac{\partial f}{\partial u^{(r)}}
    \right|_{u^{(1)}=\cdots=u^{(k)}=u_0,\,c=c^{\mathrm h}}.
    \label{eq:ppp_tiled_jacobian}
\end{equation}
Autodifferentiation through the tiling operation computes this sum
automatically. The action gradient is obtained with one reverse-mode pass,
without materializing the Jacobian.

\smallskip
\qheading{World-model training.}
The predictive training objective uses the actual recorded past and future
action chunks. With $A_\tau^{\mathrm{past}}$ denoting the recorded past
chunks and $A_\tau$ the next recorded chunk, it predicts
\begin{equation}
    \hat z_{\tau+k}
    =f\!\left(Z_\tau,[A_\tau^{\mathrm{past}},A_\tau]\right).
    \label{eq:ppp_training_prediction}
\end{equation}
The world model loss combines dimension-normalized latent prediction error
with the SIGReg regularizer introduced in
Sec.~\ref{sec:method_learning}:
\begin{equation}
    \mathcal L_{\mathrm{WM}}
    =\mathbb{E}\!\left[\frac{1}{d}
        \|\hat z_{\tau+k}-z_{\tau+k}\|^2\right]
    +\lambda_{\mathrm{SIG}}\operatorname{SIGReg}(\{z\}).
    \label{eq:ppp_world_model_loss}
\end{equation}
We set $\lambda_{\mathrm{SIG}}=0.09$. The dynamics predictor takes no
language input. The same predictive loss is retained during joint policy
training.

\subsection{Progress Calibration}
\label{sec:ppp_progress_details}

For normalized-phase supervision, consider an episode with $T_e$ raw frames and let $N_e=\lfloor T_e/k\rfloor$
be its model-grid length and $t=\tau/k$ the time in model-transition
intervals at zero-based raw-frame index $\tau$. We define
\begin{equation}
    \begin{aligned}
        \rho_t    & =\min\!\left(\frac{t}{N_e-1},1\right), \\
        p_t^\star & =\rho_t-1,
        \qquad s_e=\frac{1}{N_e-1}.
    \end{aligned}
    \label{eq:ppp_progress_labels}
\end{equation}
Here $N_e-1$ is the normalization span of the model grid; raw-tick anchors
can have fractional $t$. Clipping keeps progress at one for final raw
frames at or beyond the last model-grid timestamp. The potential targets
lie in $[-1,0]$ and calibrate normalized temporal progress, with zero at
the terminal plateau. The quantity $s_e$ is the progress increment of one
model transition before clipping.

Anchors are sampled at raw-tick resolution. A later observation is sampled
at a valid gap $\Delta\in\{1,2,4,8,16,30\}$ model transitions, and a
mismatched instruction $\tilde\ell$ is drawn from another task. Using the
main text's pair notation $i<j$, define
$\tilde p_j=p(E(o_j),\tilde\ell)$. In addition to the value and increment
losses in Eq.~\eqref{eq:progress_supervision}, we use
\begin{align}
    \mathcal L_{\mathrm{rank}}
     & =\operatorname{softplus}\!\left(-\frac{p_j-p_i}{\tau_r}\right),
    \label{eq:ppp_rank_loss}                                           \\
    \mathcal L_{\mathrm{bd}}
     & =\mathbb{E}_{x:\rho_x\geq\rho_{\mathrm{term}}}[p_x^2]
    +\mathbb{E}_{x:\rho_x<\rho_{\mathrm{term}}}
     [\,[p_x+\varepsilon_{\mathrm{nt}}]_+^2\,],
    \label{eq:ppp_boundary_loss}                                       \\
    \mathcal L_{\mathrm{lang}}
     & =\rho_j[\,m_\ell-(p_j-\tilde p_j)\,]_+,
    \label{eq:ppp_language_loss}
\end{align}
where $[v]_+=\max(v,0)$ and $x$ ranges over the sampled observations.
The full progress objective is
\begin{equation}
    \begin{aligned}
        \mathcal L_p={} &
        \lambda_{\mathrm{val}}\mathcal L_{\mathrm{val}}
        +\lambda_{\mathrm{inc}}\mathcal L_{\mathrm{inc}}
        +\lambda_{\mathrm{rank}}\mathcal L_{\mathrm{rank}}               \\
                        & +\lambda_{\mathrm{bd}}\mathcal L_{\mathrm{bd}}
        +\lambda_{\mathrm{lang}}\mathcal L_{\mathrm{lang}}.
    \end{aligned}
    \label{eq:ppp_full_progress_loss}
\end{equation}
The ranking term encourages increasing potential along demonstrations, the boundary
term encourages a near-terminal plateau around zero, and the language term
encourages higher scores for the matched instruction, weighted by temporal
progress. The boundary target is a separate regularizer and need not
coincide with the timestamp target at every near-terminal frame.

\subsection{Continuous and Discrete Policy Supervision}
\label{sec:ppp_policy_supervision}

\smallskip
\qheading{Continuous target and gain.}
At raw-tick anchor $\tau$, the pose target is
\begin{equation}
    \bar u_{E,\tau}=\frac{1}{k}
    \sum_{r=0}^{k-1}u_{E,\tau+r}.
    \label{eq:ppp_expert_target}
\end{equation}
For a command held over the query's $k$ ticks (\cref{eq:ppp_explicit_query}),
this mean is the least-squares target $\bar u_E$ in \cref{eq:action_supervision}:
$\bar u_{E,\tau}=\arg\min_u\sum_{r=0}^{k-1}\|u-u_{E,\tau+r}\|^2$.
On LIBERO the end-effector displacement grows
linearly with the gap up to five ticks, so the motion within a chunk is close
to straight, and the chunk-constant part carries $97.9\%$ of the variance of
the expert's pose commands within a chunk at $k=4$ ($93.3\%$ at $k=8$).
All LIBERO and real-robot experiments, including BC, use this $k=4$ target.
Deployment executes the predicted chunk mean for one raw tick before
recomputing both readouts (\cref{sec:ppp_optimization_details}).
The gain network described in Sec.~\ref{sec:experiment_implementation}
uses $\varepsilon_\eta=10^{-3}$ and is initialized so that
$\eta_\phi\approx1$.
Its predicted-state input is detached. The default configuration uses
$M_\phi=\eta_\phi I$ and $W_\phi=0$, so the action gradient determines
the command direction and the mobility learns its scale.

\smallskip
\qheading{Gripper loss.}
With the switching benefit $b$ of \cref{eq:gripper_policy} and the switching-region label $y$ defined below, the per-sample loss is
\begin{equation}
    \ell_{\mathrm{grip}}=\tau_e\operatorname{softplus}\!\Bigl(-\frac{(2y-1)\,b}{\tau_e}\Bigr),
    \qquad \tau_e=\bar\tau s_e,
    \label{eq:gripper_supervision}
\end{equation}
where $s_e$ is the progress increment of one model transition in episode $e$ (\cref{eq:ppp_progress_labels}).
Predicted latents are detached for this term, so it updates only $p_\theta$.

\smallskip
\qheading{Switching regions and repeated-frame contexts.}
Let $\tau^{\mathrm{sw}}$ be the next expert switch, and let
$c^{\mathrm{sw}}$ be its destination command. The switching label is
\begin{equation}
    y_\tau=\mathbf{1}\!\left[
        0\leq\tau^{\mathrm{sw}}-\tau\leq T_{c^{\mathrm{sw}}}\right],
    \label{eq:ppp_switch_region}
\end{equation}
with $T_{\mathrm{close}}=2$ and $T_{\mathrm{open}}=4$ raw ticks; anchors
without a future switch receive $y_\tau=0$.
Each anchor also supplies a repeated-frame context:
the current observation fills all $H$ context slots, with a hold-position
action history and the original label. Both original and repeated-frame
contexts contribute to Eq.~\eqref{eq:gripper_supervision}. We balance the
two classes as
\begin{equation}
    \mathcal L_{\mathrm{grip}}
    =\tfrac12\mathbb{E}[\ell_{\mathrm{grip}}\mid y=1]
    +\tfrac12\mathbb{E}[\ell_{\mathrm{grip}}\mid y=0].
    \label{eq:ppp_balanced_gripper_loss}
\end{equation}
With $\bar\tau=0.5$, the temperature controls the transition width around
zero switching benefit; the prefactor in Eq.~\eqref{eq:gripper_supervision}
bounds the loss derivative with respect to $b$ independently of temperature.

On LIBERO, the deployment switching threshold is
\begin{equation}
    \varepsilon_{\mathrm g}
    =\bar\varepsilon\,
    \operatorname{median}_{e\in\mathcal D_\ell}s_e,
    \qquad\bar\varepsilon=1,
    \label{eq:ppp_switch_threshold}
\end{equation}
where $\mathcal D_\ell$ is the demonstration set for the instructed task.
The real-robot experiments use $\varepsilon_{\mathrm g}=0$: a gripper
switches whenever its predicted benefit is positive.

\smallskip
\qheading{A shared progress field.}
Let $z_{\mathrm o}=f(z,(u_0,\mathrm{open}))$,
$z_{\mathrm c}=f(z,(u_0,\mathrm{close}))$, and
$\delta z=z_{\mathrm c}-z_{\mathrm o}$. If $p$ is continuously
differentiable along the connecting segment, then
\begin{equation}
    \begin{aligned}
        d & :=Q(u_0,\mathrm{close})-Q(u_0,\mathrm{open})                                \\
          & =\int_0^1 \nabla_z p_\theta(z_{\mathrm o}+s\delta z,\ell)^\top\delta z\,ds.
    \end{aligned}
    \label{eq:ppp_discrete_integral}
\end{equation}
This identity relates the finite gripper comparison to the same latent
progress field used for continuous actions. The switching benefit $b$ is
$d$ when the held command is open and $-d$ when it is closed. The controller
computes the two endpoint potentials directly; it does not approximate
this integral with a local gradient.

\subsection{Normalization for Control Queries}
\label{sec:ppp_query_normalization}

The predictor's output projection contains BatchNorm. With ordinary
candidate-dependent batch moments, differentiating a sum of batched query
scores gives
\begin{equation}
    \nabla_{u_i}\sum_j Q_j
    =\nabla_{u_i}Q_i+\sum_{j\ne i}\nabla_{u_i}Q_j,
    \label{eq:ppp_cross_sample_gradient}
\end{equation}
where the second term contains effects through the shared moments. We use
a separate set of normalization statistics for control queries. During
training, these moments are computed from a copy of the reference-query
batch whose candidate action is detached:
\begin{equation}
    (\mu_Q,v_Q)
    =\operatorname{moments}_{i,h}
    \varphi_{i,h}(\operatorname{sg}(u_i)),
    \label{eq:ppp_query_statistics}
\end{equation}
where $\varphi_{i,h}$ denotes the activation before this normalization
layer for sample $i$ at predictor time position $h$, conditioned on the
sample's observation history and held gripper command. The moments are
computed over all $B\times H$ positions in a batch of $B$ samples.
The moments retain their parameter gradients, but have no derivatives with
respect to the queried actions. The pose gradient and both gripper candidates
use these same moments, held fixed with respect to each candidate. This
removes cross-sample action-derivative terms from the batch-normalization
path. Accordingly, the exact derivatives in the main text are derivatives
of the query function with normalization statistics held fixed.

At deployment, we use running estimates of the query-population moments.
These are maintained separately from the normalization statistics used for
world-model prediction on recorded transitions. Training uses batch query
moments and inference uses their running estimates; both treat the moments
as constants with respect to a candidate action.

\smallskip
\qheading{Consequence for $Q$.}
World-model prediction uses normalization statistics from recorded
transitions; control queries use their own statistics to define the
candidate derivatives above. The two paths share weights but have different
latent outputs.
On $1{,}024$ demonstration anchors at the recorded command (epoch $120$),
the query branch evaluated with the world model's statistics has median
per-anchor squared-error ratios of $0.43$ (\method-GH) and $0.54$ (\method).
The ratio, $\|\hat z'-z'\|^2/\|z-z'\|^2$, compares full-latent prediction
error with copying the current latent.
At the same anchors, the median signed difference between the deployed
query's scalar value and that on the recorded next frame is $+0.13$
(\method-GH) and $+0.89$ (\method) control steps. These signed medians
summarize scalar offsets, not absolute prediction errors.
The query branch is optimized through its action-gradient and
gripper-difference readouts; closed-loop evaluation measures the resulting
control performance separately from these prediction diagnostics.
On executed counterfactual branches, the derivatives under the two
normalizations agree equally with the field under the simulator's dynamics
(\cref{sec:ppp_true_dynamics}).

A \method variant from the same branch point changes both the query
normalization to the world model's statistics and the training objective by
adding the world model's prediction loss on the query path. Its query
prediction-error ratio is $0.52$, against $0.54$ for world model prediction, and
its equal-size preference at $r=2$ (\cref{tab:tf_linearity}) rises from
$0.79$ to $0.93$. Its reference-query increment remains $Q_0=6.65$
(\cref{sec:ppp_finite_steps}), and its pooled success is $45.5$ versus
$50.5\%$ for the deployed \method over epochs $70$--$120$ ($p_{\mathrm{task}}=0.010$).
This variant has similar query and world-model prediction errors but lower
closed-loop success; the joint intervention does not isolate predictive
fidelity as the cause of the performance difference.

\subsection{Optimization and Execution}
\label{sec:ppp_optimization_details}

\smallskip
\qheading{Training schedule.}
First, the encoder and dynamics train alone from random initialization
with Eq.~\eqref{eq:ppp_world_model_loss} for $100$ epochs on all $6{,}500$ LIBERO demonstrations ($130$
tasks, including LIBERO-90), with AdamW (learning rate
$5\times10^{-5}$, weight decay $10^{-3}$, batch size $512$), five epochs
of warmup, and cosine decay to zero. The final checkpoint initializes
the encoder and dynamics of every method. Second, each policy is
trained on the same demonstrations. For the first ten epochs the world
model is frozen: \method calibrates only its potential with
$\mathcal L_p$, and the BC variants train only their heads. From epoch
$10$, the world model is trained jointly with the policy, and its
predictive loss is retained. \method's action supervision therefore
starts with world model training, whereas BC's starts against the frozen
world model. The policy learning-rate schedule is
configured for $200$ epochs; we evaluate checkpoints up to epoch $120$
and compare methods at matched epochs. In \method's joint stage,
Eq.~\eqref{eq:joint_objective} updates the parameter groups according to
Table~\ref{tab:ppp_gradient_paths}. In particular, action imitation retains
the computation graph through $g_u$ and uses double backpropagation.
The detached gain input prevents an additional action-loss path from the
gain network into the predicted state; the gradient readout remains
differentiable with respect to the potential, dynamics, and encoder.

Policy training uses AdamW with batch size $512$ and weight decay $10^{-3}$.
The learning rates are $3\times10^{-4}$ for the potential and gain networks
and $5\times10^{-5}$ for the world model, with five epochs of warmup
followed by cosine decay. Full hyperparameters are listed in
Table~\ref{tab:ppp_hyperparameters}.
Progress-fit and candidate-selection metrics use a fixed diagnostic subset of the training demonstrations.

\begin{table}[t]
    \centering
    \small
    \caption{Gradient paths during joint training.}
    \begin{tabular}{@{}lcccc@{}}
        \toprule
        Loss                         & $E$ & $f$ & $p$ & $\eta_\phi$ \\
        \midrule
        $\mathcal L_p$               & yes & --  & yes & --          \\
        $\mathcal L_{\mathrm{act}}$  & yes & yes & yes & yes         \\
        $\mathcal L_{\mathrm{grip}}$ & --  & --  & yes & --          \\
        $\mathcal L_{\mathrm{WM}}$   & yes & yes & --  & --          \\
        \bottomrule
    \end{tabular}
    \label{tab:ppp_gradient_paths}
\end{table}

\smallskip
\qheading{Execution.}
The execution rate and prediction horizon are specified in
Sec.~\ref{sec:experiment_implementation}. Each decision uses one
action-gradient evaluation and a gripper progress comparison, without
action sampling or iterative search. Pose commands are denormalized and
clipped to the robot's action limits. After one raw tick of execution,
both readouts are recomputed from the next observation. The pose gradient
is not recomputed after a gripper switch within the same decision. At
episode start, missing context is filled with the first observation and
median-pose, open-gripper history.

\smallskip
\phantomsection
\qheading{Gripper settling: motivation.}
\label{sec:ppp_gripper_settling}
In an earlier LIBERO-Spatial run without query normalization, conditioning
the pose query on a close command reduced descent before the grasp had
settled. This motivated delaying closed-gripper conditioning until physical
closure settles; \cref{tab:libero_epoch120,tab:libero_trend} report settling
comparisons for the final models.

\smallskip
\qheading{Gripper settling: implementation.}
For the LIBERO settling variant, the pose query uses a separate gripper
command history that remains open during physical closure. Finger opening
is measured as the sum of the absolute finger joint positions. Starting three
raw ticks after a close command, closure is considered settled when this
opening decreases by less than $2$\,mm in one tick, with a maximum wait of
$15$ ticks. Subsequent pose queries use the closed reference command and
record closed commands in their history.
The gripper readout continues to use the actual command history throughout;
reopening resets the settling state. This adjustment changes only the pose
query's gripper conditioning and requires no retraining.

\subsection{Training Trends}
\label{sec:ppp_training_trends}
\Cref{tab:libero_trend} reports average success at every evaluated epoch.
\method without settling exceeds BC and BC+Progress at each of them.
After epoch $40$ its success fluctuates between checkpoints ($43.4\%$ to $52.2\%$), motivating comparisons pooled over epochs $70$--$120$.
With success rates pooled over these epochs, \method without settling outperforms BC on $36$ of the $40$ tasks ($p_{\mathrm{task}}<10^{-5}$).

\smallskip
\qheading{Task-level comparisons.}
For each comparison, we aggregate evaluation outcomes within each task and compute one success-rate difference per task between the two methods, pooling across checkpoints where specified.
The sign-flip test operates on these task-level differences and does not require episodes within a task to be independent.
It assumes that, under the null hypothesis, the joint distribution of task-level differences is invariant to independent sign flips.

\input{sec/table/libero_trend.tex}

\subsection{Training Without Progress Labels}
\label{sec:ppp_nolp}

\smallskip
\qheading{Configuration.}
No $\mathcal L_p$ sets the weights of all five progress losses ($\mathcal L_{\mathrm{val}}$, $\mathcal L_{\mathrm{inc}}$, $\mathcal L_{\mathrm{rank}}$, $\mathcal L_{\mathrm{bd}}$, $\mathcal L_{\mathrm{lang}}$) to zero.
The potential is therefore trained only by $\mathcal L_{\mathrm{act}}$ through the gradient readout; the switching head, $\mathcal L_{\mathrm{WM}}$, the query normalization, the data and the optimizer are \method-GH's, and the run is deployed as \method-GH.
Language reaches the potential only through the action loss, as in BC.
It trains from initialization on BC's schedule: action loss from epoch $0$ with five epochs of learning-rate warmup, and the world model frozen in epochs $0$--$9$. It differs from BC only in the readout.
\Cref{sec:ppp_schedule_sensitivity} compares BC, No $\mathcal L_p$, and \method-GH under an alternative schedule that aligns the onset of action learning with world model updates.

\smallskip
\qheading{Results.}
\Cref{tab:ppp_nolp} gives success by suite, pooled over epochs $70$--$120$, and \cref{tab:libero_trend} gives every epoch.
The pooled comparisons in \cref{tab:ppp_nolp} are paired on the same $8{,}000$ episodes.
Without labels, No $\mathcal L_p$ beats BC early (by $5.3$--$7.8$ points at epochs $20$--$50$, $p_{\mathrm{task}}\le0.025$ at three of the four), but their pooled difference over epochs $70$--$120$ is not statistically significant.

\smallskip
\qheading{The potential without labels.}
On the training-log diagnostic subset ($650$ training demonstrations, five per task across all $130$ tasks), the label-free scalar's task-averaged Spearman correlation with progress is $0.68$--$0.71$ across evaluated checkpoints, against $0.988$--$0.997$ for \method-GH.
This statistic pools later-frame predictions within each task before averaging correlations across tasks.
On the separate $40$-task field-analysis subset, its mean per-trajectory correlation is $0.5822$ at epoch $120$ (\cref{tab:tf_alignment}).
The gauge of \cref{sec:method_learning} is visible in the trained weights.
The mobility stays near its initialization ($\eta_\phi=1.01$ at epoch $120$, median over the diagnostic states), whereas \method-GH's grows to $12.9$.
The potential's gradient grows to $13$ times \method-GH's, and its values remain uncalibrated (median $-7.9$ at the start of a rollout, against $-1.18$ for \method).

\begin{table}[t]
    \centering
    \caption{Training without progress labels: success (\%) on each suite, pooled over epochs $70$--$120$ ($2{,}000$ rollouts per suite).
        $\mathcal L_p$: progress losses used; Act.: first epoch of action supervision; the world model trains from epoch $10$ in every run.
        Below: paired differences in points, task-level sign-flip $p_{\mathrm{task}}$, and the number of tasks on which the first run is better and worse.}
    \label{tab:ppp_nolp}
    \footnotesize
    \setlength{\tabcolsep}{2.6pt}
    \begin{tabular}{@{}lccrrrrr@{}}
        \toprule
        Run                  & $\mathcal L_p$ & Act. & Spatial & Object & Goal & Long & Avg. \\
        \midrule
        BC                   & --             & 0    & 28.7    & 53.8   & 45.4 & 17.0 & 36.2 \\
        No $\mathcal L_p$    & no             & 0    & 43.8    & 40.5   & 47.7 & 22.3 & 38.6 \\
        \method-GH           & yes            & 10   & 45.1    & 53.4   & 49.9 & 25.5 & 43.5 \\
        \method w/o settling & yes            & 10   & 39.1    & 74.4   & 50.0 & 38.6 & 50.5 \\
        \bottomrule
    \end{tabular}

    \vspace{4pt}
    \begin{tabular}{@{}lrrr@{}}
        \toprule
        Comparison                       & $\Delta$ & $p_{\mathrm{task}}$ & Tasks $+/-$ \\
        \midrule
        No $\mathcal L_p$ $-$ BC         & $+2.34$  & $0.38$              & 26/12       \\
        \method-GH $-$ No $\mathcal L_p$ & $+4.90$  & $0.011$             & 27/12       \\
        \bottomrule
    \end{tabular}
\end{table}

\input{sec/X_training_schedule}

\subsection{Evaluation with a Seer Backbone}
\label{sec:app_seer}

We additionally evaluate \method with a Seer~\cite{tian2025seer} backbone to examine
its applicability beyond the LeWM-based~\cite{maes2026leworldmodel} architecture used
in our main experiments.
Seer is a predictive inverse dynamics model: a $24$-layer transformer forecasts future frames from past observations and the instruction, and action queries read that foresight to predict the next commands.
Its network never receives a candidate command, so \method's readout needs an action-conditioned query added to it.
We compare four variants:
\begin{itemize}
    \item \textbf{Seer}, the original policy with its direct action head;
    \item \textbf{Seer-AC}, which feeds the candidate command into Seer's network, so that the whole network forecasts under the command, and reads the action with Seer's head at the reference command;
    \item \textbf{Seer-AC-PPP}, the same action-conditioned network with \method's field-based readout through it in place of Seer's head;
    \item \textbf{Seer-PPP}, which keeps Seer's network action-free and unchanged and adds a small world model for \method's readout, mirroring the encoder, world model, and potential of the main experiments.
\end{itemize}
All four train on the demonstrations of all $130$ LIBERO tasks and are evaluated on the four LIBERO suites with the same $50$ initial states per task across $40$ tasks.
Seer and Seer-PPP at C50 are the Seer rows of \cref{tab:libero_epoch120}.

\smallskip
\qheading{Seer-PPP.}
The state $s$ is the output of Seer's three action queries ($3\times384$), computed by Seer's unmodified network.
A world model $f(s,u)=s+\Delta(s,u)$ of four pre-norm transformer blocks ($7.5$M parameters) reads $s$ and the command of a $3$-tick chunk.
It is trained to predict the action-query state of the window shifted by three frames, with the target under a stop-gradient.
The potential $p_\theta(f(s,u),\ell)$ is \method's FiLM-conditioned MLP on Seer's CLIP instruction embedding, and the pose and gripper follow \cref{eq:pose_policy,eq:gripper_policy} at $u_0=0$ (no motion) in raw LIBERO command units.
Each tick of the chunk is read from its own derivative, and the chunks are executed with Seer's temporal ensembling.
The switching threshold $\varepsilon_{\mathrm g}$ is one control step of predicted progress, and gripper settling is not used.
$\mathcal L_p$ trains $p_\theta$ on Seer's states of real frames.
$\mathcal L_{\mathrm{act}}$ and $\mathcal L_{\mathrm{grip}}$ reach $f$ with their gradient scaled by $0.3$ and Seer's network with it scaled by $0.03$, so that each module's action gradient stays a few times the gradient of its own predictive loss; Seer's network keeps its image-forecast loss.
Seer-PPP trains from scratch with Seer's recipe: $50$ epochs, batch $512$, AdamW at learning rate $10^{-3}$ ($3\times10^{-4}$ for $p_\theta$, the mobility, and $f$) with warm-up and cosine decay, and Seer's global gradient clipping.
The action and gripper losses start after the first epoch.

\smallskip
\qheading{Seer-AC and Seer-AC-PPP.}
Both add command-input tokens to Seer's network, so that its forecast at each step depends on that step's candidate command chunk.
Seer-AC's action queries read the forecast at the reference command; Seer-AC-PPP's potential reads the action queries' state under the candidate command, and its action and progress gradients into the network are scaled by $0.03$ to protect the forecast.
Following the protocol of the main experiments, both start from Seer's network pretrained for $50$ epochs as an action-conditioned world model and then train for $50$ epochs at learning rate $3\times10^{-4}$.
Their checkpoint C$k$ therefore follows $50$ epochs of world-model pretraining, whereas Seer and Seer-PPP train from scratch.

\begin{table}[t]
    \centering
    \caption{
        Success rates (\%) with a Seer backbone, averaged
        over 40 LIBERO tasks with 50 evaluation episodes
        per task.
        C$k$ denotes the checkpoint at training epoch $k$; for Seer-AC
        and Seer-AC-PPP, after $50$ epochs of world-model pretraining.
    }
    \label{tab:seer_backbone}
    \setlength{\tabcolsep}{3.5pt}
    \small
    \begin{tabular}{lrrrrr}
        \toprule
        Method
         & C10   & C20   & C30   & C40 & C50 \\
        \midrule
        Seer
         & 70.30 & 89.10 & 93.30
         & 96.35 & 97.30                     \\
        Seer-AC
         & 71.15 & 86.90 & 92.35
         & 94.35 & 94.70                     \\
        Seer-AC-PPP
         & 81.75 & 89.10 & 91.70
         & 93.00 & 92.35                     \\
        Seer-PPP
         & 54.45 & 80.65 & 91.15
         & 96.00 & 97.10                     \\
        \bottomrule
    \end{tabular}
\end{table}

\begin{table}[t]
    \centering
    \caption{
        Success rates (\%) with a Seer backbone at C50 on each
        LIBERO suite ($500$ rollouts per suite).
    }
    \label{tab:seer_backbone_suite}
    \setlength{\tabcolsep}{3.5pt}
    \small
    \begin{tabular}{@{}lrrrrr@{}}
        \toprule
        Method      & Spatial & Object & Goal & Long & Avg. \\
        \midrule
        Seer        & 98.8    & 99.0   & 97.8 & 93.6 & 97.3 \\
        Seer-AC     & 96.6    & 99.8   & 94.6 & 87.8 & 94.7 \\
        Seer-AC-PPP & 95.0    & 96.4   & 93.4 & 84.6 & 92.4 \\
        Seer-PPP    & 99.4    & 99.8   & 98.2 & 91.0 & 97.1 \\
        \bottomrule
    \end{tabular}
\end{table}

\smallskip
\qheading{Results.}
Seer-PPP learns more slowly than Seer early in training but reaches similar
success at C40--C50 (\cref{tab:seer_backbone}). At C50 it achieves $97.10\%$
against Seer's $97.30\%$, with no significant difference detected
($p_{\mathrm{task}}=0.83$); \cref{tab:seer_backbone_suite} gives the suite breakdown.
It exceeds Seer-AC by $2.40$ points ($p_{\mathrm{task}}=0.007$) and
Seer-AC-PPP by $4.75$ ($p_{\mathrm{task}}<10^{-5}$).
Second runs of these two variants reach $95.40\%$ and $92.40\%$;
Seer-PPP remains higher by $1.70$ and $4.70$ points, respectively
($p_{\mathrm{task}}=0.01$ and $<10^{-5}$).
At C50, Seer-AC trails Seer by $2.60$ points ($p_{\mathrm{task}}=0.003$),
and Seer-AC-PPP trails Seer-AC by $2.35$ ($p_{\mathrm{task}}=0.007$).

These results support the field readout's applicability to a second backbone.
The comparison evaluates complete training recipes: Seer-PPP and Seer share
network, data, and training schedule and differ in the readout and the added
world model; the action-conditioned variants also differ in schedule.
Seer and Seer-PPP are each a single training run.

\subsection{Direct Action-Space Scalar Potential}
\label{sec:ppp_actpot}

\smallskip
\qheading{Model and training.}
Direct-AP replaces \method-GH's composed scalar query with
\begin{equation}
    \begin{aligned}
        Q(u) & =\Phi(\bar z,u,\ell),\qquad \bar z=f(z,a_0),                      \\
        u    & =\eta_\phi(\bar z)\left.\nabla_u\Phi(\bar z,u,\ell)\right|_{u=0}.
    \end{aligned}
    \label{eq:ppp_direct_ap}
\end{equation}
The same FiLM-MLP trunk receives the six normalized pose channels concatenated to its latent input.
The reference prediction $\bar z$ is independent of the candidate $u$, so the dynamics' action Jacobian is absent from the readout.
Action imitation differentiates through $\nabla_u\Phi$ and still updates the encoder and dynamics through $\bar z$.
The mobility, gripper-switching head, world model loss, query normalization, data, optimizer, and stage schedule are \method-GH's.
The potential and policy heads train from initialization on four GPUs to epoch $120$, starting from the shared pretrained world model.
This is a comparison of action-conditioning constructions with a retained world model backbone, not an ablation of that backbone.

The state potential $p(z,\ell):=\Phi(z,0,\ell)$ receives the original progress losses.
An additional action-value loss $\mathcal L_Q$, with weight $1$, applies the same loss terms and weights to the pair
$\bigl(\operatorname{sg}\Phi(z_t,0,\ell),\Phi(\bar z,\bar u_E,\ell)\bigr)$,
with targets $p_t^\star$ and $p_{t+1}^\star$, where $t+1$ denotes the next model-grid frame and $\bar u_E$ is the expert's chunk-mean pose command.
The mismatched-instruction term uses $\Phi(\bar z,\bar u_E,\tilde\ell)$.
This loss follows the progress-supervision schedule, including the initial ten epochs with the world model frozen; action imitation and world model co-training start at epoch $10$.
This tests an alternative scalar parameterization with additional action-value supervision.

\smallskip
\qheading{Results and scope.}
\Cref{tab:libero_epoch120,tab:libero_trend} report the final checkpoint and training trajectory.
Pooled over epochs $70$, $80$, $100$, and $120$, success is $41.2/51.2/51.3/22.5\%$ on Spatial/Object/Goal/Long, or $41.6\%$ overall.
On the diagnostic subset of training demonstrations at epoch $120$, $\Phi(\cdot,0,\ell)$ has progress correlation $\rho=0.9975$; the mean absolute error of $\Phi(\bar z,\bar u_E,\ell)$ against next-frame progress is $0.099$.
The pooled advantage over BC is $5.34$ points ($p_{\mathrm{task}}=0.036$), while \method-GH's advantage over Direct-AP is $1.90$ points ($p_{\mathrm{task}}=0.351$).
Direct-AP retains a co-trained world model for its reference-state representation.
Both constructions support the same local ascent argument (\cref{sec:ppp_local_control_proof}).
The explicit $p_\theta\circ f$ construction exposes the latent action response and task differential; the direct scalar learns their combined action dependence through its values and gradient, without requiring an identifiable internal decomposition.
The comparison supports either parameterization for the gradient readout and does not establish a performance advantage for explicit factorization.

\subsection{Pose from an Action Head}
\label{sec:ppp_ah}

\smallskip
\qheading{Configuration.}
\method-AH is the converse of \method-GH: it takes the gripper from the field and the pose from an action head.
The pose is predicted by BC's policy head from the reference latent $\bar z=f(z,a_0)$, with BC's architecture and the action loss of \cref{eq:action_supervision}; the mobility is not used.
The gripper follows \method's finite-difference switching (\cref{eq:gripper_policy}) and is trained with $\mathcal L_{\mathrm{grip}}$ as in \method.
The potential, $\mathcal L_p$, the query normalization, $\mathcal L_{\mathrm{WM}}$, the data, and the optimizer are \method's.
The pose loss reaches the encoder and dynamics through $\bar z$ but not the potential, which is trained only by $\mathcal L_p$ and $\mathcal L_{\mathrm{grip}}$.
\method-AH trains from initialization under the original protocol of the main comparison, on BC's schedule: pose and gripper supervision from epoch $0$ with five epochs of learning-rate warmup, $\mathcal L_p$ from epoch $0$ as in \method, and the world model frozen in epochs $0$--$9$.
It thus differs from BC in the gripper readout and in the added potential with its progress losses, and from \method in the pose readout and in the onset of action supervision (epoch $0$ instead of $10$).
Deployment uses \method's switching threshold (\cref{eq:ppp_switch_threshold}); as for \method, the same checkpoints are evaluated with and without gripper settling.

\begin{table}[t]
    \centering
    \caption{\method-AH: LIBERO success rates (\%) by suite at epoch $120$ ($500$ rollouts per suite), on the same initial states as \cref{tab:libero_epoch120}.
        The two rows share a checkpoint and differ only in gripper settling at deployment.}
    \label{tab:ppp_ah}
    \small
    \setlength{\tabcolsep}{4pt}
    \begin{tabular}{@{}lrrrrr@{}}
        \toprule
        Method                  & Spatial & Object & Goal & Long & Avg. \\
        \midrule
        \method-AH w/o settling & 31.6    & 76.2   & 50.6 & 30.6 & 47.3 \\
        \method-AH              & 35.8    & 73.2   & 52.0 & 31.6 & 48.2 \\
        \bottomrule
    \end{tabular}
\end{table}

\begin{table}[t]
    \centering
    \caption{\method-AH: average LIBERO success rates (\%) across training epochs, on the same $2{,}000$ initial states at each checkpoint as \cref{tab:libero_trend}.
        The two rows share a checkpoint at each epoch and differ only in gripper settling at deployment.}
    \label{tab:ppp_ah_trend}
    \small
    \setlength{\tabcolsep}{2.1pt}
    \begin{tabular}{@{}lrrrrrrrr@{}}
        \toprule
                   & \multicolumn{8}{c}{Training epoch}                                                  \\
        \cmidrule(l){2-9}
        Method     & 20                                 & 30   & 40   & 50   & 70   & 80   & 100  & 120  \\
        \midrule
        \shortstack[l]{\method-AH\\w/o settling}
                   & 29.8                               & 30.2 & 31.1 & 38.8 & 44.5 & 39.9 & 43.1 & 47.3 \\
        \method-AH & 31.4                               & 32.9 & 32.2 & 39.5 & 46.9 & 42.4 & 45.6 & 48.2 \\
        \bottomrule
    \end{tabular}
\end{table}

\smallskip
\qheading{Results.}
\Cref{tab:ppp_ah} gives the epoch-$120$ success by suite and \cref{tab:ppp_ah_trend} gives every evaluated epoch; the other methods are in \cref{tab:libero_epoch120,tab:libero_trend}.
Pooled over epochs $70$--$120$ without settling, \method-AH reaches $43.7\%$, $7.44$ points above BC ($p_{\mathrm{task}}=0.0012$), level with \method-GH ($43.5\%$; $p_{\mathrm{task}}=0.95$), and $6.87$ points below \method ($p_{\mathrm{task}}=0.0018$).
It exceeds BC at every evaluated checkpoint, by $4.6$ to $11.4$ points ($p_{\mathrm{task}}<0.05$ at five of the eight), and trails \method without settling at every checkpoint from epoch $30$ on.
Its potential fits progress as closely as \method's (task-averaged Spearman $\rho=0.996$, versus $0.997$ for \method), and its pooled advantage over BC+Progress, which also trains progress losses but switches with a head, is $7.13$ points ($p_{\mathrm{task}}=0.0081$).
Settling raises \method-AH's pooled success by $2.11$ points ($p_{\mathrm{task}}=0.016$), and \method keeps its lead when both use it ($+6.34$, $p_{\mathrm{task}}=0.0053$).

\smallskip
\qheading{Where each readout helps.}
The two field readouts improve on BC in different suites.
At epoch $120$, \method-AH, with the field's gripper and a head's pose, matches \method on Object ($76.2\%$ for both without settling; BC $53.6\%$) but stays at BC's level on Spatial ($31.6\%$ versus $31.4\%$).
\method-GH, with the field's pose and a head's gripper, reaches $46.8\%$ on Spatial and $64.6\%$ on Object (\cref{tab:libero_epoch120}).
Pooled over epochs $70$--$120$, the same pattern holds: \method-AH reaches $73.9\%$ on Object and $29.0\%$ on Spatial, against $53.4\%$ and $45.1\%$ for \method-GH (\cref{tab:ppp_nolp}).
The gains of \method-AH and \method-GH over BC are of similar size ($7.4$ and $7.2$ points pooled), and \method, which reads both commands from the field, exceeds each of them by about as much again.

\smallskip
\qheading{Scope.}
\method-AH shares BC's schedule, so \method-AH $-$ BC compares the gripper readout together with the added potential and its progress losses; the BC+Progress comparison indicates that progress losses on a head-based policy do not account for the gain (\cref{tab:progress_vs_control}).
These results indicate that the field-based gripper and the field-based pose readout each contribute to \method's success under this schedule, mainly on different suites.

\subsection{Progress-Selection Baseline}
\label{sec:ppp_selection_baseline}

\smallskip
\qheading{Model and training.}
BC-MH has the architecture and losses of BC+Progress, except that its
policy network outputs $K=4$ pose hypotheses $u_1,\ldots,u_K$ and $K$
mode logits in addition to the progress estimate; the gripper head is
unchanged. Let $k_E$ index the hypothesis nearest the expert command. The
continuous-action loss of BC is applied to every hypothesis, with weight
$0.95$ for $k_E$ and $0.05/(K-1)$ for each of the others (a relaxed
winner-takes-all loss), and the mode logits are trained by cross-entropy
toward $k_E$. The selection loss generalizes the gripper loss of
Eq.~\eqref{eq:gripper_supervision} from two candidates to $K$. With
$Q_k=p(f(z,(u_k,c^{\mathrm h})),\ell)$, the scalar value assigned to
hypothesis $k$ under PPP's query construction,
\begin{equation}
    \mathcal L_{\mathrm{sel}}
    =\tau_e\,\mathrm{CE}\!\left(\operatorname{softmax}_k(Q_k/\tau_e),\,k_E\right),
    \qquad \tau_e=\bar\tau s_e,
    \label{eq:ppp_selection_loss}
\end{equation}
with $\bar\tau=0.5$ and weight $10$, as for the gripper loss. As there,
the loss updates only the policy network: the encoder, the dynamics and
the hypotheses $u_k$ are detached inside it. BC-MH is trained from
scratch for $120$ epochs on $4$ GPUs, as BC and BC+Progress. Training
and deployment use the same candidate query.

\smallskip
\qheading{Deployment rules.}
At every decision, the same checkpoint executes one hypothesis chosen by
one of four rules: the highest mode logit (own pick); the highest $Q_k$
(progress pick); the highest $Q_k$ only if it exceeds the own pick's by
more than half a control step of progress, and the own pick otherwise
(margin $0.5$); or a uniformly random hypothesis (random pick).

\smallskip
\qheading{Results.}
Table~\ref{tab:ppp_selection_rules} compares the four rules at epoch
$120$; none differs significantly from the own pick. Over all eight
evaluated checkpoints, the progress pick trails the own pick by $1.1$
points ($p_{\mathrm{task}}=0.06$). BC-MH's own pick trails BC by $5.1$
points pooled ($p_{\mathrm{task}}=0.007$).
On the diagnostic subset of training demonstrations, from epoch $70$ to
$120$, the progress pick is the hypothesis nearest the expert in
$29$--$31\%$ of samples (chance $25\%$), the mode logits in $58$--$68\%$,
and the median gap between the two highest $Q_k$ is $0.050$--$0.052$
control steps of progress. One world model step thus separates the
candidates by far less than the one-step threshold of progress-based
switching, whose trained switching benefit has a median magnitude of
$1.3$ to $3.6$ control steps in PPP from epoch $20$ to $120$.

\begin{table}[t]
    \centering
    \caption{BC-MH at epoch $120$ under four rules for choosing which pose
        hypothesis to execute, on the same $2{,}000$ episodes. Succ.\ is the
        success rate, $\Delta$ its paired difference from the own pick in
        points, and override the fraction of
        decisions whose executed hypothesis differs from the own pick, both
        in percent.}
    \label{tab:ppp_selection_rules}
    \small
    \setlength{\tabcolsep}{3.5pt}
    \begin{tabular}{@{}lrrrr@{}}
        \toprule
        Rule                 & Succ. & $\Delta$ & $p_{\mathrm{task}}$ & Override \\
        \midrule
        Own pick             & 36.3  & --       & --                  & 0.0      \\
        Progress pick        & 35.8  & $-0.45$  & 0.75                & 72.8     \\
        Progress, margin 0.5 & 36.6  & $+0.30$  & 0.61                & 9.7      \\
        Random pick          & 37.1  & $+0.80$  & 0.69                & 75.0     \\
        \bottomrule
    \end{tabular}
\end{table}

\subsection{World-Model Search Baselines}
\label{sec:ppp_search_baseline}

\smallskip
\qheading{Search on query values.}
CEM deploys PPP's epoch-$120$ checkpoint and replaces only how the pose
command is read out. Let $Q_H(u_{0:H-1})=p(z_H,\ell)$, where
$z_{h+1}=f(z_{\le h},(u_h,c^{\mathrm h}))$ is PPP's query
(Sec.~\ref{sec:ppp_query_details}) and each query output is fed back as the
newest context frame. CEM searches $H$ pose commands to maximize $Q_H$ by
the cross-entropy method~\cite{rubinstein1999cross}, with:
\begin{itemize}
    \item $300$ Gaussian samples in normalized action units, with initial
          standard deviation $1$ (the standard deviation of the training actions);
    \item $30$ elites, refit momentum $0.1$, and a floor of $0.1$ on the
          standard deviation;
    \item every sample clipped per dimension to the $0.5$--$99.5$th
          percentiles of the normalized training commands, and its norm to their
          $99$th percentile ($5.36$);
    \item the current mean always among the candidates, and the first
          command of the refit mean executed.
\end{itemize}
Like PPP, CEM replans at every tick. It warm-starts from the previous
solution, shifted by one control step every $k$ ticks, with four refits
at the first tick of each control step and one at each of the three ticks
between. The gripper is handled in one of two ways:
\begin{itemize}
    \item \emph{switch}: PPP's progress-based switch, with the same threshold;
    \item \emph{search}: searched as one Bernoulli command per step, whose
          probability is refit to the fraction of elites that close, clipped to
          $[0.05,0.95]$.
\end{itemize}
At $H=1$ and $u=0$ the objective equals PPP's query value at the
reference command. Candidates do not interact, because at deployment the
query normalization uses its running statistics.

\smallskip
\qheading{Goal-reaching search.}
CEM-Goal uses the encoder and world model from \method's epoch-$120$ checkpoint
and minimizes $\lVert z_4-z_{\mathrm{goal}}\rVert^2$. Here $z_4$ is the
world model's own four-step prediction from the executed action history,
with its own normalization statistics and without the progress model or
the instruction. It searches pose and gripper commands as above. The
benchmark's initial states have no demonstrations, so the subgoal comes
from training data:
\begin{itemize}
    \item at the start of an episode, it retrieves the training
          demonstration of the same task whose first-frame latent is nearest the
          current one;
    \item at every tick, it matches the current latent to that demonstration
          within a forward window of eight control steps, and aims four control
          steps past the match.
\end{itemize}

\smallskip
\qheading{Results.}
Table~\ref{tab:ppp_search_arms} varies the horizon and the gripper
rule. The four progress searches produce one success in $8{,}000$
rollouts.
\begin{itemize}
    \item \emph{Command size.} In failed episodes their commands stay near
          the norm cap, with a median norm of $4.4$--$4.9$ against $1.2$ for PPP.
          The hand travels $3.1$--$3.4$\,m against $0.95$\,m.
    \item \emph{Local gradient alignment.} The first-order change in $Q$
          of their commands, $g^\top a$, is at most a third of PPP's. Their
          larger query values come from its response to large commands,
          not from the local slope.
    \item \emph{Horizon and gripper.} Imagining four steps does not help.
          Searching the gripper adds switching: $18$ flips per failed episode with
          the four-step horizon.
\end{itemize}
CEM-Goal succeeds in $6.2\%$ of rollouts, most often on Goal tasks
($15.4\%$), while its searched gripper flips $11$ times per failed episode.
On an otherwise idle RTX~A5000, the one-step searches take $16$\,ms per
decision and the four-step searches $56$\,ms.

\smallskip
\qheading{Where query values can be maximized.}
Table~\ref{tab:ppp_search_radius} replays five training demonstrations of
one task per suite ($385$ decisions) through the same planner,
teacher-forced, and compares the progress search under smaller norm caps
with PPP's readout and the expert's command.
\begin{itemize}
    \item At every cap, the search finds commands with higher query
          values than the expert's own in $96$--$100\%$ of states.
    \item At the training-data cap, those commands are nearly orthogonal
          to the expert's.
    \item Shrinking the cap turns the search toward the expert, but its
          refit mean then averages elites that lie on the cap in different
          directions, and the command shrinks far below the expert's.
\end{itemize}
These probes support reading the gradient near the reference command,
where the search direction aligns more closely with the expert's.
PPP's readout also sets the command size.

\begin{table*}[t]
    \centering
    \caption{World-model search on PPP's epoch-$120$ checkpoint
        ($2{,}000$ rollouts per row, the same initial states). $H$ is the
        imagined horizon in control steps. Gripper: PPP's progress-based
        switch, or searched (Bernoulli). Success in \%. $\lVert a\rVert$ (the
        normalized pose command), path (the end-effector path length) and
        flips (gripper switches) are medians over failed episodes.}
    \label{tab:ppp_search_arms}
    \small
    \setlength{\tabcolsep}{4.5pt}
    \begin{tabular}{@{}lclrrrrrrrr@{}}
        \toprule
        Method           & $H$ & Gripper & Spatial & Object & Goal & Long & Avg. & $\lVert a\rVert$ & Path (m) & Flips \\
        \midrule
        PPP w/o settling & --  & switch  & 40.4    & 76.2   & 52.4 & 37.0 & 51.5 & 1.16             & 0.95     & 2     \\
        CEM              & 1   & switch  & 0.0     & 0.0    & 0.0  & 0.0  & 0.0  & 4.87             & 3.28     & 3     \\
        CEM              & 1   & search  & 0.0     & 0.0    & 0.2  & 0.0  & 0.05 & 4.87             & 3.34     & 6     \\
        CEM              & 4   & switch  & 0.0     & 0.0    & 0.0  & 0.0  & 0.0  & 4.39             & 3.13     & 3     \\
        CEM              & 4   & search  & 0.0     & 0.0    & 0.0  & 0.0  & 0.0  & 4.44             & 3.42     & 18    \\
        CEM-Goal         & 4   & search  & 4.6     & 1.4    & 15.4 & 3.2  & 6.2  & 3.89             & 2.35     & 11    \\
        \bottomrule
    \end{tabular}
\end{table*}

\begin{table}[t]
    \centering
    \caption{Teacher-forced replays of training demonstrations: the
        progress search under norm caps, against PPP's readout. For $H=4$ the
        whole four-command plan is compared with the expert's next four commands. Cos.\
        is the cosine to the expert's command, $\lVert a\rVert$ the median
        command norm (expert: $2.39$), and $>$Expert the fraction of
        states where the command's predicted progress exceeds that of the
        expert's command.}
    \label{tab:ppp_search_radius}
    \small
    \setlength{\tabcolsep}{4pt}
    \begin{tabular}{@{}lrrrr@{}}
        \toprule
        Readout        & Cap  & Cos. & $\lVert a\rVert$ & $>$Expert \\
        \midrule
        PPP (gradient) & --   & 0.97 & 2.39             & 52\%      \\
        CEM, $H=1$     & 5.36 & 0.17 & 4.57             & 96\%      \\
        CEM, $H=1$     & 2.0  & 0.59 & 0.54             & 98\%      \\
        CEM, $H=1$     & 1.0  & 0.78 & 0.38             & 100\%     \\
        CEM, $H=4$     & 5.36 & 0.15 & 4.88             & 100\%     \\
        CEM, $H=4$     & 2.0  & 0.24 & 1.54             & 100\%     \\
        CEM, $H=4$     & 1.0  & 0.46 & 0.68             & 99\%      \\
        \bottomrule
    \end{tabular}
\end{table}

\begin{table}[t]
    \centering
    \small
    \caption{Architecture and optimization settings.}
    \begin{tabular}{@{}p{0.36\linewidth}p{0.59\linewidth}@{}}
        \toprule
        Component               & Setting                                                                  \\
        \midrule
        Latent and context      & $d=384$, $H=3$, $k=4$ raw ticks                                          \\
        Dynamics predictor      & 6 layers, 16 heads, MLP width 2048, dropout 0.1                          \\
        SIGReg weight           & $\lambda_{\mathrm{SIG}}=0.09$                                            \\
        Potential               & 4 residual FiLM blocks, width 1024, FiLM hidden width 256                \\
        Scalar gain             & 2 hidden layers of width 256, scalar output, $\varepsilon_\eta=10^{-3}$  \\
        Progress weights        & $\lambda_{\mathrm{val}}=\lambda_{\mathrm{inc}}=\lambda_{\mathrm{bd}}=1$;
        $\lambda_{\mathrm{rank}}=\lambda_{\mathrm{lang}}=0.5$                                              \\
        Progress settings       & $\tau_r=0.05$, $\rho_{\mathrm{term}}=0.95$,
        $\varepsilon_{\mathrm{nt}}=0.05$, $m_\ell=0.2$                                                     \\
        Action weights          & $\lambda_{\mathrm{act}}=\lambda_{\mathrm{vec}}=\lambda_{\mathrm{dir}}=1$ \\
        Gripper settings        & $\lambda_{\mathrm{grip}}=10$, $\bar\tau=0.5$,
        $\bar\varepsilon=1$                                                                                \\
        Prediction weight       & $\lambda_{\mathrm{WM}}=1$                                                \\
        Optimizer               & AdamW, batch size 512, weight decay $10^{-3}$,
        gradient-norm clipping at 1                                                                        \\
        Learning rates          & $3\times10^{-4}$ for potential and gain;
        $5\times10^{-5}$ for encoder and dynamics                                                          \\
        World-model pretraining & 100 epochs, learning rate $5\times10^{-5}$, 5-epoch warmup,
        cosine decay to 0                                                                                  \\
        Schedule                & 5-epoch warmup, cosine decay over 200 configured epochs to
        $3\times10^{-5}$; evaluated to epoch 120; world model frozen for epochs 0--9                       \\
        Deployment              & 20\,Hz replanning; 4-tick prediction, 1-tick execution                   \\
        \bottomrule
    \end{tabular}
    \label{tab:ppp_hyperparameters}
\end{table}

\input{sec/compute_cost.tex}

\section{Real-Robot Evaluation}
\label{sec:real_robot_details}

\smallskip
\qheading{Training.}
The two Franka tasks follow the LIBERO recipe (\cref{sec:ppp_optimization_details}) with
the following changes. Each task has its own world model, trained from
random initialization for $100$ epochs on that task's demonstrations;
\method and BC both start from its final checkpoint. The policies are
trained for $300$ epochs, with the world model frozen for the first $15$
and eight epochs of warmup, and the final checkpoint is deployed. To
limit overfitting to the small demonstration sets, every training image
of the world model and of both policies is augmented per camera view
with a random shift of up to $\pm6$ pixels, a brightness gain in
$[0.9,1.1]$, and per-channel gains in $[0.95,1.05]$; evaluation and
deployment images are not augmented. Rotation is not teleoperated, so the
rotation channels of the command are zero.

\subsection{Bun Picking}
\label{sec:real_bun_details}

\smallskip
\qheading{Data and conditioning.}
The $400$ demonstrations contain $101$, $101$, $99$, and $99$ episodes
for green, purple, white, and yellow, respectively. They comprise
$73{,}500$ synchronized observation/action samples at $20$\,Hz,
or approximately $1$\,h of processed demonstration data. The two
camera streams observe the same trajectories and are counted once
when computing duration. Each color is represented by a distinct
one-hot vector in the first four coordinates of the $384$-dimensional
instruction input, with the remaining coordinates set to zero.

\smallskip
\qheading{Evaluation and annotation.}
Tables~\ref{tab:real_bun_ppp_layouts} and
\ref{tab:real_bun_bc_layouts} give all $80$ trial annotations.
A successful trial places the bun of the instructed color in the steamer.
For each failure, the reviewer records one
primary observed mode: approaching or attempting to manipulate the
wrong target, inaccurate end-effector positioning, or inaccurate
gripper operation. Each label summarizes the dominant visible outcome;
multiple errors can contribute to the same trial.

All trials are retained in the denominator, including six failures
that ended with runtime errors. Arm motion stopped with a
\texttt{cartesian\_reflex} error in PPP layout 1/white and 4/white,
and BC layout 1/yellow, 8/white, and 9/purple. BC layout 5/yellow
ended with a gripper timeout. Their primary video-review labels
are retained in the tables; the logged termination event is distinct
from the observed failure mode.

\input{sec/table/real_bun_layouts.tex}

\subsection{Doll Straightening}
\label{sec:real_stitch_details}

\smallskip
\qheading{Task and data.}
A plush doll is dropped on the table in a random pose. The task is to
leave it lying on its back, belly up and centered, with its head toward
the robot base; every demonstration ends in this state. The $150$
demonstrations were recorded in three sessions, with starting poses
spread across the table. In $40$ of them, the doll starts belly up or
face up and is only pushed into place. In the other $110$, it starts
face down or on its side, and the operator grasps and lifts it so that
it rolls over before pushing it into place; $17$ of these retry a
failed flip. The demonstrations comprise $74{,}895$ synchronized
samples at $20$\,Hz, approximately $1$\,h, and last $5.9$ to
$53.0$\,s (median $24.9$\,s). The task has a single instruction, so
the instruction input is one constant one-hot vector and the
instruction contrast loss is disabled ($\lambda_{\mathrm{lang}}=0$).

\smallskip
\qheading{Progress labels.}
Demonstration length here is set mostly by the starting pose: push-only
demonstrations last about $17$\,s on average and those with a flip
about $28$\,s. Normalized time (Eq.~\eqref{eq:ppp_progress_labels})
would therefore give the same doll pose different labels in different
demonstrations. For this task, \method instead regresses absolute
time-to-go,
\begin{equation}
    p_t^\star=-\frac{\max(N_e-1-t,0)}{T_u},
    \qquad s_e=\frac{1}{T_u},
    \label{eq:ppp_stitch_labels}
\end{equation}
with $T_u=123$ model transitions, the span of the median demonstration.
A demonstration of median length thus keeps its normalized-time labels,
and targets range from $-2.14$ at the start of the longest demonstration
to $0$ at the end. The boundary loss treats the last $6$ transitions
($1.2$\,s, $\lfloor(1-\rho_{\mathrm{term}})T_u\rfloor$) as terminal.
All other settings match bun picking. BC has no progress output, so the
labels affect only \method.

In \cref{fig:real_rollouts}, $p_t=p_\theta(E(o_t),\ell)$ is converted
to predicted time-to-go as $-0.2\,T_u p_t$ seconds; the normalized-phase
tasks plot $1+p_t$. Numbered images match curve samples by camera frame ID.

\smallskip
\qheading{Evaluation.}
Each method is evaluated in $12$ trials, all of which are retained in
the denominator. Table~\ref{tab:real_stitch_trials} lists every trial.
The time to success
is averaged over successful trials only, so the two averages in
\cref{tab:real_stitch_results} are taken over different sets of
trials. The success-count difference is two trials ($9/12$ versus
$7/12$), so we treat the success-rate comparison as preliminary.
The individual timings also favor \method beyond the mean:
successful-trial medians are $33$\,s versus $46$\,s, and $9/12$
versus $6/12$ recorded trials finish within $60$\,s.
\Cref{fig:stitch_field} visualizes \method's field on this task.

\input{sec/table/real_stitch_trials.tex}

\subsection{Bimanual Egg Pouring}
\label{sec:real_egg_details}

This experiment tests whether one scalar field can supply multiple
discrete control signals, using two grippers alongside a joint
continuous command for both arms.

\smallskip
\qheading{Task, data, and training.}
Two xArm7 arms grasp a steel bowl containing an egg and a white plate,
respectively. The left arm tilts the bowl over the plate held by the
right arm; the demonstrations then set down and release both objects.
We collect $150$ successful teleoperated episodes, comprising $46{,}619$
synchronized samples at $20$\,Hz. A front camera and one wrist camera
per arm supply three RGB views, encoded with the same shared ViT and
projected to one $384$-dimensional latent. Each arm has six pose
coordinates and a binary gripper command; both translation and rotation
are teleoperated. The action layout is
$(u_{\mathrm L},u_{\mathrm R},c_{\mathrm L},c_{\mathrm R})\in
    \mathbb R^{12}\times\{\mathrm{open},\mathrm{close}\}^2$.
This single-task model uses a constant one-hot instruction and no
instruction contrast loss. Progress supervision uses normalized phase,
as in bun picking. We retain the image augmentation and $100$-epoch
world-model pretraining followed by $300$-epoch policy training described
above, with co-training from epoch $15$; the final checkpoint is deployed.

\smallskip
\qheading{Multiple discrete coordinates.}
For $G$ binary grippers, write the held-command vector as
$\mathbf c^{\mathrm h}=(c^{\mathrm h}_1,\ldots,c^{\mathrm h}_G)$ and
let $\mathbf c^{\mathrm{flip}(j)}$ change only its $j$th entry.
The same scalar query $Q(u,\mathbf c)$ supplies
\begin{equation}
    \begin{aligned}
        g_u & =\nabla_u Q(u_0,\mathbf c^{\mathrm h}),\qquad
        u=u_0+\eta_\phi(\bar z)g_u,                                         \\
        b_j & =Q(u_0,\mathbf c^{\mathrm{flip}(j)})
        -Q(u_0,\mathbf c^{\mathrm h}),                                      \\
        c_j & =\begin{cases}
                   c^{\mathrm{flip}(j)}_j, & b_j>\varepsilon_{\mathrm g,j}, \\
                   c^{\mathrm h}_j,        & \text{otherwise.}
               \end{cases}
    \end{aligned}
    \label{eq:ppp_multigrip}
\end{equation}
Here $u_0=0\in\mathbb R^{6G}$ and
$\bar z=f(z,(u_0,\mathbf c^{\mathrm h}))$.
For two arms, one gradient covers all $12$ pose coordinates and one
positive scalar mobility scales it. Each gripper comparison holds the
other command fixed at the same reference. The resulting switches are
applied together, without sequential updates or scoring joint flips;
thus the discrete query count grows linearly with $G$. The shared
world model and field still condition on the full scene and both arms.

Switching-region labels (\cref{eq:ppp_switch_region}) are computed from
each gripper's own next expert switch. The loss in
\cref{eq:ppp_balanced_gripper_loss} pools sample--gripper pairs and
balances switching versus holding across this pool. Repeated-frame
contexts retain both held commands, and this loss still updates only
the potential. Pose supervision uses the joint command, with the
mean-squared term averaged over $12$ coordinates.

\smallskip
\qheading{Deployment.}
The controller replans both arms at $20$\,Hz and uses
$\varepsilon_{\mathrm g,j}=0$ for both grippers, as in the Franka
experiments. The policy completes the egg transfer in one of five attempts ($20\%$).
\Cref{fig:real_rollouts}(c) uses a successful checkpoint-$300$ rollout,
with recorded camera frames synchronized to the unsmoothed prediction
$1+p_\theta(E(o_t),\ell)$ by camera frame ID.

\input{sec/X_task_field}

%% file: sec/table/libero_trend.tex
\begin{table}[!ht]
    \centering
    \caption{Average LIBERO success rates (\%) across training epochs,
    evaluated on the same $2{,}000$ initial states at each checkpoint.
    The two PPP rows share a checkpoint at each epoch and differ only
    in whether gripper settling is enabled during deployment
    (Sec.~\ref{sec:exp_gripper_settling}); all other rows use no
    deployment adjustment. No $\mathcal L_p$: \method-GH without
    progress losses (\cref{sec:ppp_nolp}).
    Bold denotes the best result per column.}
    \label{tab:libero_trend}
    \small
    \setlength{\tabcolsep}{2.1pt}
    \begin{tabular}{@{}lrrrrrrrr@{}}
        \toprule
                         & \multicolumn{8}{c}{Training epoch}                                                                                                                 \\
        \cmidrule(l){2-9}
        Method           & 20                                 & 30            & 40            & 50            & 70            & 80            & 100           & 120           \\
        \midrule
        BC               & 20.5                               & 20.6          & 26.3          & 27.4          & 33.7          & 35.3          & 37.8          & 38.1          \\
        BC+Progress      & 20.9                               & 24.0          & 26.2          & 30.4          & 34.1          & 34.0          & 39.4          & 38.8          \\
        Direct-AP        & 26.5                               & 32.2          & 31.5          & 37.5          & 39.4          & 41.6          & 40.5          & 44.8          \\
        No $\mathcal L_p$ & 26.2                               & 26.6          & 31.6          & 35.2          & 37.7          & 36.9          & 37.8          & 41.8          \\
        PPP-GH           & 29.7                               & 32.8          & 34.8          & 36.3          & 41.9          & 38.9          & 45.1          & 48.0          \\
        PPP w/o settling & 28.9                               & 39.1          & 48.1          & 43.4          & 49.0          & 52.2          & 49.5          & 51.5          \\
        PPP              & \textbf{31.5}                      & \textbf{41.0} & \textbf{50.0} & \textbf{45.7} & \textbf{50.0} & \textbf{53.6} & \textbf{50.6} & \textbf{54.3} \\
        \bottomrule
    \end{tabular}
\end{table}

%% file: sec/X_training_schedule.tex
\subsection{Sensitivity to Policy Training Schedule}
\label{sec:ppp_schedule_sensitivity}

\smallskip
\qheading{Original protocol.}
The main LIBERO comparison retains the protocol of \cref{sec:ppp_optimization_details}: a common pretrained world model, training data, learning-rate schedule, and world model unfreezing epoch.
During the first ten policy epochs, \method and \method-GH calibrate their potentials with $\mathcal L_p$, whereas BC trains its action head and No $\mathcal L_p$ trains its gradient readout by action imitation.
From epoch $10$, all jointly train the world model; action supervision begins at this point for \method and \method-GH.
This compares complete training procedures, including their early objectives, but does not isolate policy structure from the conditions at the onset of action learning.

\smallskip
\qheading{Branch protocol.}
We compare BC-branch, No $\mathcal L_p$-branch, \method-GH-branch, and \method-branch.
The first three use a learned gripper-switching head and inherit the same epoch-$10$ world model state used to initialize \method-GH's joint stage.
\method-branch uses field-based switching and inherits the epoch-$10$ checkpoint used to initialize full \method's joint stage, from a separate initial-stage run.
In each case, the encoder and dynamics retain their pretrained weights, which were frozen during the initial stage, and inherit their source checkpoint's BatchNorm running statistics.
BC's action and switching heads use their original epoch-$0$ initialization.
For all field variants, the potential is reset to its original epoch-$0$ weights; the gain and, where present, switching head retain their initial weights, having received no updates in the initial stage.
Thus no progress-trained potential is transferred to a branch run.
The field variants initialize their query-normalization statistics from the inherited prediction-normalization statistics, as in their original runs.

All four runs start with empty AdamW moment states, restore their source checkpoint's random state, and retain the epoch-$10$ learning-rate schedule and epoch-indexed data sampling.
The first update therefore uses the near-peak learning rates, without repeating warmup, and jointly trains the policy and world model.
The data, batch size, loss definitions, and deployment rules remain those of each method.
\method-GH-branch applies $\mathcal L_p$ and action supervision together from this first update; No $\mathcal L_p$-branch differs only by setting all progress-loss weights to zero.
\method-branch likewise applies progress, pose, and field-based gripper supervision from the first update.
Its two deployment variants evaluate the same checkpoints with and without gripper settling, using the original \method deployment settings (\cref{sec:ppp_gripper_settling}).
Epoch labels retain the original counter: checkpoint $e$ contains $e-10$ epochs of action supervision in every branch run, versus $e$ for original BC and No $\mathcal L_p$.
Original \method and \method-GH also receive $e-10$ action-supervised epochs, preceded by ten epochs of progress calibration.
Equal checkpoint labels therefore match action-update budgets within the branch comparison, but not across every original and branch pair.

\input{sec/table/branch.tex}

\smallskip
\qheading{Results and scope.}
\Cref{tab:branch_trend} reports the checkpoint trajectories, and \cref{tab:branch_epoch120} gives the final per-suite breakdown, using the same evaluation initial states as the main comparison.
Relative to their original runs (\cref{tab:libero_trend}), BC and No $\mathcal L_p$ both improve under the branch protocol, including late in training.
BC-branch is therefore a stronger direct-policy baseline than the original BC.
No $\mathcal L_p$-branch leads BC-branch at several intermediate checkpoints, while their final average success rates are close; this does not establish equivalence.
Adding progress supervision to the branch gradient policy with a learned gripper head does not yield a consistent success-rate benefit across checkpoints.
No $\mathcal L_p$-branch removes progress losses from \method-GH, not from full \method.
These results support learning an effective continuous readout without progress labels; the evidence for value calibration and progress monitoring is separate (\cref{sec:exp_rq3,sec:ppp_rollout_progress}).

Full \method retains higher final average success than BC-branch and No $\mathcal L_p$-branch under both its original and branch protocols, including without settling (\cref{tab:libero_epoch120,tab:branch_epoch120}).
\method-branch also improves over \method-GH-branch and its own counterpart under the original protocol in final average success.
Settling improves the final average for \method-branch, but its effect varies across checkpoints.
The comparison between \method-branch and \method-GH-branch matches the training schedule but inherits normalization statistics and random states from different source runs, so it does not isolate the gripper readout alone.
The schedule interventions change multiple training conditions together, including action-learning onset, learning-rate warmup, normalization state, potential initialization, optimizer state, and data order, so the performance changes cannot be attributed to any one factor alone.

%% file: sec/table/branch.tex
\begin{table}[!ht]
    \centering
    \caption{Branch-schedule LIBERO success rates (\%), averaged over the same $2{,}000$ evaluation initial states at each checkpoint.
        Epoch labels retain the original counter; action supervision starts at epoch $10$ in every row.
        The two \method-branch rows use field-based gripper switching, share a checkpoint at each epoch, and differ only in settling at deployment; the other methods use a learned gripper-switching head.
        No $\mathcal L_p$-branch is \method-GH-branch without progress losses (\cref{sec:ppp_schedule_sensitivity}).}
    \label{tab:branch_trend}
    \small
    \setlength{\tabcolsep}{2.1pt}
    \begin{tabular}{@{}lrrrrrrrr@{}}
        \toprule
                          & \multicolumn{8}{c}{Training epoch}                                                  \\
        \cmidrule(l){2-9}
        Method            & 20                                 & 30   & 40   & 50   & 70   & 80   & 100  & 120  \\
        \midrule
        BC-branch                & 27.4                        & 35.8 & 40.5 & 34.6 & 43.2 & 40.2 & 47.0 & 46.8 \\
        No $\mathcal L_p$-branch & 27.2                        & 30.8 & 39.3 & 42.8 & 49.8 & 46.6 & 50.9 & 46.3 \\
        \method-GH-branch        & 32.0                        & 34.7 & 38.0 & 41.0 & 38.9 & 44.0 & 45.7 & 44.4 \\
        \midrule
        \shortstack[l]{\method-branch\\w/o settling}
                                & 37.0                        & 42.8 & 49.2 & 47.8 & 49.7 & 52.5 & 55.1 & 55.0 \\
        \method-branch           & 36.8                        & 44.0 & 48.8 & 46.9 & 51.0 & 55.5 & 55.5 & 56.3 \\
        \bottomrule
    \end{tabular}
\end{table}

\begin{table}[!ht]
    \centering
    \caption{Branch-schedule LIBERO success rates (\%) by suite at epoch $120$, with $500$ evaluation rollouts per suite on the same initial states as \cref{tab:libero_epoch120}.
        The two \method-branch rows share a checkpoint and differ only in settling at deployment.}
    \label{tab:branch_epoch120}
    \small
    \setlength{\tabcolsep}{4pt}
    \begin{tabular}{@{}lrrrrr@{}}
        \toprule
        Method            & Spatial & Object & Goal & Long & Avg. \\
        \midrule
        BC-branch                & 46.2 & 54.2 & 57.0 & 30.0 & 46.8 \\
        No $\mathcal L_p$-branch & 54.4 & 43.2 & 51.0 & 36.6 & 46.3 \\
        \method-GH-branch        & 46.8 & 55.2 & 47.8 & 27.8 & 44.4 \\
        \midrule
        \shortstack[l]{\method-branch\\w/o settling}
                                & 45.6 & 88.0 & 48.6 & 37.6 & 55.0 \\
        \method-branch           & 50.2 & 86.2 & 49.8 & 39.0 & 56.3 \\
        \bottomrule
    \end{tabular}
\end{table}

%% file: sec/compute_cost.tex
\subsection{Parameter Counts and Arithmetic Cost}
\label{sec:ppp_compute_cost}

\input{sec/table/compute_cost.tex}

\smallskip
\qheading{Counting convention.}
\Cref{tab:ppp_compute_cost} reports registered model parameters and
executed matrix-multiplication and convolution FLOPs for the LIBERO
models, counting a multiply--accumulate as two operations.
We trace the deployed planner and joint-training computation with
PyTorch~2.11's operator-level FLOP counter, including the action-gradient
backward pass and the outer loss backward pass during training.
Attention uses its mathematical implementation so that its matrix
products are counted explicitly.
These are dense-arithmetic counts: elementwise operations, activations,
normalization, softmax, reductions, data movement, optimizer updates,
and distributed communication are excluded.
The counts are obtained in FP32 on CPU without timing measurements and
do not depend on GPU load or clock speed.

\smallskip
\qheading{Deployment.}
Each decision has batch size one, two new $128\times128$ RGB images,
and three cached observation latents.
The deployed planner encodes only the newest pair of images; it reuses
the historical latents.
The readout count includes the reference dynamics prediction, the pose
readout, and the gripper decision.
For \method it includes one action-gradient evaluation and both gripper
potential evaluations as executed by the implementation; for BC it
includes the direct pose and switching heads.
Instruction embeddings are computed before control, so the shared text
encoder is excluded from both parameter and per-decision FLOP counts.
Optional progress logging is also excluded.
Shared visual-encoder weights are counted once despite processing both
views. Parameter totals include all registered network weights,
including the ViT pooler whose output the downstream model does not use.

\smallskip
\qheading{Joint training.}
Training counts include all configured losses and their backward passes
after world model unfreezing, using seeded synthetic tensors with the
training input shapes.
The per-sample count includes context encoding and the next-frame
world-model target. Progress-supervised methods additionally encode the
later frame used for progress pairs.
It also includes query-normalization passes and the original and
repeated-frame gripper contexts where applicable.
Thus, the training comparison measures each complete objective; its
difference includes these additional evaluations as well as differentiation
through the action gradient.
Under this counting convention, \method uses $1.09\times$ BC's
per-decision arithmetic and $1.28\times$ its per-sample training
arithmetic. Relative to BC+Progress, which also encodes progress pairs,
the training ratio is $1.02\times$.

%% file: sec/table/compute_cost.tex
\begin{table*}[t]
    \centering
    \caption{Network size and dense-arithmetic cost of the LIBERO implementations.
    Parameters are in millions; costs are in GFLOPs, with two operations per multiply--accumulate.
    Inference includes the new two-camera observation and the complete action readout, with cached history latents.
    Training includes all losses and their backward passes, normalized per sample; see \cref{sec:ppp_compute_cost} for the counting scope.}
    \label{tab:ppp_compute_cost}
    \small
    \setlength{\tabcolsep}{7pt}
    \begin{tabular}{@{}lrrrrr@{}}
        \toprule
        & & \multicolumn{3}{c}{Inference per decision} & Training per sample \\
        \cmidrule(lr){3-5}
        Method & Parameters (M) & Encoder & Readout & Total & Forward + backward \\
        \midrule
        BC & 61.816 & 5.757 & 0.181 & 5.938 & 69.805 \\
        BC+Progress & 61.817 & 5.757 & 0.181 & 5.938 & 87.197 \\
        \method-GH & 61.976 & 5.757 & 0.357 & 6.115 & 88.188 \\
        \method & 61.712 & 5.757 & 0.718 & 6.475 & 89.079 \\
        \bottomrule
    \end{tabular}
\end{table*}

%% file: sec/table/real_bun_layouts.tex
\begin{table}[t]
    \centering
    \caption{Bun picking: PPP annotations by layout and instructed
    color. Each cell is one trial; failure labels indicate the primary
    observed failure mode: Target = wrong target, Pose = pose
    inaccuracy, and Gripper = gripper inaccuracy.}
    \label{tab:real_bun_ppp_layouts}
    \small
    \setlength{\tabcolsep}{3pt}
    \renewcommand{\arraystretch}{1.25}
    \begin{tabular*}{\linewidth}{@{\extracolsep{\fill}}ccccc@{}}
        \toprule
        Layout & green & purple & white & yellow \\
        \midrule
        0 & \textbf{Success} & \textbf{Success} & Pose & Target \\
        1 & \textbf{Success} & \textbf{Success} & Target & \textbf{Success} \\
        2 & Pose & \textbf{Success} & Pose & \textbf{Success} \\
        3 & Target & \textbf{Success} & Pose & \textbf{Success} \\
        4 & \textbf{Success} & \textbf{Success} & Pose & \textbf{Success} \\
        5 & \textbf{Success} & \textbf{Success} & Target & \textbf{Success} \\
        6 & \textbf{Success} & Target & Pose & Pose \\
        7 & \textbf{Success} & \textbf{Success} & Target & Target \\
        8 & Target & \textbf{Success} & \textbf{Success} & Target \\
        9 & Pose & Gripper & \textbf{Success} & Target \\
        \bottomrule
    \end{tabular*}
\end{table}

\begin{table}[!ht]
    \centering
    \caption{Bun picking: BC annotations by layout and instructed
    color. Each cell is one trial; failure labels indicate the primary
    observed failure mode: Target = wrong target, Pose = pose
    inaccuracy, and Gripper = gripper inaccuracy.}
    \label{tab:real_bun_bc_layouts}
    \small
    \setlength{\tabcolsep}{3pt}
    \renewcommand{\arraystretch}{1.25}
    \begin{tabular*}{\linewidth}{@{\extracolsep{\fill}}ccccc@{}}
        \toprule
        Layout & green & purple & white & yellow \\
        \midrule
        0 & Target & Pose & Pose & \textbf{Success} \\
        1 & \textbf{Success} & Target & Gripper & Pose \\
        2 & Target & Pose & \textbf{Success} & Target \\
        3 & Gripper & Gripper & Target & Target \\
        4 & Target & \textbf{Success} & Pose & Target \\
        5 & Gripper & Gripper & Target & Gripper \\
        6 & Pose & Pose & Pose & Pose \\
        7 & Pose & Gripper & Pose & Pose \\
        8 & Target & Gripper & Target & Target \\
        9 & Gripper & Pose & Target & \textbf{Success} \\
        \bottomrule
    \end{tabular*}
\end{table}

%% file: sec/table/real_stitch_trials.tex
\begin{table}[t]
    \centering
    \caption{Doll straightening: all $12$ trials per method, with time
    to success for successful trials. The final row gives success rates
    and mean times over successes.}
    \label{tab:real_stitch_trials}
    \small
    \setlength{\tabcolsep}{3pt}
    \renewcommand{\arraystretch}{1.25}
    \begin{tabular*}{\linewidth}{@{\extracolsep{\fill}}ccrcr@{}}
        \toprule
        & \multicolumn{2}{c}{PPP} & \multicolumn{2}{c}{BC} \\
        \cmidrule(lr){2-3}\cmidrule(l){4-5}
        Trial & Outcome & Time (s) & Outcome & Time (s) \\
        \midrule
        0  & \textbf{Success} & 13 & \textbf{Success} & 8 \\
        1  & Failure          & -- & Failure          & -- \\
        2  & \textbf{Success} & 55 & Failure          & -- \\
        3  & \textbf{Success} & 33 & \textbf{Success} & 52 \\
        4  & \textbf{Success} & 19 & \textbf{Success} & 51 \\
        5  & Failure          & -- & Failure          & -- \\
        6  & \textbf{Success} & 18 & \textbf{Success} & 14 \\
        7  & Failure          & -- & Failure          & -- \\
        8  & \textbf{Success} & 43 & \textbf{Success} & 46 \\
        9  & \textbf{Success} & 43 & \textbf{Success} & 115 \\
        10 & \textbf{Success} & 10 & \textbf{Success} & 36 \\
        11 & \textbf{Success} & 52 & Failure          & -- \\
        \midrule
        Total & 9/12 (75.0\%) & 31.8 & 7/12 (58.3\%) & 46.0 \\
        \bottomrule
    \end{tabular*}
\end{table}

%% file: sec/X_task_field.tex
\section{Task-Field Analysis}
\label{sec:ppp_task_field_analysis}

\begin{figure*}[t]
    \centering
    \includegraphics[width=\linewidth]{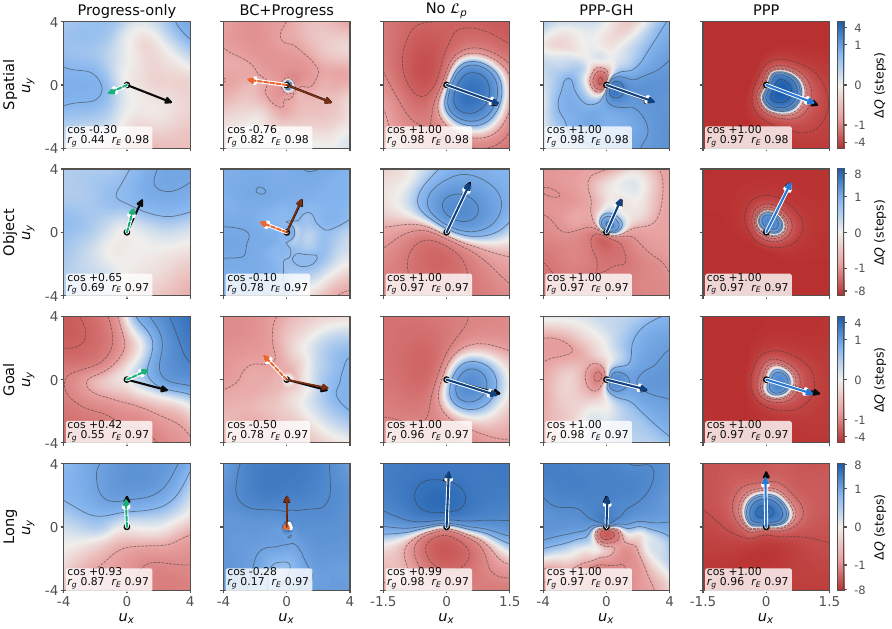}
    \caption{Action-space slices of $Q$ in the $(u_x,u_y)$ plane at one rule-selected state per suite (rows), for the progress-only field, BC+Progress, No $\mathcal L_p$, \method-GH, and \method (columns; epoch $10$, then $120$).
        No $\mathcal L_p$ is trained without progress labels, so its scale is free; its $\Delta Q$ is multiplied by one constant ($0.078$), chosen so that its median $\|g_a\|$ over the diagnostic states equals \method-GH's, which leaves shapes, directions and signs unchanged.
        Black: the expert command.
        Solid colour: the deployed command (No $\mathcal L_p$, \method-GH and \method).
        Dashed colour: the gradient direction scaled to the expert's norm (progress-only and BC+Progress; BC+Progress's own BC command in dark brown).
        cos is the six-dimensional cosine between $g_a$ and $\bar u_E$.
        Colour: $\Delta Q=Q(u)-Q(u_0)$ in control steps, symmetric-log scale shared within a row; for \method, the elevated $Q(u_0)$ is associated with gripper-margin training (\cref{sec:ppp_finite_steps}).
        The No $\mathcal L_p$ and \method columns are zoomed to $[-1.5,1.5]^2$, and their arrows are shortened by a common factor (\cref{sec:ppp_field_vis}).}
    \label{fig:slices_all}
\end{figure*}

\subsection{Diagnostic Protocol}
\label{sec:ppp_gradient_protocol}

\smallskip
\qheading{States.}
For each of the $40$ benchmark tasks we draw ten demonstrations with a fixed seed.
We take every second model-grid anchor in the training anchor range, giving $7{,}849$ states from $400$ trajectories.
States are built by the training sampler's own code at phase zero.
All LIBERO demonstrations are used for training, so these are training states.

\smallskip
\qheading{Queries.}
Every quantity is computed with the deployed query in evaluation mode, as $Q_{z,\ell}(u,c_{\mathrm h})=p_\theta(f(Z_\tau,[\bar A_\tau,\operatorname{tile}_k(u,c_{\mathrm h})]),\ell)$ (\cref{eq:ppp_explicit_query}).
The context consists of three frames four ticks apart, the history's pose channels are at the corpus median, its gripper channels are the recorded commands, and $c_{\mathrm h}$ is the held gripper command.
$g_a$ is one reverse-mode derivative of $Q$ with respect to $u$ at $u_0=0$.
The expert target is the mean of the four normalized expert commands that the query's held chunk replaces, exactly the target of \cref{eq:action_supervision}.
For BC+Progress, the potential is the trunk's progress channel, composed with its own world model in the same query.

\smallskip
\qheading{Validation.}
Autograd gradients agree with central finite differences of $Q$ (step sizes $10^{-2}$ and $3\times10^{-3}$; median relative error below $10^{-3}$ at the better step).
For a field run, the command computed here equals the deployed planner's output for the same context to a relative $2\times10^{-6}$, and for BC+Progress the planner's BC command matches as well.

\smallskip
\qheading{Statistics.}
Neighbouring anchors of a demonstration are correlated, so each statistic is a mean of per-trajectory means.
Progress correlation is computed within each trajectory and then averaged; this differs from the $130$-task training-log diagnostic, which pools sampled predictions within each task before averaging correlations (\cref{sec:ppp_nolp}).
Intervals are percentile bootstraps over the $400$ trajectories, and paired differences are tested with a sign-flip test over the $40$ per-task differences.
Directional metrics exclude states whose expert command is near zero ($\|\bar u_E\|<0.25$, about a tenth of the median; $0.23\%$ of states).

\input{sec/table/tf_alignment.tex}

\smallskip
\phantomsection
\qheading{Frozen-world-model runs.}
\label{sec:ppp_frozen_runs}
Two runs hold the dynamics fixed.
Both continue from \method-GH's own epoch-$10$ checkpoint, the end of its progress-only stage (progress-only (a) in \cref{tab:tf_alignment}), with \method-GH's data, batch size, optimizer state, and learning-rate schedule.
Both keep pretrained encoder and dynamics weights fixed and omit the world model loss.
The progress-only run continues the progress losses alone to epoch $120$; its gain and switching head are never trained, so it is measured only through $g_a$.
The frozen-WM run adds \method-GH's action loss and switching head from epoch $10$.
Its action loss can therefore change only the potential and the mobility, not $J_f$.
A third run deploys the progress-only field: it freezes the epoch-$120$ progress-only potential and trains only the scalar gain and a switching head, with \method-GH's action and switching losses, for $20$ epochs.
Its command direction is thus exactly the progress-only gradient.

\subsection{Results in Detail}
\label{sec:ppp_field_results}

\smallskip
\qheading{Values and directions.}
\Cref{tab:tf_alignment} reports value fit and gradient alignment across training;
brackets below give $95\%$ trajectory-bootstrap intervals for paired differences.
The two independent progress-only fields have mean cosine $0.25$ with the expert,
positive on three quarters of states; BC+Progress's gradient drifts toward
orthogonality as its value fit improves.
At epoch $120$, \method and \method-GH reach cosine $0.995$.
Their paired gains over progress-only (b) and BC+Progress, respectively, are
$0.749$ [0.738, 0.760] and $0.915$ [0.903, 0.927], positive on all $40$ tasks
with $p_{\mathrm{task}}<10^{-4}$ in both comparisons.
The \method $-$ \method-GH difference is $-0.0001$ ($p_{\mathrm{task}}=0.61$).

Continuing progress-only training to epoch $120$ lowers value error
($0.069\rightarrow0.051$) but changes cosine by $-0.024$ [$-0.034$, $-0.014$]
(lower on $25/40$ tasks, $p_{\mathrm{task}}=0.037$).
With the world model frozen, \method-GH's losses raise cosine to $0.947$,
a gain of $0.722$ [0.708, 0.736] over the frozen progress-only field.
Joint training adds $0.048$ [0.045, 0.052]; both gains hold on all $40$ tasks
with $p_{\mathrm{task}}<10^{-4}$.
The jointly trained model reaches $0.95$ within ten action-supervised epochs,
a level the frozen run approaches only at epoch $120$.
The orthogonal-energy column shows what a scalar gain cannot express:
about $80\%$ of expert-command energy for progress-only fields, against $1\%$ for \method.
Neighbouring states two grid steps apart have gradient-direction similarity
$0.78$ for both \method and the expert commands.

Without progress losses, epoch-$120$ cosine reaches $0.996$, compared with
\method-GH's $0.995$ (\method-GH $-$ No $\mathcal L_p$: $-0.0008$,
positive on $10/40$ tasks, $p_{\mathrm{task}}=0.0004$).
Its per-trajectory progress correlation is $0.58$, illustrating the separate
role of value calibration.

\smallskip
\qheading{Closed loop.}
Both deployable frozen-world-model runs use the protocol of \cref{tab:libero_epoch120}.
With \method-GH's losses, success is $5.6$, $5.1$, $4.9$, and $4.4\%$ at epochs $70$, $80$, $100$, and $120$ ($5.0\%$ pooled, against $43.5\%$ for \method-GH; lower on $39$ of the $40$ tasks, $p_{\mathrm{task}}<10^{-5}$).
At epoch $120$, median hand paths are similar ($0.96$ versus $0.91$\,m for \method-GH), but the gripper closes in only $62\%$ of episodes (\method-GH: $84\%$), with success in $6.9\%$ of episodes with a close (\method-GH: $54.1\%$).
The progress-only field with a trained gain succeeds in $0$ and $1$ of $2{,}000$ episodes after $10$ and $20$ epochs of gain training.
\Cref{tab:progress_vs_control} (progress, gain) compares the latter with the frozen-WM epoch-$120$ checkpoint ($4.4\%$ success).
Its median command norm is $0.24$, against $1.71$ for \method-GH, and the hand travels a median $0.33$\,m.

\begin{figure*}[t]
    \centering
    \includegraphics[width=\linewidth]{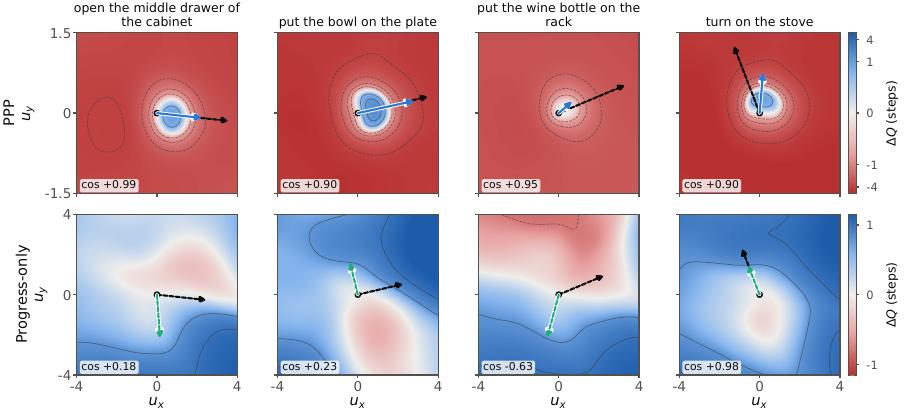}
    \caption{One LIBERO-Goal scene, at the planner's episode-start state, under four of its ten instructions.
        Top: \method (epoch $120$) and its command (blue), zoomed to $[-1.5,1.5]^2$ with arrows shortened by a common factor.
        Bottom: the progress-only field (epoch $10$) and its gradient direction (green, dashed).
        Black dashed: the instructed task's mean initial hand motion over its $50$ demonstrations, drawn as a direction.
        cos is computed in three-dimensional translation.}
    \label{fig:instruction_goal}
\end{figure*}

\input{sec/table/tf_linearity.tex}

\smallskip
\phantomsection
\qheading{Finite query values and their reference.}
\label{sec:ppp_finite_steps}
\Cref{tab:tf_alignment} reports scalar increments relative to the current observation: $Q_0=Q(u_0)-p_\theta(z,\ell)$ and $Q_E=Q(\bar u_E)-p_\theta(z,\ell)$, summarized by their medians.
In \method, query values are supervised differently from recorded-state potentials: pose imitation trains the derivative at $u_0$, and $\mathcal L_{\mathrm{grip}}$ trains the difference between $Q(u_0,\mathrm{open})$ and $Q(u_0,\mathrm{close})$, updating only $p_\theta$.
The logistic gripper loss increases the correct switch/hold margin without fixing absolute query values; reference queries lack direct progress-value targets.
The reference $u_0=0$ is the normalized corpus median command, not necessarily a physical no-op, and is used for readout rather than execution.

Under every calibrated field, the demonstrated next state lies $1.1$--$1.2$ control steps above the current one.
\method-GH stays on this scale throughout training ($Q_0=0.40$ and $Q_E=1.14$ at epoch $120$, and $1.14$ for its deployed command).
For \method, $Q_E$ stays positive ($0.57$ at epoch $120$; $0.55$ for its deployed command), while $Q_0$ rises from $0.94$ at epoch $20$ to $6.92$ at epoch $120$, higher than \method-GH's on all $40$ tasks.
These query-latent values differ from the observed increment $p_\theta(E(o_{t+k}),\ell)-p_\theta(E(o_t),\ell)$, which the simulator measures as $0.00$ control steps for the reference command and $0.99$ for \method's deployed command (\cref{sec:ppp_true_dynamics}).
The median per-state difference $Q(\bar u_E)-Q(u_0)$ is $-5.8$ control steps, comparing the expert query with the elevated reference rather than a decrease along the observed trajectory; it is not the difference of the $Q_0$ and $Q_E$ medians.

Reference values are highest on hold states: $+11.4$ control steps with the hand open and no switch within eight steps, $+5.4$ with the hand closed under the same condition, and $-4.7$ within eight steps before an opening (\method-GH: $+0.8$, $+0.1$, $-0.1$).
Continuing \method from epoch $40$ with a label-smoothed $\mathcal L_{\mathrm{grip}}$, whose minimizer is a margin of two control steps, reduces $Q_0$ at epoch $120$ to $3.66$, with similar measured closed-loop success ($50.2\%$ against $50.5\%$ pooled over epochs $70$--$120$, $p_{\mathrm{task}}=0.76$).
This intervention supports an association with gripper-margin training; the logistic objective alone does not require the absolute reference value to rise.
A positive local derivative can coexist with lower finite-command values outside the Taylor neighborhood below.

\begin{figure*}[t]
    \centering
    \includegraphics[width=\linewidth]{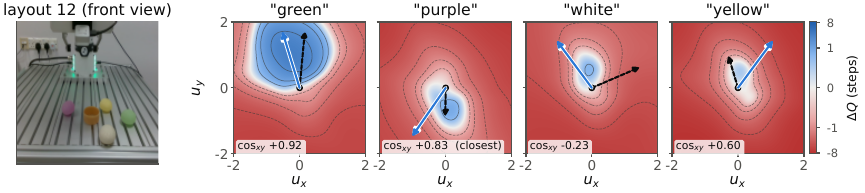}
    \caption{Real robot: one shared layout at the episode-start state, under the four color instructions (\method, epoch $300$, $(u_x,u_y)$ table plane in the robot base frame, normalized units, zoomed to $[-2,2]^2$).
        Blue: the direction of \method's command (drawn at a fixed length, since commands are small at the first decision).
        Black dashed: the direction of the same layout's demonstration for that color over its first $2$\,s.
        ``closest'': the command is closer to this color's demonstration than to the other three.}
    \label{fig:instruction_bun}
\end{figure*}

\input{sec/table/tf_instruction.tex}

\begin{figure*}[!t]
    \centering
    \includegraphics[width=0.9\linewidth]{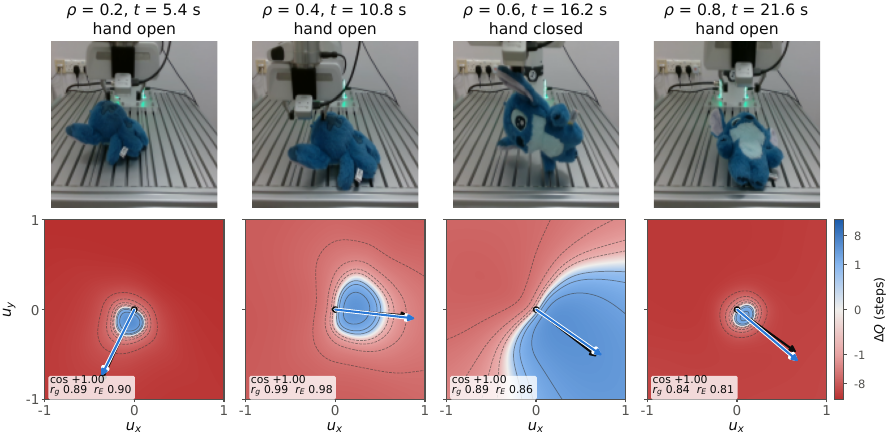}
    \caption{Real robot, doll straightening (single instruction): \method's field (epoch $300$) at four phases of one seeded single-grasp demonstration (training states).
        Top: the front view at each state.
        Bottom: $(u_x,u_y)$ slices in the robot base frame, normalized units, zoomed to $[-1,1]^2$.
        Black: the expert command; blue: \method's deployed command (projections, shortened by a common factor in each panel).
        cos is the six-dimensional cosine between $g_a$ and $\bar u_E$.
        Colour: $\Delta Q$ in time-to-go control steps; shared symmetric-log scale.}
    \label{fig:stitch_field}
\end{figure*}

\smallskip
\qheading{Local linearity.}
\Cref{tab:tf_linearity} gives the data of \cref{fig:local_linearity}.
For \method and \method-GH, $Q$ is close to linear up to $r\approx0.1$ (about $4\%$ of a median expert command).
At $r=1$, \method's median ratio $-0.27$ indicates a finite query below $Q(a_0)$ along $\hat g_a$, while its median cosine $-0.58$ compares gradients at $a_0$ and $a_0+\hat u_E$.
These fixed-observation diagnostics measure candidate-action curvature; the supervised readout recomputes the reference gradient (expert alignment $0.995$) after each observation.

With the world model frozen, the aligned field stays much closer to linear (ratio $0.81$ and gradient cosine $0.77$ at $r=1$), and its gradient is about nine times less steep (\cref{tab:tf_alignment}).
In these runs, jointly trained fields have a shorter linear range.
The label-free field has the same local profile as \method-GH (ratio $0.85$ at $r=0.3$ and $0.23$ at $r=1$, where its gradient reverses, cos $-0.66$).
BC+Progress's progress channel is trained on values, and its $Q$ varies irregularly even at small $r$.
The progress-only field is the most linear, but its gradient is misaligned.

The $Q{-}p$ rows show \method-GH's $Q$ rising along the gradient to $2.1$ control steps at $r=0.3$, then returning to $1.1$ at the expert's norm and beyond: small commands yield larger increments than expert-sized commands. With the world model frozen, it rises monotonically ($0.29$ to $1.50$).
\method's $Q$ starts from its elevated reference value and falls to $1.1$ by $r=1$.
Comparing two commands of equal size removes the level at $a_0$.
At $r=2$, about the expert's norm, $Q$ prefers the expert direction to its opposite on $98\%$ of states for \method-GH and the frozen-WM field, $90\%$ for BC+Progress, $83\%$ for the progress-only field, and $79\%$ for \method (pref.\ rows).
Reference values and directional preference vary separately: world model normalization with an added query prediction loss retains the high $Q_0$ but raises expert-direction preference to $93\%$ (\cref{sec:ppp_query_normalization}).
Away from $a_0$, jointly trained $Q$ ranks directions and, for \method-GH, matches the demonstrated step at the expert's norm, but does not calibrate progress for arbitrary finite commands.

\smallskip
\qheading{Instruction conditioning.}
\Cref{tab:tf_instruction} tests instruction dependence before task-specific motion, using the planner's episode-start state: the first frame in every context slot and padded history.
Robot queries use $25$ shared layouts recorded from the home pose, each with one demonstration per color; LIBERO-Goal uses ten demonstrations per task across ten tasks sharing one initial layout.
Both \method and BC turn their commands toward the instructed target about twice as often as chance on the robot, and about three times as often as chance on LIBERO-Goal.
Progress gradients without action supervision are at chance, also after $120$ epochs of progress training.
With the world model frozen, \method-GH's command reaches the instructed target twice as often as chance, less often than the co-trained fields.

\begin{table*}[t]
    \centering
    \caption{Predicted progress along the $2{,}000$ closed-loop rollouts of each epoch-$120$ model.
        $\rho_t$: median per-rollout Spearman correlation of $p_t$ with time, on successful (S) and failed (F) rollouts.
        End: median $p$ at the last tick (No $\mathcal L_p$: after the affine fit, marked $^\ast$).
        AUROC for success from the value at the last tick, and for eventual success from the value at $5$\,s, with $95\%$ task-bootstrap intervals.}
    \label{tab:ppp_rollout_progress}
    \small
    \setlength{\tabcolsep}{5pt}
    \begin{tabular}{@{}lrrrrrrr@{}}
        \toprule
                          &       & \multicolumn{2}{c}{$\rho_t$} & \multicolumn{2}{c}{End} & \multicolumn{2}{c}{AUROC}                                                                            \\
        \cmidrule(lr){3-4}\cmidrule(lr){5-6}\cmidrule(l){7-8}
        Scalar            & Succ. & S                            & F                       & S                         & F            & end                         & $5$\,s                      \\
        \midrule
        \method           & 51.5  & 1.00                         & 0.67                    & $-0.03$                   & $-0.45$      & 0.84 {\scriptsize[.78,.90]} & 0.78 {\scriptsize[.71,.84]} \\
        BC+Progress       & 38.8  & 1.00                         & 0.69                    & $-0.04$                   & $-0.48$      & 0.89 {\scriptsize[.84,.93]} & 0.75 {\scriptsize[.69,.81]} \\
        No $\mathcal L_p$ & 41.8  & 0.68                         & 0.30                    & $-0.38^\ast$              & $-0.50^\ast$ & 0.73 {\scriptsize[.66,.79]} & 0.56 {\scriptsize[.49,.63]} \\
        \bottomrule
    \end{tabular}
\end{table*}

\subsection{The Field Under the True Dynamics}
\label{sec:ppp_true_dynamics}

\input{sec/table/tf_realized.tex}

We now evaluate the same field on simulator-reached states instead of the control query's predicted latents.

\smallskip
\qheading{Protocol.}
From a restored simulator state, each candidate command is held for the world model's horizon (one control step, four ticks). The field's own encoder and potential score the reached frame: $Q_{\mathrm{real}}(a)=p_\theta(E(o^{\mathrm{sim}}_{t+k}(a)),\ell)$.
The candidates are:
the reference command $a_0$ and its gripper alternative;
$a_0\pm\epsilon e_i$ on the six pose axes ($\epsilon\in\{0.1,0.3,1\}$);
$a_0\pm r\hat d$ for $r\in\{0.3,1,2.3\}$, along the $\hat g_a$ of \method and \method-GH, the expert direction, and two random directions;
and each field's deployed command.
All fields score the same executed branches, also rendered after a second control step.
Restoration re-synchronizes the operational-space controller and internal gripper command, reproducing uninterrupted simulation (state difference $2\times10^{-15}$, identical frames); repeated branches are bit-identical.
Demonstration states are $600$ training anchors in recorded layouts, with contexts rendered from recorded states.
Rollout states are four decisions per episode from each policy's own evaluation-protocol rollouts on five official initial states per task, branched after the episode.
Statistics are task means over the $40$ tasks, with sign-flip tests.

\smallskip
\qheading{Direction.}
$Q_{\mathrm{real}}(a_0+\hat g_a)>Q_{\mathrm{real}}(a_0-\hat g_a)$ holds on $72\%$ of demonstration states for \method and $75\%$ for \method-GH (\cref{tab:tf_realized}).
These rates are similar to the expert's ($74\%$ and $76\%$; $p_{\mathrm{task}}=0.16$ and $0.36$), and exceed random directions ($49\%$; $p_{\mathrm{task}}<10^{-5}$).
Rates remain similar on own-policy rollouts ($72\%$ and $68\%$) and at the expert's norm ($r=2.3$: $75\%$ and $77\%$), rising to $81\%$ and $82\%$ after two control steps ($p_{\mathrm{task}}<10^{-3}$), consistent with a short-horizon effect.

\smallskip
\qheading{Values.}
\method's deployed command is predicted to advance $1.29$ control steps and realizes $0.99$ (\method-GH: $1.18$ and $1.06$; neither difference is significant).
For the expert command the values are $1.24$ and $0.94$ ($1.15$ and $1.04$).
Executing $a_0$ realizes $0.00$ control steps against the query's $6.25$ (\cref{sec:ppp_finite_steps}). Its realized value stays within $\pm1.2$ in every gripper state and checkpoint, whereas the query value grows with training: the elevation belongs to the query.
The realized gripper contrast selects the expert's switch at $83\%$ of switch moments but favors switching on most hold states ($44\%$ correct).
Hold decisions therefore come from the trained margin ($95\%$ correct), not from the realized field.

\smallskip
\qheading{Gradient and magnitude.}
Imitation aligns $g_a$ with demonstrations, not the realized steepest ascent.
$\cos(g_a,g_{\mathrm{real}})$ is $0.26$ for \method and $0.28$ for \method-GH, equal to the expert direction's agreement with the same $g_{\mathrm{real}}$ ($0.26$ and $0.29$; $p_{\mathrm{task}}=0.87$ and $0.33$).
The realized slope is reproducible: finite differences at $\epsilon=0.3$ and $1$ agree at cosine $0.70$.
The co-trained gradient is also steeper than the realized one: within $\|a-a_0\|\le0.3$, \method has $12.1$ control steps per normalized unit against a realized $1.8$.
The frozen-WM field is not steeper ($0.9$ against $2.4$), consistent with the shorter linear range of jointly trained fields (\cref{tab:tf_linearity}).
On these executed branches, \method's deployed query is $101$ times worse than copying $z_t$ at its command, and $0.97$ times under the world model's statistics, yet the two derivatives agree with $g_{\mathrm{real}}$ equally ($0.26$ and $0.27$; \cref{sec:ppp_query_normalization}).
Such agreement does not require action supervision: the progress-only field also agrees with its realized slope ($0.30$), despite expert misalignment and closed-loop failure (\cref{tab:progress_vs_control}).

\smallskip
\qheading{Consistency and outcome.}
On successful versus failed rollout states, \method's $\cos(g_a,g_{\mathrm{real}})$ is $0.38$ versus $0.14$ (\method-GH: $0.38$ versus $0.12$), finite-step preference is $0.82$ versus $0.63$ (\method-GH: $0.81$ versus $0.58$), and its deployed command realizes $1.29$ versus $0.29$ control steps.
The realized field is steeper on failed states ($2.3$ versus $1.3$ control steps per unit), rather than flat. These observational comparisons do not establish that disagreement causes failure.

\smallskip
\qheading{Ascending the realized field.}
A controller estimates $g_{\mathrm{real}}$ at every decision from the twelve axis branches ($\epsilon=0.5$) and executes it at \method's command norm, with \method's gripper decision.
A matched \method arm runs and discards the same branches, differing only in executed pose direction.
Both use five official initial states per task and replan every control step instead of every tick; their rates are therefore not comparable with \cref{tab:libero_epoch120}.
Realized ascent succeeds in $1.5\%$ of $200$ episodes and \method in $66.0\%$ (better on $38$ of $40$ tasks, worse on none, $p_{\mathrm{task}}<10^{-5}$).
Exact dynamics alone do not yield a policy: imitation selects the demonstrated ascent direction.

\subsection{Field Visualizations}
\label{sec:ppp_field_vis}

\smallskip
\qheading{Slices.}
For a fixed state and instruction, we evaluate $\Delta Q(\delta u_i,\delta u_j)=Q(u_0+\delta u_ie_i+\delta u_je_j)-Q(u_0)$ on a $61\times61$ grid over $[-4,4]^2$ normalized units, with other pose coordinates and the gripper fixed at the reference.
The slice shows $p_\theta\circ f$, combining the field and learned dynamics, in control steps of progress of the state's demonstration.
Arrows project full six-dimensional vectors onto the plotted plane; $r_g=\|Pg_a\|/\|g_a\|$ and $r_E=\|P\bar u_E\|/\|\bar u_E\|$ report the captured norm fractions.
Each row shares a symmetric-log color scale, linear within $\pm0.5$ steps, so \method's large negative query values away from the reference do not obscure other fields.
\method's query potential rises only within about one normalized unit of the reference; we therefore recompute its panels on $81\times81$ grids over smaller windows: $[-1.5,1.5]^2$ on LIBERO, $[-2,2]^2$ for bun picking, and $[-1,1]^2$ for doll straightening.
The equally local label-free field uses \method's LIBERO zoom.
Zoomed panels retain the full-range color scale and shorten all arrows by a common factor to fit the longest, preserving relative lengths.

\smallskip
\qheading{Selection rule.}
For \cref{fig:teaser,fig:slices_all}, a rule fixed before drawing any slices takes one seeded draw per suite from states with $\rho\in[0.15,0.75]$, expert-command norm at least the suite median, and at least $80\%$ of that norm in the plotted plane.
The instruction figures use one seeded LIBERO-Goal demonstration and one seeded shared robot layout.
The doll-straightening figure uses one seeded demonstration among the $93$ with a single grasp, at fractions $\rho=0.2$, $0.4$, $0.6$, and $0.8$ of its length.

\smallskip
\qheading{Doll straightening.}
\Cref{fig:stitch_field} shows \method's field across states of the second real-robot task, which has a single instruction.
At all four training states, the deployed command aligns with the expert's ($\cos\geq0.996$) and is within $8\%$ of its norm.
As on LIBERO, the query potential rises only near the reference, except when holding the lifted doll ($\rho=0.6$), where it rises broadly along the expert's direction.
Here $\Delta Q$ is in time-to-go control steps, $1/T_u$ (\cref{sec:real_stitch_details}).

\subsection{Progress Along Closed-Loop Rollouts}
\label{sec:ppp_rollout_progress}

\smallskip
\qheading{Protocol.}
Each closed-loop LIBERO evaluation records $p_t=p_\theta(E(o_t),\ell)$ after every executed tick, using the planner's own context.
\Cref{fig:rollout_progress} and \cref{tab:ppp_rollout_progress} use the epoch-$120$ evaluations of \method without settling, BC+Progress (its progress output, which its actions do not use), and No $\mathcal L_p$, on the same $2{,}000$ initial states.
Each model follows its own closed-loop trajectories, ending at first success or the suite's time budget ($300$ ticks, $520$ on Long).
Panel (a) shows unsmoothed traces from different initial states of \emph{put both the alphabet soup and the tomato sauce in the basket}: a success, a failure with the first item placed and the second held, and a failure after dropping the first item.
For the population panels, each trace is resampled to $101$ points over its duration.
The affine map for the label-free scalar is the least-squares fit of the label $\rho-1$ on its values over the $7{,}849$ diagnostic states; it explains $24\%$ of the label variance, against $99\%$ for \method.
The $5$\,s value averages the five ticks ending at tick $100$; early AUROC uses only rollouts still running then. Of \method's $1{,}691$ such rollouts, $43\%$ later succeed.
AUROC intervals are percentile bootstraps over the $40$ tasks ($2{,}000$ resamples).

\smallskip
\qheading{Results.}
On successful rollouts, both calibrated scalars tend to rise and end near the terminal label; failed rollouts tend to end below it (\cref{tab:ppp_rollout_progress}).
\method's observed-state values yield success AUROCs of $0.84$ at rollout end and $0.78$ at $5$\,s; BC+Progress scores $0.89$ and $0.75$ on its own rollouts.
With the untuned threshold $p\geq-0.1$ (five-tick running mean; a tenth of a demonstration), $70\%$ of \method's successful rollouts register completion, a median $0.3$\,s before the simulator's success check ends the episode, and $30\%$ of its failed rollouts register it at some point (BC+Progress: $70\%$, $0.35$\,s, $30\%$).
The calibrated field thus measures demonstration-relative progress, which is informative about success but does not verify it.
Both calibrated fields have higher outcome AUROCs than No $\mathcal L_p$ on their respective rollouts; these metrics are invariant to positive affine rescaling.

%% file: sec/table/tf_alignment.tex
\begin{table*}[t]
    \centering
    \caption{Task-field diagnostics on $7{,}849$ LIBERO training states from $400$ trajectories (\cref{sec:ppp_gradient_protocol}).
    Progress-only (a)/(b): the epoch-$10$ fields that initialize \method-GH/\method; (a) continues with progress losses alone.
    Frozen WM uses \method-GH's losses with pretrained encoder and dynamics fixed (\cref{sec:ppp_frozen_runs}).
    $\rho$: mean per-trajectory Spearman correlation with progress; MAE: mean absolute error against $\rho_t-1$.
    cos: mean of per-trajectory mean $\cos(g_a,\bar u_E)$, with a $95\%$ bootstrap interval over trajectories; $>0$: fraction with $g_a^\top\bar u_E>0$.
    Orth.: expert-command energy fraction orthogonal to $g_a$, $1-\cos^2$.
    $\|g_a\|$: median gradient norm in control steps per normalized command unit.
    $Q_0$, $Q_E$: median increments $Q(u_0)-p_\theta(z_t,\ell)$ and $Q(\bar u_E)-p_\theta(z_t,\ell)$ in control steps; these are query values, not observed progress gains (\cref{sec:ppp_finite_steps}).
    Scale-dependent columns are dashed for No $\mathcal L_p$.
    States with $\|\bar u_E\|<0.25$ ($0.23\%$) are excluded from directional and value columns.
    Paired comparisons are reported in \cref{sec:ppp_field_results}.}
    \label{tab:tf_alignment}
    \footnotesize
    \setlength{\tabcolsep}{3.5pt}
    \begin{tabular}{@{}lrrrrcrrrrr@{}}
        \toprule
        Field & Epoch & $\rho$ & MAE & cos & 95\% CI & $>0$ & Orth. & $\|g_a\|$ & $Q_0$ & $Q_E$ \\
        \midrule
        Progress-only (a) & 10 & 0.9897 & 0.069 & 0.249 & [0.238, 0.260] & 0.75 & 0.80 & 0.51 & 0.45 & 1.11 \\
        Progress-only (a) & 20 & 0.9925 & 0.062 & 0.245 & [0.233, 0.257] & 0.75 & 0.80 & 0.50 & 0.45 & 1.08 \\
        Progress-only (a) & 40 & 0.9969 & 0.058 & 0.242 & [0.229, 0.255] & 0.75 & 0.80 & 0.51 & 0.47 & 1.11 \\
        Progress-only (a) & 70 & 0.9952 & 0.061 & 0.235 & [0.221, 0.248] & 0.74 & 0.81 & 0.52 & 0.48 & 1.12 \\
        Progress-only (a) & 120 & 0.9992 & 0.051 & 0.225 & [0.211, 0.238] & 0.73 & 0.81 & 0.54 & 0.50 & 1.13 \\
        Progress-only (b) & 10 & 0.9894 & 0.075 & 0.246 & [0.234, 0.258] & 0.75 & 0.80 & 0.51 & 0.44 & 1.12 \\
        BC+Progress & 10 & 0.9939 & 0.069 & 0.218 & [0.205, 0.231] & 0.73 & 0.80 & 0.60 & 0.48 & 1.13 \\
        BC+Progress & 20 & 0.9994 & 0.059 & 0.070 & [0.058, 0.083] & 0.55 & 0.80 & 1.78 & 0.16 & 1.05 \\
        BC+Progress & 40 & 0.9999 & 0.051 & 0.064 & [0.053, 0.076] & 0.56 & 0.81 & 3.01 & 0.11 & 0.98 \\
        BC+Progress & 70 & 1.0000 & 0.048 & 0.081 & [0.068, 0.094] & 0.58 & 0.82 & 3.46 & 0.10 & 0.92 \\
        BC+Progress & 120 & 0.9997 & 0.047 & 0.080 & [0.068, 0.093] & 0.58 & 0.82 & 4.18 & 0.05 & 0.77 \\
        \method-GH, frozen WM & 20 & 0.9900 & 0.069 & 0.822 & [0.814, 0.830] & 0.98 & 0.27 & 0.83 & 0.19 & 1.08 \\
        \method-GH, frozen WM & 40 & 0.9956 & 0.062 & 0.875 & [0.869, 0.881] & 0.99 & 0.20 & 0.82 & 0.21 & 1.12 \\
        \method-GH, frozen WM & 70 & 0.9970 & 0.062 & 0.910 & [0.905, 0.915] & 0.99 & 0.15 & 0.84 & 0.24 & 1.19 \\
        \method-GH, frozen WM & 120 & 0.9984 & 0.055 & 0.947 & [0.944, 0.950] & 1.00 & 0.09 & 0.83 & 0.26 & 1.23 \\
        No $\mathcal L_p$ & 20 & 0.6098 & -- & 0.955 & [0.953, 0.957] & 1.00 & 0.08 & -- & -- & -- \\
        No $\mathcal L_p$ & 40 & 0.5933 & -- & 0.983 & [0.983, 0.984] & 1.00 & 0.03 & -- & -- & -- \\
        No $\mathcal L_p$ & 70 & 0.5878 & -- & 0.992 & [0.992, 0.992] & 1.00 & 0.02 & -- & -- & -- \\
        No $\mathcal L_p$ & 120 & 0.5822 & -- & 0.996 & [0.996, 0.996] & 1.00 & 0.01 & -- & -- & -- \\
        \method-GH & 20 & 0.9994 & 0.053 & 0.950 & [0.948, 0.952] & 1.00 & 0.09 & 3.07 & 0.30 & 1.24 \\
        \method-GH & 40 & 0.9999 & 0.054 & 0.983 & [0.982, 0.983] & 1.00 & 0.03 & 4.32 & 0.29 & 1.24 \\
        \method-GH & 70 & 1.0000 & 0.048 & 0.992 & [0.991, 0.992] & 1.00 & 0.02 & 5.70 & 0.38 & 1.27 \\
        \method-GH & 120 & 0.9998 & 0.044 & 0.995 & [0.995, 0.996] & 1.00 & 0.01 & 7.25 & 0.40 & 1.14 \\
        \method & 20 & 0.9991 & 0.073 & 0.950 & [0.948, 0.952] & 1.00 & 0.09 & 4.37 & 0.94 & 1.93 \\
        \method & 40 & 0.9998 & 0.056 & 0.981 & [0.980, 0.982] & 1.00 & 0.04 & 8.11 & 2.26 & 1.89 \\
        \method & 70 & 1.0000 & 0.051 & 0.991 & [0.991, 0.991] & 1.00 & 0.02 & 13.40 & 4.61 & 1.04 \\
        \method & 120 & 0.9998 & 0.043 & 0.995 & [0.995, 0.995] & 1.00 & 0.01 & 17.29 & 6.92 & 0.57 \\
        \bottomrule
    \end{tabular}
\end{table*}

%% file: sec/table/tf_linearity.tex
\begin{table}[!tbp]
    \centering
    \caption{Finite-step structure of $Q$ at epoch $120$, with observations fixed (medians over diagnostic states unless noted).
    Ratio: $[Q(a_0+r\hat g_a)-Q(a_0)]/(r\|g_a\|)$; $1$ for linear $Q$, negative when $Q<Q(a_0)$.
    cos: $\cos(g_a(a_0),g_a(a_0+r\hat u_E))$; negative means a rotation $>90^\circ$. Expert alignment at $a_0$ is separate ($0.995$ for \method; \cref{tab:tf_alignment}).
    $Q{-}p$: $Q(a_0+r\hat g_a)-p_\theta(z_t,\ell)$ in control steps (demonstrated increment $\approx1.2$; dashed for No $\mathcal L_p$, whose units are free).
    pref.: mean per-trajectory fraction with $Q(a_0+r\hat u_E)>Q(a_0-r\hat u_E)$, comparing equal-size commands.
    $r$: normalized command units; median expert norm $2.3$.}
    \label{tab:tf_linearity}
    \footnotesize
    \setlength{\tabcolsep}{1.6pt}
    \begin{tabular}{@{}llrrrrrr@{}}
        \toprule
        Field & & \multicolumn{6}{c}{step size $r$} \\
        \cmidrule(l){3-8}
         & & 0.03 & 0.1 & 0.3 & 1 & 2 & 4 \\
        \midrule
        Progress-only (a) & ratio & 1.01 & 1.02 & 1.04 & 0.99 & 0.83 & 0.55 \\
         & cos & 1.00 & 1.00 & 0.98 & 0.84 & 0.67 & 0.44 \\
         & $Q{-}p$ & 0.51 & 0.56 & 0.68 & 1.07 & 1.43 & 1.71 \\
         & pref. & 0.73 & 0.73 & 0.74 & 0.77 & 0.83 & 0.85 \\
        BC+Progress & ratio & 1.31 & 1.29 & 0.32 & 0.11 & 0.07 & 0.04 \\
         & cos & 0.91 & 0.51 & 0.02 & 0.12 & 0.14 & 0.08 \\
         & $Q{-}p$ & 0.19 & 0.53 & 0.32 & 0.36 & 0.48 & 0.54 \\
         & pref. & 0.58 & 0.60 & 0.70 & 0.86 & 0.90 & 0.91 \\
        \method-GH, frozen WM & ratio & 1.00 & 1.00 & 0.98 & 0.81 & 0.57 & 0.37 \\
         & cos & 1.00 & 1.00 & 0.99 & 0.77 & 0.37 & 0.16 \\
         & $Q{-}p$ & 0.29 & 0.35 & 0.51 & 0.94 & 1.22 & 1.50 \\
         & pref. & 1.00 & 1.00 & 1.00 & 1.00 & 0.98 & 0.93 \\
        No $\mathcal L_p$ & ratio & 1.00 & 0.98 & 0.85 & 0.23 & 0.06 & 0.03 \\
         & cos & 1.00 & 1.00 & 0.96 & $-$0.66 & $-$0.13 & $-$0.15 \\
         & $Q{-}p$ & -- & -- & -- & -- & -- & -- \\
         & pref. & 1.00 & 1.00 & 1.00 & 0.99 & 0.93 & 0.85 \\
        \method-GH & ratio & 1.00 & 0.98 & 0.81 & 0.16 & 0.05 & 0.03 \\
         & cos & 1.00 & 1.00 & 0.92 & $-$0.66 & 0.22 & $-$0.09 \\
         & $Q{-}p$ & 0.61 & 1.12 & 2.11 & 1.50 & 1.09 & 1.14 \\
         & pref. & 1.00 & 1.00 & 1.00 & 1.00 & 0.98 & 0.96 \\
        \method & ratio & 0.99 & 0.94 & 0.49 & $-$0.27 & $-$0.15 & $-$0.08 \\
         & cos & 1.00 & 0.99 & $-$0.43 & $-$0.58 & 0.03 & $-$0.26 \\
         & $Q{-}p$ & 7.44 & 8.65 & 8.39 & 1.13 & 0.54 & 0.30 \\
         & pref. & 1.00 & 1.00 & 0.99 & 0.89 & 0.79 & 0.71 \\
        \bottomrule
    \end{tabular}
\end{table}

%% file: sec/table/tf_instruction.tex
\begin{table}[ht]
    \centering
    \caption{Instruction conditioning at episode start (first frame repeated across context, open hand), under all instructions trained in the scene; epoch $120$ unless noted.
    Hit: fraction of (state, instruction) queries whose command/gradient direction is closest to the instructed reference.
    References: the instructed color's demonstration in the same layout ($2$-s hand displacement, table plane; robot), or the task's mean $1$-s hand displacement over $50$ demonstrations (LIBERO-Goal; pairwise reference cosine $0.63$).
    Swap cos: mean cosine between readouts for swapped and original instructions; lower indicates greater directional change.}
    \label{tab:tf_instruction}
    \footnotesize
    \setlength{\tabcolsep}{2.5pt}
    \begin{tabular}{@{}llrrr@{}}
        \toprule
        Setting & Readout & Hit & Chance & Swap cos \\
        \midrule
        Robot & \method (300), command & 0.52 & 0.25 & 0.40 \\
         & BC (300), cmd. & 0.51 & 0.25 & 0.15 \\
        LIBERO-Goal & Progress-only (10), grad. & 0.13 & 0.10 & 0.31 \\
         & Progress-only, grad. & 0.10 & 0.10 & 0.57 \\
         & BC+Progress, grad. & 0.10 & 0.10 & 0.33 \\
         & BC+Progress, BC cmd. & 0.33 & 0.10 & 0.71 \\
         & \method-GH, frozen WM, cmd. & 0.21 & 0.10 & 0.59 \\
         & \method-GH, command & 0.33 & 0.10 & 0.68 \\
         & \method, command & 0.29 & 0.10 & 0.71 \\
        \bottomrule
    \end{tabular}
\end{table}

%% file: sec/table/tf_realized.tex
\begin{table*}[!tbp]
    \centering
    \caption{The field under true dynamics, epoch $120$ (\cref{sec:ppp_true_dynamics}).
        $Q_{\mathrm{real}}(a)=p_\theta(E(o^{\mathrm{sim}}_{t+k}(a)),\ell)$ scores the frame reached by holding $a$ for $k=4$ ticks (world-model horizon) from the restored state, using the field's encoder and potential.
        Demo: $600$ training states ($40$ tasks, $3$ demonstrations, $5$ anchors); rollouts: $800$ states per policy ($40$ tasks, $5$ initial states, $4$ decisions).
        pref.: fraction of states with $Q_{\mathrm{real}}(a_0+r\hat d)>Q_{\mathrm{real}}(a_0-r\hat d)$ at $r=1$, along $\hat d=\hat g_a$, expert $\hat u_E$, or a random unit direction.
        cos: $\cos(g_a,g_{\mathrm{real}})$ and $\cos(\bar u_E,g_{\mathrm{real}})$; $g_{\mathrm{real}}$ is the least-squares slope of $Q_{\mathrm{real}}$ over executed commands with $\|a-a_0\|\le1$.
        $Q_0$, $Q_{\hat a}$: query-predicted / realized increments over $p_\theta(z_t,\ell)$ (control steps) at $a_0$ and the deployed command, respectively.
        Means over $40$ tasks; ``--'': not executed for that field.}
    \label{tab:tf_realized}
    \footnotesize
    \setlength{\tabcolsep}{3pt}
    \begin{tabular}{@{}llccccccc@{}}
        \toprule
        Field & States & pref.\ $\hat g_a$ & pref.\ $\hat u_E$ & pref.\ random & $\cos g_a$ & $\cos\bar u_E$ & $Q_0$ & $Q_{\hat a}$ \\
        \midrule
        \method & demo & 0.72 & 0.74 & 0.49 & 0.26 & 0.26 & 6.25 / 0.00 & 1.29 / 0.99 \\
         & own rollouts & 0.72 & -- & 0.50 & 0.26 & -- & 6.54 / $-$0.08 & 2.18 / 0.86 \\
        \method-GH & demo & 0.75 & 0.76 & 0.49 & 0.28 & 0.29 & 0.23 / $-$0.15 & 1.18 / 1.06 \\
         & own rollouts & 0.68 & -- & 0.50 & 0.23 & -- & 0.70 / $-$0.21 & 1.29 / 0.79 \\
        \method-GH, frozen WM & demo & -- & -- & -- & 0.31 & 0.30 & 0.29 / $-$0.05 & -- \\
         & \method's rollouts & -- & -- & -- & 0.39 & -- & 0.45 / $-$0.45 & -- \\
        Progress-only, frozen WM & demo & -- & -- & -- & 0.30 & 0.26 & 0.75 / $-$0.19 & -- \\
         & \method's rollouts & -- & -- & -- & 0.30 & -- & 0.80 / $-$0.59 & -- \\
        \bottomrule
    \end{tabular}
\end{table*}

%% file: refers.bib
@String(ICCV= {Int. Conf. Comput. Vis.})

@String(ECCV= {Eur. Conf. Comput. Vis.})

@String(NIPS= {Adv. Neural Inform. Process. Syst.})

@String(ICLR = {Int. Conf. Learn. Represent.})

@String(AAAI = {AAAI})

@String(ICCV  = {ICCV})

@String(ECCV  = {ECCV})

@String(NIPS  = {NeurIPS})

@String(ICML  = {ICML})

@String(RSS   = {RSS})

@String(CoRL  = {CoRL})

@String(EMNLP_IJCNLP = {EMNLP-IJCNLP})

@String(ICLR  = {ICLR})

@inproceedings{maes2026leworldmodel,
  title     = {{LeWorldModel}: Stable End-to-End Joint-Embedding Predictive Architecture from Pixels},
  author    = {Maes, Lucas and Le Lidec, Quentin and Scieur, Damien and LeCun, Yann and Balestriero, Randall},
  booktitle = NIPS,
  year      = {2026}
}

@article{mclachlan1998unified,
  title   = {Unified Approach to {Hamiltonian} Systems, {Poisson} Systems, Gradient Systems, and Systems with {Lyapunov} Functions or First Integrals},
  author  = {McLachlan, Robert I. and Quispel, G. R. W. and Robidoux, Nicolas},
  journal = {Physical Review Letters},
  volume  = {81},
  number  = {12},
  pages   = {2399--2403},
  year    = {1998}
}

@article{balestriero2025lejepa,
  title   = {{LeJEPA}: Provable and Scalable Self-Supervised Learning Without the Heuristics},
  author  = {Balestriero, Randall and LeCun, Yann},
  journal = {arXiv preprint arXiv:2511.08544},
  year    = {2025}
}

@inproceedings{perez2018film,
  title     = {{FiLM}: Visual Reasoning with a General Conditioning Layer},
  author    = {Perez, Ethan and Strub, Florian and de Vries, Harm and Dumoulin, Vincent and Courville, Aaron},
  booktitle = AAAI,
  year      = {2018}
}

@inproceedings{reimers2019sentencebert,
  title     = {Sentence-{BERT}: Sentence Embeddings using Siamese {BERT}-Networks},
  author    = {Reimers, Nils and Gurevych, Iryna},
  booktitle = EMNLP_IJCNLP,
  year      = {2019}
}

@inproceedings{wang2020minilm,
  title     = {{MiniLM}: Deep Self-Attention Distillation for Task-Agnostic Compression of Pre-Trained Transformers},
  author    = {Wang, Wenhui and Wei, Furu and Dong, Li and Bao, Hangbo and Yang, Nan and Zhou, Ming},
  booktitle = NIPS,
  year      = {2020}
}

@inproceedings{liu2023libero,
  title     = {{LIBERO}: Benchmarking Knowledge Transfer for Lifelong Robot Learning},
  author    = {Liu, Bo and Zhu, Yifeng and Gao, Chongkai and Feng, Yihao and Liu, Qiang and Zhu, Yuke and Stone, Peter},
  booktitle = NIPS,
  year      = {2023}
}

@inproceedings{chi2023diffusionpolicy,
  title     = {{Diffusion Policy}: Visuomotor Policy Learning via Action Diffusion},
  author    = {Chi, Cheng and Feng, Siyuan and Du, Yilun and Xu, Zhenjia and Cousineau, Eric and Burchfiel, Benjamin CM and Song, Shuran},
  booktitle = RSS,
  year      = {2023}
}

@inproceedings{kim2025openvla,
  title     = {{OpenVLA}: An Open-Source Vision-Language-Action Model},
  author    = {Kim, Moo Jin and Pertsch, Karl and Karamcheti, Siddharth and Xiao, Ted and Balakrishna, Ashwin and Nair, Suraj and Rafailov, Rafael and Foster, Ethan P. and Sanketi, Pannag R. and Vuong, Quan and Kollar, Thomas and Burchfiel, Benjamin and Tedrake, Russ and Sadigh, Dorsa and Levine, Sergey and Liang, Percy and Finn, Chelsea},
  booktitle = CoRL,
  year      = {2025}
}

@inproceedings{black2025pi0,
  title     = {{$\pi_0$}: A Vision-Language-Action Flow Model for General Robot Control},
  author    = {Black, Kevin and Brown, Noah and Driess, Danny and Esmail, Adnan and Equi, Michael Robert and Finn, Chelsea and Fusai, Niccolo and Groom, Lachy and Hausman, Karol and Ichter, Brian and Jakubczak, Szymon and Jones, Tim and Ke, Liyiming and Levine, Sergey and Li-Bell, Adrian and Mothukuri, Mohith and Nair, Suraj and Pertsch, Karl and Shi, Lucy Xiaoyang and Smith, Laura and Tanner, James and Vuong, Quan and Walling, Anna and Wang, Haohuan and Zhilinsky, Ury},
  booktitle = RSS,
  year      = {2025}
}

@inproceedings{zhou2025dinowm,
  title     = {{DINO-WM}: World Models on Pre-trained Visual Features Enable Zero-shot Planning},
  author    = {Zhou, Gaoyue and Pan, Hengkai and LeCun, Yann and Pinto, Lerrel},
  booktitle = ICML,
  year      = {2025}
}

@article{assran2025vjepa2,
  title   = {{V-JEPA 2}: Self-Supervised Video Models Enable Understanding, Prediction and Planning},
  author  = {Assran, Mahmoud and Bardes, Adrien and Fan, David and Garrido, Quentin and Howes, Russell and Komeili, Mojtaba and Muckley, Matthew and Rizvi, Ammar and Roberts, Claire and Sinha, Koustuv and Zholus, Artem and Arnaud, Sergio and Gejji, Abha and Martin, Ada and Robert Hogan, Francois and Dugas, Daniel and Bojanowski, Piotr and Khalidov, Vasil and Labatut, Patrick and Massa, Francisco and Szafraniec, Marc and Krishnakumar, Kapil and Li, Yong and Ma, Xiaodong and Chandar, Sarath and Meier, Franziska and LeCun, Yann and Rabbat, Michael and Ballas, Nicolas},
  journal = {arXiv preprint arXiv:2506.09985},
  year    = {2025}
}

@inproceedings{lyu2025dywa,
  title     = {{DyWA}: Dynamics-adaptive World Action Model for Generalizable Non-prehensile Manipulation},
  author    = {Lyu, Jiangran and Li, Ziming and Shi, Xuesong and Xu, Chaoyi and Wang, Yizhou and Wang, He},
  booktitle = ICCV,
  year      = {2025}
}

@article{ye2026dreamzero,
  title   = {World Action Models Are Zero-shot Policies},
  author  = {Ye, Seonghyeon and Ge, Yunhao and Zheng, Kaiyuan and Gao, Shenyuan and Yu, Sihyun and Kurian, George and Indupuru, Suneel and Tan, You Liang and Zhu, Chuning and Xiang, Jiannan and Malik, Ayaan and Lee, Kyungmin and Liang, William and Ranawaka, Nadun and Gu, Jiasheng and Xu, Yinzhen and Wang, Guanzhi and Hu, Fengyuan and Narayan, Avnish and Bjorck, Johan and Wang, Jing and Kim, Gwanghyun and Niu, Dantong and Zheng, Ruijie and Xie, Yuqi and Wu, Jimmy and Wang, Qi and Julian, Ryan and Xu, Danfei and Du, Yilun and Chebotar, Yevgen and Reed, Scott and Kautz, Jan and Zhu, Yuke and Fan, Linxi ``Jim'' and Jang, Joel},
  journal = {arXiv preprint arXiv:2602.15922},
  year    = {2026}
}

@inproceedings{ma2023vip,
  title     = {{VIP}: Towards Universal Visual Reward and Representation via Value-Implicit Pre-Training},
  author    = {Ma, Yecheng Jason and Sodhani, Shagun and Jayaraman, Dinesh and Bastani, Osbert and Kumar, Vikash and Zhang, Amy},
  booktitle = ICLR,
  year      = {2023}
}

@inproceedings{ma2023liv,
  title     = {{LIV}: Language-Image Representations and Rewards for Robotic Control},
  author    = {Ma, Yecheng Jason and Kumar, Vikash and Zhang, Amy and Bastani, Osbert and Jayaraman, Dinesh},
  booktitle = ICML,
  year      = {2023}
}

@inproceedings{zhang2026vlac,
  title     = {A Generalist Pair-wise Progress Critic Model for Vision-Language-Action Robots},
  author    = {Zhang, Qi and Zhai, Shaopeng and Zhang, Shengzhe and Liu, Litao and Zhang, Tianyi and Huang, Fuxian and Zhou, Ming},
  booktitle = ICML,
  year      = {2026}
}

@inproceedings{liang2026robometer,
  title     = {{Robometer}: Scaling General-Purpose Robotic Reward Models via Trajectory Comparisons},
  author    = {Liang, Anthony and Korkmaz, Yigit and Zhang, Jiahui and Hwang, Minyoung and Anwar, Abrar and Kaushik, Sidhant and Shah, Aditya and Huang, Alex S. and Zettlemoyer, Luke and Fox, Dieter and Xiang, Yu and Li, Anqi and Bobu, Andreea and Gupta, Abhishek and Tu, Stephen and B{\i}y{\i}k, Erdem and Zhang, Jesse},
  booktitle = RSS,
  year      = {2026}
}

@inproceedings{zhao2026tapsampling,
  title     = {{TapSampling}: Inference-Time Sampling with a Task-Progress-Understanding Verifier for Robotic Manipulation},
  author    = {Zhao, Sizhe and Zhang, Shengping and Yang, Shuo and Zhao, Weiyu and Wang, Shuigen and Ji, Xiangyang},
  booktitle = ICML,
  year      = {2026}
}

@article{yan2026progressvla,
  title   = {{ProgressVLA}: Progress-Guided Diffusion Policy for Vision-Language Robotic Manipulation},
  author  = {Yan, Hongyu and Li, Qiwei and Yang, Jiaolong and Mu, Yadong},
  journal = {arXiv preprint arXiv:2603.27670},
  year    = {2026}
}

@article{khansari2017potential,
  title   = {Learning Potential Functions from Human Demonstrations with Encapsulated Dynamic and Compliant Behaviors},
  author  = {Khansari-Zadeh, Seyed Mohammad and Khatib, Oussama},
  journal = {Autonomous Robots},
  volume  = {41},
  number  = {1},
  pages   = {45--69},
  year    = {2017}
}

@inproceedings{florence2022implicit,
  title     = {{Implicit Behavioral Cloning}},
  author    = {Florence, Pete and Lynch, Corey and Zeng, Andy and Ramirez, Oscar A. and Wahid, Ayzaan and Downs, Laura and Wong, Adrian and Lee, Johnny and Mordatch, Igor and Tompson, Jonathan},
  booktitle = CoRL,
  year      = {2022}
}

@inproceedings{srinivas2018universal,
  title     = {{Universal Planning Networks}: Learning Generalizable Representations for Visuomotor Control},
  author    = {Srinivas, Aravind and Jabri, Allan and Abbeel, Pieter and Levine, Sergey and Finn, Chelsea},
  booktitle = ICML,
  year      = {2018}
}

@inproceedings{wan2024difftori,
  title     = {{DiffTORI}: Differentiable Trajectory Optimization for Deep Reinforcement and Imitation Learning},
  author    = {Wan, Weikang and Wang, Ziyu and Wang, Yufei and Erickson, Zackory and Held, David},
  booktitle = NIPS,
  year      = {2024}
}

@inproceedings{psenka2026grasp,
  title     = {Parallel Stochastic Gradient-Based Planning for World Models},
  author    = {Psenka, Michael and Rabbat, Michael and Krishnapriyan, Aditi S. and LeCun, Yann and Bar, Amir},
  booktitle = ICML,
  year      = {2026}
}

@article{rubinstein1999cross,
  title   = {The Cross-Entropy Method for Combinatorial and Continuous Optimization},
  author  = {Rubinstein, Reuven},
  journal = {Methodology and Computing in Applied Probability},
  volume  = {1},
  number  = {2},
  pages   = {127--190},
  year    = {1999}
}

@inproceedings{hafner2020dreamer,
  title     = {Dream to Control: Learning Behaviors by Latent Imagination},
  author    = {Hafner, Danijar and Lillicrap, Timothy and Ba, Jimmy and Norouzi, Mohammad},
  booktitle = ICLR,
  year      = {2020}
}

@inproceedings{janner2022diffuser,
  title     = {Planning with Diffusion for Flexible Behavior Synthesis},
  author    = {Janner, Michael and Du, Yilun and Tenenbaum, Joshua B. and Levine, Sergey},
  booktitle = ICML,
  year      = {2022}
}

@inproceedings{abbeel2004apprenticeship,
  title     = {Apprenticeship Learning via Inverse Reinforcement Learning},
  author    = {Abbeel, Pieter and Ng, Andrew Y.},
  booktitle = ICML,
  year      = {2004}
}

@inproceedings{ziebart2008maxent,
  title     = {Maximum Entropy Inverse Reinforcement Learning},
  author    = {Ziebart, Brian D. and Maas, Andrew and Bagnell, J. Andrew and Dey, Anind K.},
  booktitle = AAAI,
  year      = {2008}
}

@article{khatib1986potential,
  title   = {Real-Time Obstacle Avoidance for Manipulators and Mobile Robots},
  author  = {Khatib, Oussama},
  journal = {The International Journal of Robotics Research},
  volume  = {5},
  number  = {1},
  pages   = {90--98},
  year    = {1986}
}

@article{rimon1992navigation,
  title   = {Exact Robot Navigation Using Artificial Potential Functions},
  author  = {Rimon, Elon and Koditschek, Daniel E.},
  journal = {IEEE Transactions on Robotics and Automation},
  volume  = {8},
  number  = {5},
  pages   = {501--518},
  year    = {1992}
}

@article{khansari2011seds,
  title   = {Learning Stable Nonlinear Dynamical Systems With {Gaussian} Mixture Models},
  author  = {Khansari-Zadeh, Seyed Mohammad and Billard, Aude},
  journal = {IEEE Transactions on Robotics},
  volume  = {27},
  number  = {5},
  pages   = {943--957},
  year    = {2011}
}

@inproceedings{hansen2022tdmpc,
  title     = {Temporal Difference Learning for Model Predictive Control},
  author    = {Hansen, Nicklas A. and Su, Hao and Wang, Xiaolong},
  booktitle = ICML,
  year      = {2022}
}

@inproceedings{czarnecki2017sobolev,
  title     = {Sobolev Training for Neural Networks},
  author    = {Czarnecki, Wojciech M. and Osindero, Simon and Jaderberg, Max and {\'S}wirszcz, Grzegorz and Pascanu, Razvan},
  booktitle = NIPS,
  year      = {2017}
}

@inproceedings{zhong2026gomp,
  title     = {Grasp-Oriented Non-Prehensile Manipulation via Learning a Graspability Field},
  author    = {Zhong, Licheng and Lee, Gim Hee},
  booktitle = ECCV,
  year      = {2026}
}

@inproceedings{zhao2023act,
  title     = {Learning Fine-Grained Bimanual Manipulation with Low-Cost Hardware},
  author    = {Zhao, Tony Z. and Kumar, Vikash and Levine, Sergey and Finn, Chelsea},
  booktitle = RSS,
  year      = {2023}
}

@inproceedings{ghosh2024octo,
  title     = {{Octo}: An Open-Source Generalist Robot Policy},
  author    = {Ghosh, Dibya and Walke, Homer Rich and Pertsch, Karl and Black, Kevin and Mees, Oier and Dasari, Sudeep and Hejna, Joey and Kreiman, Tobias and Xu, Charles and Luo, Jianlan and Tan, You Liang and Chen, Lawrence Yunliang and Vuong, Quan and Xiao, Ted and Sanketi, Pannag R. and Sadigh, Dorsa and Finn, Chelsea and Levine, Sergey},
  booktitle = RSS,
  year      = {2024}
}

@inproceedings{zitkovich2023rt2,
  title     = {{RT-2}: Vision-Language-Action Models Transfer Web Knowledge to Robotic Control},
  author    = {Zitkovich, Brianna and Yu, Tianhe and Xu, Sichun and Xu, Peng and Xiao, Ted and Xia, Fei and Wu, Jialin and Wohlhart, Paul and Welker, Stefan and Wahid, Ayzaan and Vuong, Quan and Vanhoucke, Vincent and Tran, Huong and Soricut, Radu and Singh, Anikait and Singh, Jaspiar and Sermanet, Pierre and Sanketi, Pannag R. and Salazar, Grecia and Ryoo, Michael S. and Reymann, Krista and Rao, Kanishka and Pertsch, Karl and Mordatch, Igor and Michalewski, Henryk and Lu, Yao and Levine, Sergey and Lee, Lisa and Lee, Tsang-Wei Edward and Leal, Isabel and Kuang, Yuheng and Kalashnikov, Dmitry and Julian, Ryan and Joshi, Nikhil J. and Irpan, Alex and Ichter, Brian and Hsu, Jasmine and Herzog, Alexander and Hausman, Karol and Gopalakrishnan, Keerthana and Fu, Chuyuan and Florence, Pete and Finn, Chelsea and Dubey, Kumar Avinava and Driess, Danny and Ding, Tianli and Choromanski, Krzysztof Marcin and Chen, Xi and Chebotar, Yevgen and Carbajal, Justice and Brown, Noah and Brohan, Anthony and Arenas, Montserrat Gonzalez and Han, Kehang},
  booktitle = CoRL,
  year      = {2023}
}

@inproceedings{wu2024gr1,
  title     = {Unleashing Large-Scale Video Generative Pre-training for Visual Robot Manipulation},
  author    = {Wu, Hongtao and Jing, Ya and Cheang, Chilam and Chen, Guangzeng and Xu, Jiafeng and Li, Xinghang and Liu, Minghuan and Li, Hang and Kong, Tao},
  booktitle = ICLR,
  year      = {2024}
}

@inproceedings{tian2025seer,
  title     = {Predictive Inverse Dynamics Models are Scalable Learners for Robotic Manipulation},
  author    = {Tian, Yang and Yang, Sizhe and Zeng, Jia and Wang, Ping and Lin, Dahua and Dong, Hao and Pang, Jiangmiao},
  booktitle = ICLR,
  year      = {2025}
}

@inproceedings{heess2015svg,
  title     = {Learning Continuous Control Policies by Stochastic Value Gradients},
  author    = {Heess, Nicolas and Wayne, Gregory and Silver, David and Lillicrap, Timothy and Erez, Tom and Tassa, Yuval},
  booktitle = NIPS,
  year      = {2015}
}

@inproceedings{ze2024dp3,
  title     = {{3D Diffusion Policy}: Generalizable Visuomotor Policy Learning via Simple {3D} Representations},
  author    = {Ze, Yanjie and Zhang, Gu and Zhang, Kangning and Hu, Chenyuan and Wang, Muhan and Xu, Huazhe},
  booktitle = RSS,
  year      = {2024}
}

@inproceedings{rana2020euclideanizing,
  title     = {Euclideanizing Flows: Diffeomorphic Reduction for Learning Stable Dynamical Systems},
  author    = {Rana, Muhammad Asif and Li, Anqi and Fox, Dieter and Boots, Byron and Ramos, Fabio and Ratliff, Nathan},
  booktitle = {L4DC},
  year      = {2020}
}

@inproceedings{amos2018differentiablempc,
  title     = {Differentiable {MPC} for End-to-end Planning and Control},
  author    = {Amos, Brandon and Jimenez, Ivan and Sacks, Jacob and Boots, Byron and Kolter, J. Zico},
  booktitle = NIPS,
  year      = {2018}
}

@inproceedings{hansen2024tdmpc2,
  title     = {{TD-MPC2}: Scalable, Robust World Models for Continuous Control},
  author    = {Hansen, Nick and Su, Hao and Wang, Xiaolong},
  booktitle = ICLR,
  year      = {2024}
}

@inproceedings{cheng2020rmpflow,
  title     = {{RMPflow}: A Computational Graph for Automatic Motion Policy Generation},
  author    = {Cheng, Ching-An and Mukadam, Mustafa and Issac, Jan and Birchfield, Stan and Fox, Dieter and Boots, Byron and Ratliff, Nathan},
  booktitle = {Algorithmic Foundations of Robotics XIII},
  pages     = {441--457},
  year      = {2020}
}

@inproceedings{mukadam2020rmpfusion,
  title     = {Riemannian Motion Policy Fusion through Learnable {Lyapunov} Function Reshaping},
  author    = {Mukadam, Mustafa and Cheng, Ching-An and Fox, Dieter and Boots, Byron and Ratliff, Nathan},
  booktitle = CoRL,
  year      = {2020}
}

@inproceedings{sontakke2023roboclip,
  title     = {{RoboCLIP}: One Demonstration is Enough to Learn Robot Policies},
  author    = {Sontakke, Sumedh and Zhang, Jesse and Arnold, S{\'e}b and Pertsch, Karl and B{\i}y{\i}k, Erdem and Sadigh, Dorsa and Finn, Chelsea and Itti, Laurent},
  booktitle = NIPS,
  year      = {2023}
}
